%% file: latex/acl_latex.tex
\documentclass[11pt]{article}

\usepackage[preprint]{latex/acl}

\usepackage{times}
\usepackage{latexsym}

\usepackage[T1]{fontenc}

\usepackage[utf8]{inputenc}

\usepackage{microtype}

\usepackage{inconsolata}

\usepackage{graphicx}

\usepackage[inline]{enumitem}

\usepackage[british]{babel}
\usepackage[babel=true]{csquotes}

\usepackage{amsmath}
\usepackage{amssymb}
\DeclareMathOperator*{\argmax}{arg\,max}

\usepackage{listings}
\usepackage{tabularx}
\usepackage{booktabs}
\usepackage[table]{xcolor}
\usepackage{graphicx}
\usepackage{pgf}
\usepackage{multirow}
\usepackage{nicematrix}

\usepackage{contour}
\usepackage{ulem}
\contourlength{0.8pt}

\usepackage{pgfplots}
\pgfplotsset{compat=1.18}
\usetikzlibrary{pgfplots.statistics}

\newcommand{\gradientcell}[7]{
    \def\printValue##1{\ifnum#7=2 \else \ifnum#7=1 \textbf{#1}\else #1\fi\fi}
    \ifdimcomp{#1pt}{>}{#3 pt}{\cellcolor{#5!100.0!#4!#6}\printValue{#1}}{
    \ifdimcomp{#1pt}{<}{#2 pt}{\cellcolor{#5!0.0!#4!#6}\printValue{#1}}{
         \pgfmathparse{int(round(100*(#1/(#3-#2))-(#2 *(100/(#3-#2)))))}
        \xdef\tempa{\pgfmathresult}
        \cellcolor{#5!\tempa!#4!#6}\hspace*{-9pt}\printValue{#1}\hspace*{-10pt}
    }}
}

\usepackage{placeins}

\newcommand{\percentGrad}[2]{\gradientcell{#1}{0}{100}{cyan}{yellow}{70}{#2}}

\newcolumntype{C}{>{\centering\arraybackslash}X}

\usepackage[acronym]{glossaries}
\glsdisablehyper
\newacronym{llm}{LLM}{large language model}
\newacronym{llms}{LLMs}{large language models}
\newacronym{cla}{CLA}{cross-lingual alignment}
\newacronym{pmi}{PMI}{point-wise mutual information}
\newacronym{pwm}{p-wm}{position-weighted mean}
\newacronym{anc}{ANC}{average neuron-wise correlation}

\let\origfootnote\footnote
\renewcommand{\footnote}[1]{\kern.1em\origfootnote{#1}}
\newcommand{\punctfootnote}[1]{\kern-.1em\origfootnote{#1}}

\title{Predicting Multilingual Classification and Translation Performance of LLMs with Cross-Lingual Alignment -- Is English Enough?}

\author{
    \textbf{Adnan Al Ali\textsuperscript{1},}
    \textbf{Kathy Hämmerl\textsuperscript{2,3},}
    \\
    \textbf{Jindřich Libovický\textsuperscript{1},}
    \textbf{Alexander Fraser\textsuperscript{2,3}}
    \\
    \textsuperscript{1}Charles University, Faculty of Mathematics and Physics, Czech Republic \\
    \textsuperscript{2}Technical University of Munich, Germany \\
    \textsuperscript{3}Munich Center for Machine Learning \\
   \small{
   \textbf{Correspondence:} \href{mailto:alali@ufal.mff.cuni.cz}{\texttt{alali@ufal.mff.cuni.cz}}
 }
}

\begin{document}
\maketitle
\begin{abstract}

Multilingual \acrfull{llms} have been shown to perform better on non-English classification tasks when the~representations of the~given language are more aligned to English within the~model.
Several \acrfull{cla} scores have been proposed for use with LLMs, along with multiple approaches for extracting embeddings from the~models.
We provide a~comparative analysis of 27 \acrshort{cla} score variants, examining how they differ and how well each predicts downstream performance across three tasks.
Crucially, while \acrshort{llms} are widely used for generative tasks such as machine translation, prior work has focused almost exclusively on classification.
We therefore investigate whether \acrshort{cla} scores are similarly predictive of translation performance.
To enable computing correlations across target languages, we propose a \acrshort{pmi}-based translation metric, which
is less dependent on the~target language and correlates strongly with \textsc{chrF}.
We find that \acrshort{cla} with English predicts translation quality comparably to or better than source--target \acrshort{cla}, providing new evidence that \acrshort{llms} use English as an internal pivot language.

\end{abstract}

\section{Introduction}

Multilingual \acrfull{llms} exhibit remarkable multilingual capabilities \citep{singh-etal-2025-global} in tasks such as machine translation \citep{goyal-etal-2022-flores}, reading comprehension \citep{bandarkar-etal-2024-belebele}, and text classification \citep{adelani-etal-2024-sib}, among others. While these capabilities have been extensively benchmarked, we have little understanding of how multilinguality works.

\Acrfull{cla} scores can reliably predict LLM performance on non-English downstream tasks without needing to explicitly run the~tasks, as \citet{ravisankar2026mapenglishrolecrosslingual-dali} and \citet{kargaran-etal-2025-mexa} show.
They specifically measure alignment scores with English, based on the~claim that \acrshort{llms} implicitly translate prompts to English in their hidden states \citep{wendler-etal-2024-llamas-think}.
Alignment scores can be obtained from embeddings \citep[\it inter alia]{kargaran-etal-2025-mexa,del-fishel-2022-cross-anc,artetxe-schwenk-2019-margin-xsim} or tokenisation \citep{hammerl-etal-2025-beyond, limisiewicz-etal-2023-tokenization}.

In this work, we comprehensively review existing methods for obtaining sentence-embedding-based \acrshort{cla} scores, including sentence embedding extraction.
We test the~correlation between the~scores and performance on three tasks:
\begin{enumerate*}[label=(\arabic*)]
    \item \emph{SIB-200} (text categorisation; \citealp{adelani-etal-2024-sib}),
    \item \emph{Belebele} (reading comprehension; \citealp{bandarkar-etal-2024-belebele}), and
    \item \emph{Flores-200} \& \emph{BOUQuET} (machine translation; \citealp{nllbteam2022languageleftbehindscaling-flores200}; \citealp{andrews-etal-2025-bouquet}).
\end{enumerate*}

\acrshort{cla} has been studied extensively on encoder-only models, but only a~limited number of recent papers focus on decoder-only models \citep{hammerl-etal-2025-beyond, ravisankar2026mapenglishrolecrosslingual-dali, kargaran-etal-2025-mexa}, and none address the~translation task, which remains understudied despite its popularity.

Therefore, our work concentrates on translation. To our knowledge, no prior work has inspected the~impact of \acrshort{cla} scores on translation performance. 
We aim to answer an essential question: when translating from language \texttt{src} to \texttt{tgt},  which \acrshort{cla} is more predictive of quality:
\begin{enumerate*}[label=(\arabic*)]
 \item \texttt{src-tgt},
 \item \texttt{src-en}, or
 \item \texttt{en-tgt}?
\end{enumerate*}
To answer this, we propose a~translation evaluation metric based on \acrfull{pmi} which performs similarly to \textsc{chrF} but depends less on the~target language -- a~property crucial to our analysis.

We analyse the~contributions of \acrshort{cla} scores to prediction by finding optimal weights in an ensemble of the~different alignment directions.
Our findings suggest that \acrshort{cla} scores with English (\texttt{src-en} and \texttt{en-tgt}) are generally more predictive, contributing new evidence for the~claim that \acrshort{llms} use English as a~pivot language from a~yet unexplored perspective.

The~main contributions of this work are:
\begin{enumerate*}[label=\arabic*:]
\item A~comprehensive review of existing methods for measuring \acrshort{cla} scores from sentence embeddings.
\item A new multilingual \textbf{fewshot} method for extracting sentence embeddings from LLMs (Section~\ref{sec:experiments-cla}).
\item Expanding the~study of \acrshort{cla} in \acrshort{llm}s to machine translation (Section~\ref{sec:translation}).
\item A~new translation evaluation metric that is less target-language dependent (Section~\ref{sec:translation-pmi}).
\item New evidence for the~claim that \acrshort{llm}s use English as an internal pivot language.
\end{enumerate*}
We publish our code on GitHub.\footnote{\url{https://github.com/KathyHaem/cla-metrics}}

\section{Background and Related Work}\label{sec:background}

\paragraph{Cross-lingual alignment.} \acrshort{cla}, in the~broad sense, is \enquote{the~similar representation of similar meanings regardless of the~language} \citep{hammerl-etal-2024-understanding}. Previous work has used various methods of measuring this concept to estimate zero-shot cross-lingual transfer in encoder-only models \citep{conneau-etal-2020-emerging,del-fishel-2022-cross-anc,limisiewicz-etal-2023-tokenization} and decoder-only performance based on alignment with English \citep{kargaran-etal-2025-mexa,ravisankar2026mapenglishrolecrosslingual-dali}. While we focus on the~latter, methods developed for encoder models can be adapted to decoders with minimal modifications.

Embedding-based \acrshort{cla} has also been studied in encoder-decoder machine translation. \citet{kudugunta-etal-2019-investigating} and \citet{qu-etal-2025-languages} used the similarities of the embeddings to show that when the target language is known, the encoder will map the input sentence into that language's space (as opposed to a language-independent space). Compared to that, the sentence embeddings obtained from an \acrshort{llm} are completely agnostic to the target language or the translation task as a whole, which considerably changes the predictive properties of the alignment. 

\paragraph{Obtaining text embeddings.}
Embedding-level alignment scores (described below) require a~fixed-length vector representation for each sentence.\punctfootnote{While we work with sentences in the~traditional linguistic sense, any reasonably long chunk of text can be considered a~sentence in this context.} In contrast, \acrshort{llm}s assign a~hidden state (embedding) to individual tokens. Various approaches for pooling token embeddings into a~single vector exist; we refer to these as \emph{sentence representations}.

Taking the \textbf{mean} of the~embeddings is a~straightforward option; however, \citet{jiang-etal-2022-promptbert} found that this approach with BERT \citep{devlin-etal-2019-bert} performs worse than averaging static embeddings for semantic similarity tasks.

In decoder-only Transformers \citep{transformers}, attention is masked so that each token can attend only to previous tokens, meaning later tokens carry more information. \Acrlong{pwm} (\textbf{\acrshort{pwm}}; \citealp{muennighoff2022sgptgptsentenceembeddings}) leverages this by computing a~weighted mean of the~embeddings with weights proportional to position index. Alternatively, one can simply use the \textbf{last} token embedding.

To better aggregate sentence meaning, \citeauthor{jiang-etal-2024-scaling} (\citeyear{jiang-etal-2022-promptbert}, \citeyear{jiang-etal-2024-scaling}) propose the \textbf{prompt} \colorbox[RGB]{230,230,230}{This sentence: "\texttt{[text]}" means in one word: "}. However, this approach would require a~human translation of the~prompt into every language considered, which is difficult to obtain; otherwise, the~embeddings would implicitly reflect alignment between English and the~inspected language.

For completeness, we use all sentence representations listed here in our experiments, including the~English \textbf{prompt}, and extend this list with our novel approach in Section~\ref{sec:experiments-cla}.

\paragraph{Embedding-level alignment scores.}

Consider a~parallel dataset of~\(D\) sentences in the~source and target languages, represented as \(d\)-dimensional vectors. Suppose the~embeddings are stored in matrices \(S \in \mathbb{R}^{d\times D}\) (source) and \(T \in \mathbb{R}^{d\times D}\) (target). We consider the~following metrics.

The most straightforward approach, average \textbf{cos}ine similarity, is the~per-sentence-pair average of
\begin{equation}
\cos(S_{*i},T_{*i}) = \frac{S_{*i}^{}T_{*i}^\top}{\lVert S_{*i} \rVert \lVert T_{*i} \rVert},
\end{equation}
However, \citet{ethayarajh-2019-contextual} noted that embeddings occupy an anisotropic space, making direct comparisons unreliable.

\emph{xSIM} \citep{artetxe-schwenk-2019-margin-xsim} uses cosine similarity indirectly by computing accuracy in parallel corpus retrieval. In its simplest form \(\textbf{abs}\), the~retrieval similarity between two embeddings is their cosine similarity \(\cos(S_{*i}, T_{*i})\).
To mitigate scale inconsistencies, the \textbf{dist} and \textbf{ratio} variants use a~margin between the~cosine similarity of \(S_{*i}\) and \(T_{*i}\) and the~average cosine similarity of their $k$ nearest neighbours within their respective languages:
\begin{multline}
\mathrm{margin}\Bigl(
  \cos(S_{*i},T_{*i}), \\
  \frac{1}{2k}
  \Bigl(
    \sum\nolimits_{S_{*j}\in \mathrm{kNN}(S_{*i})}\cos(S_{*i},S_{*j}) \\
    + \sum\nolimits_{T_{*j}\in \mathrm{kNN}(T_{*i})}\cos(T_{*i},T_{*j})
  \Bigr)
\Bigr)
\end{multline}
where margin \(\mathrm{margin}(a,b)\) is either~\(a-b\) (for \textbf{dist}) or~\(a/b\) (for \textbf{ratio}).

\textbf{Dali}\textsubscript{weak} \citep{ravisankar2026mapenglishrolecrosslingual-dali} computes retrieval accuracy on a~corpus where each sample consists of a~context~\(c\) and completions \(a_1, \dots, a_n\), of which exactly one is correct (\(a^\ast\)). Retrieval is successful if the~correct source-language context-completion pair has greater cosine similarity to its target-language translation than to any incorrect target-language completion:
\begin{equation}
\argmax_{a^\mathrm{tgt}} \bigl(\cos(c^\mathrm{src}a^{\ast\mathrm{src}},c^\mathrm{tgt}a^{\mathrm{tgt}})\bigr) = a^{*\mathrm{tgt}}.
\end{equation}
In our experiments, we use Belebele \citep{bandarkar-etal-2024-belebele} as the~dataset of contexts and completions.

Finally, \acrlong{anc} (\textbf{ANC}; \citealp{del-fishel-2022-cross-anc}) leverages the~other dimension of the~\(S\) and \(T\) matrices; for each position (\enquote{neuron}) \(j \in \{1, \dots, d\}\) in the~embedding vector, calculates the~Pearson correlation coefficient over the~\(j\)-th row of the~dataset \(\rho(S_{j*}, T_{j*})\) and returns the~per-neuron average, rather than the~working with the~individual sentence embeddings stored in the~columns.

We test all the~listed methods in our experiments.

\paragraph{Tokeniser-level alignment scores.}

Vocabulary overlap between languages has long been considered a~factor in cross-lingual transfer \citep[e.g.,][]{wu-dredze-2019-beto}.
Rather than simply counting shared tokens, \citet{limisiewicz-etal-2023-tokenization} use the~similarity of two languages' token distributions to predict downstream transfer performance on certain tasks.
However, very few tokens are shared across script boundaries. 

\citet{hammerl-etal-2025-beyond} show that subword token \textit{alignability} is more predictive of downstream transfer than literal subword token distribution overlap.
They define token alignability via the~log-likelihood score of the \textbf{Eflomal} word aligner \citep{ostling2016efficient}, averaged over all sentence pairs in a~parallel corpus.
The~eflomal score acts as a~proxy for how well the~tokeniser supports learning shared semantics across languages, and operates quite differently from embedding-level alignment scores.

\paragraph{English as a~pivot language.}

Contemporary \acrshort{llms} are trained predominantly on English data \citep{openai2024gpt4technicalreport}, yet perform well in other languages. \citet{wendler-etal-2024-llamas-think} investigated this by inspecting the~hidden layers of Llama 2 \citep{touvron2023llama2openfoundation}, whose pre-training data is 89.7\% English. They prompted the~model to translate a~word between two non-English languages, decoded the~intermediate hidden states, and found that the~resulting tokens were often the~English equivalent of the~translated word, suggesting that the~model operates internally in English even when prompted in another language. We provide additional evidence for this claim from a~different perspective using novel methodology.

\citet{kargaran-etal-2025-mexa} and \citet{ravisankar2026mapenglishrolecrosslingual-dali} build on this claim, showing that retrieval-based \acrshort{cla} between a~given language and English correlates well with monolingual performance in that language on tasks such as reading comprehension, language understanding, and narrative understanding.

Explicitly using a high-resource language, such as English, as a pivot has been a common practice in machine translation. The techniques include translating into the pivot language as an intermediate step, generating pivot-based synthetic data, or transferring a model trained on the pivot language \citep{kim-etal-2019-pivot}. Evidence presented above suggests that this approach emerges in \acrshort{llms} as well. We aim to further study this behaviour using \acrshort{cla}.

\begin{figure}[h]
    \centering
    \includegraphics[width=\columnwidth]{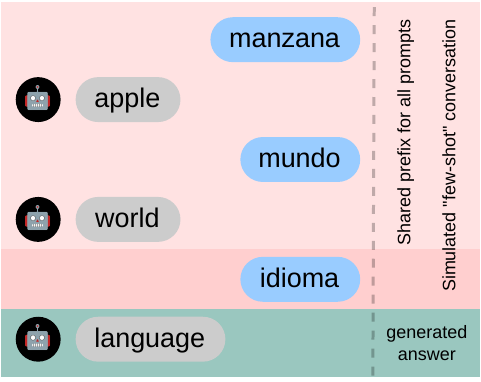}
    \caption{Example two-shot prompting conversation for translation from Spanish to English.}
    \label{fig:prompting}
\end{figure}

\paragraph{Few-shot prompting.}
Few-shot prompting involves presenting an \acrshort{llm} with multiple solved examples of a task, followed by an unsolved instance. The model then follows the previous pattern and generates the answer for that instance. This approach is typical for non-instruction-tuned models \citep{lms-are-fewshot}. In some of our experiments, we opt for a few-shot approach and exploit its absence of an explicit instruction, since translating a prompt into all languages would be unfeasible, and using a shared English prompt would implicitly embed the alignment with English.

We further expand the few-shot learning approach to instruction-tuned \acrshort{llms}, where we apply a chat template to the traditional few-shot prompt to simulate a conversation without an explicit instruction. Figure~\ref{fig:prompting} shows an example few-shot prompt for translation.

\section{Experimental Setup}\label{sec:setup}

In our experiments, we use five methods to represent sentences as fixed-length vectors and six methods to measure \acrshort{cla} scores from those representations, together with a~tokeniser-level \acrshort{cla} metric, yielding 27 feasible combinations in total. We examine how these combinations correlate with performance on downstream tasks.

We conduct our experiments with the~following models in their base and instruction-tuned versions: \texttt{Qwen3-14B[-Base]} \citep{yang2025qwen3technicalreport}, \texttt{gemma-3-12b-\-\{pt,it\}} \citep{gemmateam2025gemma3technicalreport}, and \texttt{Ministral-3-\-14B-\{Base,Instruct\}-\-2512} \citep{liu2026ministral3}. To ensure a fair comparison, we disable the chain-of-thought phase for all models. In all of the experiments, we work with 44~languages of various families, resource-richnesses, and writing systems, listed in Appendix~\ref{app:langs}. Throughout, we use the~Pearson correlation coefficient as the~measure of correlation.

\subsection{Computing CLA Scores}\label{sec:experiments-cla}

We compute all alignments on the~44-language subset of the \emph{Flores-200}-\texttt{dev} dataset (\citealp{goyal-etal-2022-flores}; 997 samples).
For the~embedding-level \acrshort{cla} scores, we use the~sentence representation approaches described in Section~\ref{sec:background}: \textbf{mean}, \textbf{\acrshort{pwm}} (\acrlong{pwm}; \citealp{muennighoff2022sgptgptsentenceembeddings}), \textbf{last} token, and English \textbf{prompt} \citep{jiang-etal-2024-scaling}.

Additionally, \emph{we propose a~new embedding extraction method:} \textbf{fewshot}, which guides the \acrshort{llm} to concentrate the~meaning of a~sentence into a~single token.
This is similar to the \textbf{prompt} method \citep{jiang-etal-2024-scaling}, but \textbf{fewshot} relies neither on an English prompt -- which carries implicit information about alignment with English -- nor on translating the~prompt into each language.

The \textbf{fewshot} approach is inspired by \emph{DefSent} \citep{tsukagoshi-etal-2021-defsent}, who fine-tuned a BERT model to predict a masked word following its definition. Then, they used the embeddings of the \texttt{[MASK]} token preceded by a sentence to embed. Unlike in DefSent, we do not fine-tune the model; we prompt it with a few-shot prompt to predict an entity from its description. We use the~sentence to be embedded as the~last description and extract embeddings from the~token preceding the~predicted entity. As the source of our few-shot data, for each language, we take the~first five Wikidata entities \citep{vrandevcic2014wikidata}, sorted by QID, with descriptions of at least 10 characters. Listing~\ref{lst:fewshot-prompt} in Appendix~\ref{app:prompts} shows an example English prompt.

Since all alignment scores are calculated per layer, we select the~layer with the \emph{maximum} alignment score, following \citet{kargaran-etal-2025-mexa}.
Next, we pair the~sentence representation approaches with the \acrshort{cla} metrics described in Section~\ref{sec:background}: \textbf{cos}ine similarity, \emph{xSIM} scores (\textbf{abs}, \textbf{dist}, \textbf{ratio}; \citealp{artetxe-schwenk-2019-margin-xsim}), \textbf{Dali}\footnote{We only pair \textbf{Dali} with the \textbf{last} token sentence representation, following \citet{ravisankar2026mapenglishrolecrosslingual-dali}. Moreover, \textbf{Dali} scores are estimated on the Belebele dataset, instead of Flores.} \citep{ravisankar2026mapenglishrolecrosslingual-dali}, and \textbf{ANC} \citep{del-fishel-2022-cross-anc}.

For tokeniser-level \acrshort{cla} scores, we follow \citet{hammerl-etal-2025-beyond} and use the \textbf{Eflo}mal score \citep{ostling2016efficient} to estimate token alignability in the~parallel corpora. Unlike \citet{hammerl-etal-2025-beyond}, we do \emph{not} estimate the~priors on a~larger dataset -- due to data scarcity -- and instead compute the~score directly from \emph{Flores-200}-\texttt{dev}. Since a~lower \textbf{Eflo}mal score indicates better alignability, we negate it to obtain positive expected correlations with task performance.

\paragraph{Correlations within the alignment scores.\hfill{}}\label{sec:corr-within-cla}
The~alignment of a~given language to English is strongly correlated with its alignment to other non-English languages.
To demonstrate this, we correlate each \texttt{src} language's alignment to English with its average alignment to all non-English languages (\texttt{src-[\textasciitilde{}en]}) in Table~\ref{tab:corr-Qwen3-14B} (for \texttt{Qwen3}).
This finding implies that alignments cannot be viewed as complementary in an ensemble combining multiple alignment directions (such as Equation~\ref{eq:big-ansamble}): even if both \texttt{src-en} and \texttt{src-tgt} \acrshort{cla}s correlate with translation quality, their combination may not yield a~higher correlation. This means that it is not sufficient to observe the correlations individually; instead, we focus on how much one alignment direction can improve others in the translation task (Section~\ref{sec:translation}).

\input{latex/tables/corr_with_cla_en_Qwen3-14B}

\section{SIB-200 Benchmark}

The~task in \emph{SIB-200} \citep{adelani-etal-2024-sib} is to categorise a~text segment into one of seven topics. As the~category is in English, we use an English prompt, stated in Listings~\ref{lst:sib-200-prompt} (base models) and~\ref{lst:sib-200-prompt-instruction} (instruction-tuned models) in Appendix~\ref{app:prompts}. We compute the~macro‑averaged \(F_1\)~score on the~\texttt{train} split (701 samples) and correlate it with the \texttt{src-en} \acrshort{cla} score across each \texttt{src} language, reporting the~per‑language correlation for each \acrshort{cla} metric and representation.

\subsection{Results and Discussion}

\begin{figure}[t]
    \centering
    \includegraphics[width=\columnwidth,trim= 0 0 0 20,clip]{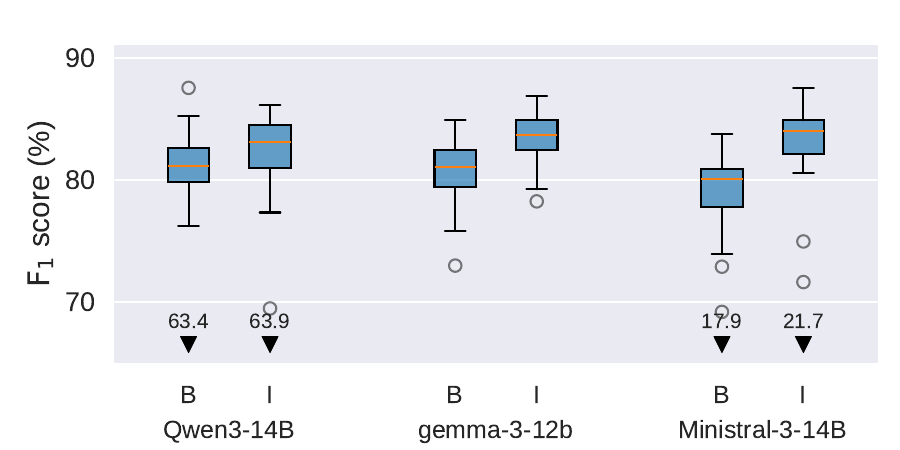}
    \caption{Macro \(F_1\)~score distribution across languages for \textbf{SIB-200}. Cropped outliers are marked by \(\blacktriangledown\).}
    \label{fig:performanceTasksib-200}
\end{figure}

Overall, the~models achieved satisfactory performance on SIB-200, as shown in Figure~\ref{fig:performanceTasksib-200}. \texttt{Ministral-3} achieved the~lowest score on Amharic, indicating its set of supported languages is more limited than the~other models.

As Table~\ref{tab:overview-corr-en-task-sib-200} shows, \texttt{src-en} \acrshort{cla} metrics correlate moderately to strongly with performance, with \textbf{prompt} repeatedly appearing as the~best-performing sentence representation. This is likely because, in this task, we prompt the models in English, and the \textbf{prompt} representation implicitly encodes understanding of English prompts. We show detailed results in Table~\ref{tab:corr-en-task-sib-200-all-HIGHEST} in Appendix~\ref{app:results}.

\input{latex/tables/overview_corr_with_sib-200_en}

\section{Belebele Benchmark}

\begin{figure}[t]
    \centering
    \includegraphics[width=\columnwidth,trim= 0 0 0 20,clip]{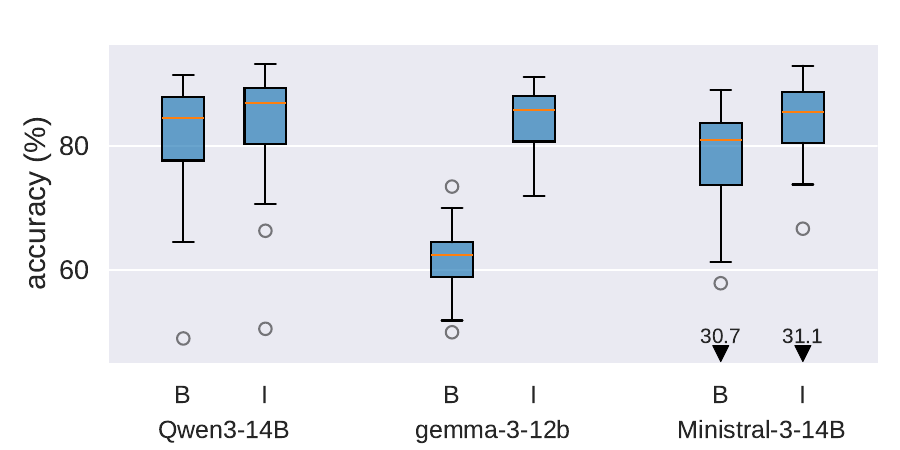}
    \caption{Accuracy distribution across languages for \textbf{Belebele}. Cropped outliers are marked by \(\blacktriangledown\).}
    \label{fig:performanceTaskBelebele}
\end{figure}

\emph{Belebele} (\citealp{bandarkar-etal-2024-belebele}; 900 samples) is a~multiple-choice reading comprehension dataset where the~task is to select the~correct answer from four options given a~context and a~question. Each part (context, question, options) is translated into each supported language, yielding a~fully monolingual task in each language (unlike SIB-200). To avoid using English in the~prompt, we use a~few-shot prompting approach that shows three answered examples before the~sample to be solved.

The~prompting template adds only newline characters to separate the~parts of the~text (or role switching in the case of instruction-tuned model). We do \emph{not} introduce option numbering or field names (such as \enquote{Context:}). The \acrshort{llm} answers by explicitly stating one of the~options, without sampling during generation. To save resources, we limit generated tokens to the~smallest prefix that uniquely distinguishes each answer and check whether the~generated prefix matches that of the~correct answer. Listings~\ref{lst:belebele-prompt} and~\ref{lst:belebele-prompt-instruct} in Appendix~\ref{app:prompts} show an example prompt for base and instruction-tuned models, respectively.

We measure prediction accuracy for each language and inspect the~correlation between accuracy and the \texttt{src-en} \acrshort{cla} score over each \texttt{src} language.

\subsection{Results and Discussion}

Figure~\ref{fig:performanceTaskBelebele} shows that the~models  again achieved solid results, with multiple models exceeding 80\% accuracy on average. As with SIB-200, \texttt{Ministral-3} achieved the~lowest score on Amharic.

As Table~\ref{tab:overview-corr-en-task-belebele} shows, the~correlation between \texttt{src-en} \acrshort{cla} metrics and Belebele performance is moderate to strong, with \textbf{p-wm} repeatedly appearing as the~optimal representation and \textbf{ANC} as the~optimal \acrshort{cla} metric. Detailed correlations per model are in Table~\ref{tab:corr-en-task-belebele-all-HIGHEST} in Appendix~\ref{app:results}.

\input{latex/tables/overview_corr_with_belebele_en}

\section{Translation Benchmark}\label{sec:translation}

Unlike the~previous monolingual tasks, translation involves two languages: \texttt{src} and \texttt{tgt}. A~natural question is whether the \texttt{src-tgt} \acrshort{cla} score predicts translation quality well. Alternatively, if \acrshort{llm}s use English as a~pivot language, \texttt{src-en} and \texttt{en-tgt} \acrshort{cla} may correlate more strongly with translation quality.

We primarily use the~\texttt{devtest} split of \emph{Flores-200} for translation (1011~samples) and additionally the \texttt{dev} split of the \emph{BOUQuET} dataset \citep{andrews-etal-2025-bouquet} for out-of-domain validation. We consider all the possible pairs of the 44 languages in both directions. To avoid English prompts that might bias alignment measurements, we use a~few-shot prompt of five examples, followed by the~sentence to translate. An example is in Listing~\ref{lst:translation-prompt}, Appendix~\ref{app:prompts}. We generate translations without sampling, limiting the~number of tokens to 1.5 times the~length of the~reference.

We evaluate translations using the \textsc{chrF} score\footnote{Using default parameters: \texttt{ncorder=6}, \texttt{nworder=2}, \(\beta\)\texttt{=2}.} \citep{popovic-2015-chrf} at the~system level for each language pair. A~limitation of \textsc{chrF} is that its value depends strongly on the~target language's writing system, as characters carry different amounts of information across writing systems.

This restricts us to measuring correlations within a~fixed target language and averaging across target languages. Thus, we can assess the performance of \texttt{src-tgt} and \texttt{src-en} \acrshort{cla} scores, but assessing \texttt{en-tgt} is impossible in this setting, as it is constant under a~fixed target language. We tackle this problem in the~following section.

\subsection{PMI as a~Translation Quality Metric}\label{sec:translation-pmi}

To overcome the~target-language sensitivity of \textsc{chrF}, we propose a~task-specific machine translation metric: \acrfull{pmi} between the~source sentence and its translation.

We aim to reduce target-language dependence by normalising the~likelihood of a~translation given the~source sentence (\(p(y|x)\)) by its unconditional (prior) likelihood (\(p(y)\)) and compute the~logarithm of the~value:
\begin{equation}\label{eq:pmi}
    \log\frac{p(y|x)}{p(y)} = \log\frac{p(x,y)}{p(x)p(y)} = \mathrm{PMI}(x,y)
\end{equation}

We compute the~conditional log-likelihood using the~same prompt as for generating translations. For the~prior log-likelihood, we use a~prompt with the~same five sentences but only in the~target language to introduce the~domain and language. An example is in Listing~\ref{lst:translation-prior-prompt}, Appendix~\ref{app:prompts}. In both cases, we concatenate the~few-shot prefix (\(m\)~tokens) and the~translation (\(n\)~tokens) into a~single string~\(s\) and compute the~length-normalised log-likelihood of the~translation:
\begin{equation}
    \log p(\cdot)=\frac{1}{n}\sum_{i=m+1}^{m+n}\log p(s_i | s_1, \dots,s_{i-1})
\end{equation}

We validate this metric via its strong Pearson correlation with \textsc{chrF}, measured within a~fixed target language, shown in Table~\ref{tab:chrf-pmi-corr}. The correlation is particularly strong for base models. A notable outlier is \texttt{gemma-3} instr. on the Flores dataset. A possible explanation is training data contamination, which could skew the prior distribution.

Intuitively, \acrshort{pmi} should achieve target-language independence, as all the~language-specific factors that influence the~conditional probability of the~translation are also included in the~prior probability, and the~metric measures the~relative increase in probability between the~two. We empirically discuss target-language independence in Appendix~\ref{sec:target-indep}.

\input{latex/tables/chrf_pmi_corr}

Note that we propose \acrshort{pmi} for our specific experimental setup, rather than 
as
a~translation evaluation metric for general use, as its value is unbounded and may vary across models. Nonetheless, it may be used in any experiments in which the exact value is not important, only the correlation within a fixed model.

\subsection{\texorpdfstring{Analysis with \textsc{chrF}}{Correlation Analysis with chrF}}\label{sec:chrf-analysis}

Table~\ref{tab:translation-performance-by-language-short} shows \textsc{chrF} scores for selected target languages; the~complete lists are in Tables~\ref{tab:translation-performance-by-language} (Flores) and~\ref{tab:translation-performance-by-language-bouquet} (BOUQuET), Appendix~\ref{app:results}.
For all target languages, \texttt{gemma-3} achieved the~highest results.
The scores for \texttt{Ministral-3}-instruct are low because it frequently failed to follow the few-shot template and provide a translation. 
As noted above, performance across target languages is not directly comparable: e.g., \textsc{chrF} is lower for Chinese than for other high-resource languages, because Chinese characters carry more information per character.

\input{latex/tables/translation_performance_by_language-short}
\begin{figure}
    \centering
    \includegraphics[width=\columnwidth,trim= 0 10 0 10,clip]{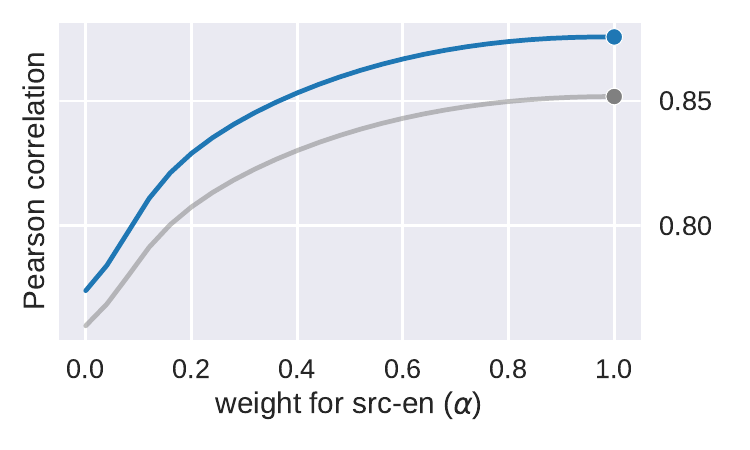}
    \caption{Per-target-language average Pearson correlation between translation \textsc{chrF} and the \(\alpha\)-weighted convex combination of \texttt{src-en} \texttt{src-tgt} \acrshort{cla} scores for \textbf{Ministral-3 base}, \textbf{ANC}, \textbf{mean} representations. \colorbox[RGB]{155,170,255}{\textbf{Blue}} plot: Flores, \colorbox[RGB]{200,200,200}{\textbf{grey}} plot: BOUQuET.}
    \label{fig:alpha-vs-corr}
\end{figure}

Correlations with \texttt{src-en} and \texttt{src-tgt} are shown in Table~\ref{tab:overview-all-corr-combined-task-chrf} top and middle (overview) and Table~\ref{tab:corr-both-with-chrf-all-HIGHEST}, Appendix~\ref{app:results} (per model).
The \textbf{ANC} metric repeatedly achieves the~best results across models. There is no dominant sentence representation that wins consistently across all models. In most cases, \texttt{src-en} \acrshort{cla} correlates more strongly with \textsc{chrF} than \texttt{src-tgt} does. 
Given the~correlations between \acrshort{cla} metrics described in Section~\ref{sec:corr-within-cla}, we next ask whether \texttt{src-tgt} carries any association beyond that contained in \texttt{src-en}.

To assess this, we measure the~improvement in correlation from a~convex combination of the~two. For each sentence representation and \acrshort{cla} metric pair, we find the~optimal convex combination weight~\(\alpha\), maximising the~correlation of
\begin{equation}
    \alpha\mathrm{\acrshort{cla}}_{\mathtt{src-en}} + (1-\alpha)\mathrm{\acrshort{cla}}_{\mathtt{src-tgt}}
\end{equation}
with \textsc{chrF} averaged across target languages. We search over $[0,1]$ with step size $1/25$.

The~lower part of Table~\ref{tab:overview-all-corr-combined-task-chrf} shows the~correlations with this~optimal convex combination. 
Detailed results are in Tables~\ref{tab:corr-combination-with-chrf-combined} and~\ref{tab:corr-combination-with-chrf-all-HIGHEST-validation}, Appendix~\ref{app:results}. Although both \texttt{src-en} and \texttt{src-tgt} correlate well with \textsc{chrF}, combining them increases the~correlation only slightly.  
Figure~\ref{fig:alpha-vs-corr} illustrates the~relationship between~\(\alpha\) and the~correlation for Ministral-3 base, the~model with the~strongest overall correlation. 
The BOUQuET validation shows that the correlation is retained even when the alignment and translation datasets come from different domains.

\input{latex/tables/table_wide_corr_chrf}

\subsection{Analysis with PMI}\label{sec:corr-analysis-pmi}

Unlike with \textsc{chrF}, we do not fix the~target language when analysing \acrshort{pmi}. Instead, we compare all language pairs simultaneously.
In this setting, using only one alignment from \texttt{src-en} or \texttt{en-tgt} is impractical, as it would yield duplicate alignment scores.\punctfootnote{For example, if we used only \texttt{src-en} \acrshort{cla} scores, the~pairs \texttt{it-ar} and \texttt{it-pt} would both be assigned the~same \texttt{it-en} \acrshort{cla} score.}
Therefore, we omit the~single-alignment experiment and proceed directly to analysing the~convex combination correlation. This time, we have three weights, with two degrees of freedom: \(\alpha \in [0,1], \beta \in [0, 1-\alpha]\):
\begin{multline}\label{eq:big-ansamble}
    \alpha\mathrm{\acrshort{cla}}_{\mathtt{src-en}} + \beta\mathrm{\acrshort{cla}}_{\mathtt{en-tgt}} \\
    +  (1-\alpha-\beta)\mathrm{\acrshort{cla}}_{\mathtt{src-tgt}}
\end{multline}

\input{latex/tables/corr_pmi_overview_combined}

\begin{figure*}
    \centering
    \includegraphics[width=\textwidth,trim= 0 10 0 30,clip]{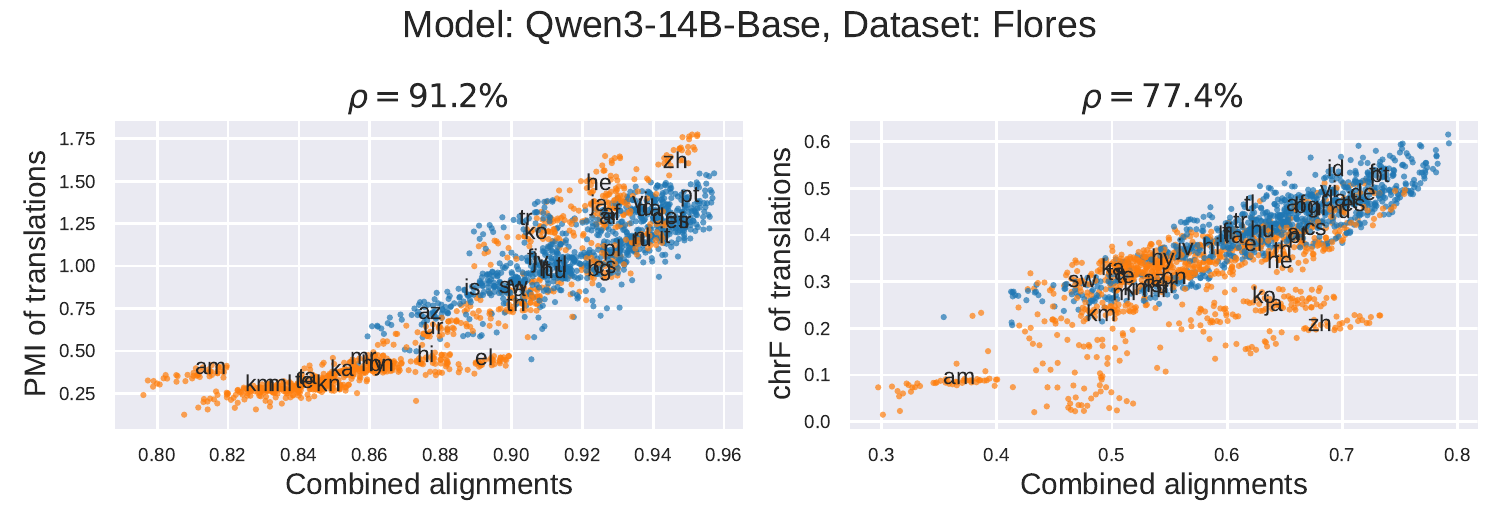}
    \caption{\textsc{chrF} (right) clusters target languages by character information density, while clusters are considerably less pronounced for \acrshort{pmi} (left). Each point is a~language pair; non-Latin-script target languages are in orange. Both metrics are correlated with their respective optimal convex combination of \texttt{src-en}, \texttt{en-tgt}, and \texttt{src-tgt} \acrshort{cla} scores, using the~best-correlating sentence representation and \acrshort{cla} metric. Model: \texttt{Qwen3-Base}.
    }
    \label{fig:pmi-vs-chrf}
\end{figure*}

\begin{figure}[t]
    \centering
    \includegraphics[width=\columnwidth]{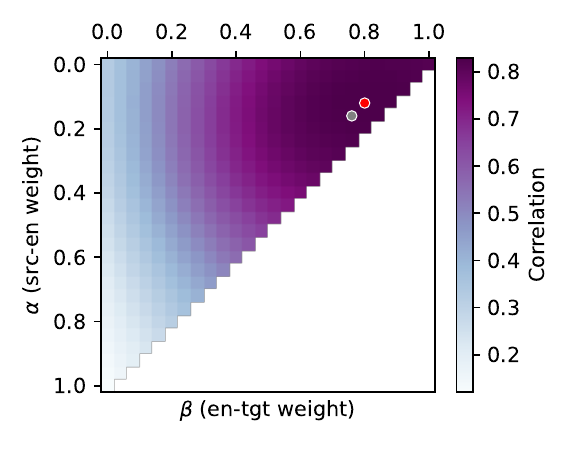}
    \caption{Correlation heatmap for the~$\alpha$ and~$\beta$ grid search for the~best-performing sentence representation/\acrshort{cla} metric pair for \textbf{\texttt{Qwen3} instr.} when predicting \textbf{Flores} translation PMI. Values on the~diagonal correspond to zero \texttt{src-tgt} weight. The~best correlation is marked red. The best combination for \textbf{BOUQuET} is marked grey.}
    \label{fig:corr-heatmap}
\end{figure}

We show the~overview of the~results of this experiment in Table~\ref{tab:corrPmiOverviewCombined} (top) and the~details in Tables~\ref{tab:corr-pmi-combined}, \ref{gammas-pmi-all-HIGHEST-pmi} and~\ref{tab:validation-pmi-all-HIGHEST-pmi}, Appendix~\ref{app:results}. Figure~\ref{fig:corr-heatmap} illustrates the~grid search for~$\alpha$ and~$\beta$ as a~heatmap. The~results indicate moderate to strong correlations, also retained on the validation data. The \textbf{p-wm} sentence representation consistently performs the~best, while \textbf{cos} appears to be the~optimal metric for multiple models. Importantly, the~weight of \acrshort{cla}\textsubscript{\texttt{src-tgt}} in the~optimal convex combination is consistently near-zero or zero, suggesting that \textbf{most information about the~translation quality is contained in the \acrshort{cla}\textsubscript{\texttt{src-en}} and the \acrshort{cla}\textsubscript{\texttt{en-tgt}} scores}.

Additionally, the~results we show in Figure~\ref{fig:pmi-vs-chrf} also confirm our previous design decisions about \acrshort{pmi}, showing that it can be used as a~target-language independent metric (discussion in Appendix~\ref{sec:target-indep}).

\subsection{What About Related Languages?}\label{sec:related-langs}

The~findings from the~previous section show that the~alignment \texttt{src-tgt} has only little relevance for translation quality, implying that \acrshort{llm}s might internally translate first to English and then to the~target language, as \citet{wendler-etal-2024-llamas-think} suggested. However, in a~practical setting, translating via English makes little sense when the~source and target languages are closely related.

To examine this specific case, we create a~new set of seven pairs of mutually intelligible languages, listed in Appendix~\ref{app:langs}. Considering both directions, we work with 14 data points. We conduct the~same experiments on these language pairs as in the~previous section. Table~\ref{tab:corrPmiOverviewCombined} (bottom) shows vastly different results than before: the~weight for the \texttt{src-tgt} \acrshort{cla} score is now dominant (\(\ge 84\%\)), suggesting that the \acrshort{llm}s may not use English as a~pivot point when direct translation is trivial.

\section{CLA Metrics Evaluation}\label{sec:cla-metrics-eval}

The~results across all three tasks allow us to make several recommendations regarding CLA metric choice.
While there is no one consistently best-performing approach to measuring \acrlong{cla}, the \acrlong{pwm} (\textbf{\acrshort{pwm}}; \citealp{muennighoff2022sgptgptsentenceembeddings}) sentence representation wins in more settings than any other representation. 
Similarly, there was no single preferred alignment scoring method, but \acrlong{anc} (\textbf{\acrshort{anc}}; \citealp{del-fishel-2022-cross-anc}) wins in more settings than any other metric.

Therefore, we suggest using \textbf{\acrshort{pwm}} in combination with \textbf{\acrshort{anc}} in future experiments. However, in cases when sample-level alignment is needed, \acrshort{anc} is not applicable, and instead, we recommend using a~retrieval-based metric (the~simplest case is xSIM \textbf{abs}; \citealp{artetxe-schwenk-2019-margin-xsim}).

Compared to the~embedding-level \acrshort{cla} scores, the~tokeniser-level \textbf{Eflo}mal score \citep{hammerl-etal-2025-beyond} achieves considerably worse correlation scores, except for the~similar languages task. None\-theless, in cases where the~model's hidden states are unavailable (such as in closed-weight models), it can serve as a~rough estimate of the~model's expected performance.

\section{Conclusions}

We comprehensively review 27 approaches to measuring \acrlong{cla} from sentence embeddings by combining five sentence representation methods -- one of which, \textbf{fewshot}, we newly propose -- and six alignment scoring methods, together with one tokeniser-level metric. Our newly proposed method outperforms existing methods in numerous setups, showing potential for further development.
We work with 44 languages of varying linguistic properties and three \acrshort{llm} families.

We assess how predictive \acrshort{cla} scores are for performance on three common NLP tasks: topic classification (SIB-200), reading comprehension (Belebele), and non-English-to-non-English machine translation. 
For monolingual tasks, we show that the~performance in the \texttt{src} language is well correlated with the \texttt{src-en} \acrshort{cla} score, which is consistent with the~findings from previous work \citep{kargaran-etal-2025-mexa, ravisankar2026mapenglishrolecrosslingual-dali}.

In the~machine translation task, we address whether alignment between the~source and target languages (\texttt{src-tgt}) is more predictive of translation quality than alignment via an intermediate English step (\texttt{src-en} and \texttt{en-tgt}). First, we measure translation quality using the \textsc{chrF} score, which only allows us to analyse \texttt{src-tgt} and \texttt{src-en}. We find that \texttt{src-en} is generally more predictive of quality and only improves slightly by combining it with the \texttt{src-tgt} score.

Next, to include the \texttt{en-tgt} score in our analysis, we propose \acrfull{pmi} for translation---a~proxy for \textsc{chrF}---which we show to be less dependent on the~target language. Using this method, we show that the \texttt{src-tgt} score does not further improve the~predictions of the \texttt{src-en} and \texttt{en-tgt} combined. This contributes evidence to the~claim that \acrshort{llm}s use English as a~pivot language \citep{wendler-etal-2024-llamas-think} from a~yet unstudied perspective.

We summarise the~predictive performance of different \acrshort{cla} scores into three recommendations for best performance estimates:
\begin{enumerate*}[label=(\arabic*)]
    \item Use \textbf{\acrshort{pwm}}+\textbf{\acrshort{anc}} for system-level tasks scoring.
    \item Use \textbf{\acrshort{pwm}}+\textbf{abs} when sample-level scores are needed.
    \item Use \textbf{Eflo}mal only when model weights are unavailable. 
\end{enumerate*}

\section*{Limitations}

\paragraph{Domain coverage.}

Highly multilingual parallel data remains scarce, and the~Flores-200 dataset, while the~best available resource for our purposes, covers a~single domain.
Consequently, our \acrshort{cla} scores and two downstream tasks share the~same domain, which might inflate the~observed correlations compared to what one would expect in more diverse settings.
Controlling for domain effects would require multilingual parallel corpora of comparable quality across multiple domains. Such a dataset (BOUQuET) exists for translation, which we use for validation. However, to our knowledge, there is no suitable alternative to the SIB-200 and Belebele dataset that covers a different domain with the same level of multilinguality.

\paragraph{The~few-shot sentence representation.}

Our proposed \textbf{fewshot} embedding method relies on Wikidata entity descriptions, which vary in quality and may not describe the~same entities across all 44 languages.
A~more principled approach would use verified multilingual dictionary definitions, but no such resource exists with the~language coverage we require. Nonetheless, our proposed method remained the best-performing for certain setups, highlighting its potential.

\paragraph{Target-language independence of PMI.}

While we show empirically that our proposed \acrshort{pmi} metric clusters target languages less severely than \textsc{chrF}, full independence cannot be formally established.
The~metric remains sensitive to model-specific calibration, and the~degree of residual dependence likely varies across models.
We view this as an invitation for future work rather than a~deficiency of the~current proposal.

\paragraph{Related languages experiment.}

Our side experiment with related languages (Section~\ref{sec:related-langs}) covers only 7 pairs of related languages and is evaluated only on base models. Furthermore, token overlap between related languages may skew the PMI and \textsc{chrF} distributions. We invite future work to examine the role of English as a pivot language in more detail when translating between related languages.

\paragraph{Language coverage.}

We selected the set of 44 languages we work with to cover various families,
resource-richnesses, and writing systems. 17 of those languages (37\%) are European, which is not representative of the global population. Nonetheless, our experiments (Figure~\ref{fig:pmi-vs-chrf}) show that the script and the resource-richness of the language affect the results more than the language's geographical origin.

\section*{Ethical Considerations}
We do not anticipate any negative ethical implications arising from this study. We used AI-assisted coding (GitHub Copilot) and AI tools (Claude, ChatGPT, and Grammarly) for grammar checking and minor sentence rephrasing.

The inspected open-weight models are each licensed under the \texttt{Apache 2.0} license (\texttt{Qwen3} and \texttt{Ministral-3}) and the \texttt{gemma} license (\texttt{gemma-3}). The BOUQuET dataset is licensed under \texttt{CC-BY-4.0}, while the remaining datasets we work with are licensed under \texttt{CC-BY-SA-4.0}.

\section*{Acknowledgments}
Adnan was supported by the HumanAId project \texttt{CZ.02.01.01/00/23\_025/0008691} of the Czech Ministry of Education. Jindřich was supported by the CUNI project \texttt{PRIMUS/23/SCI/023} and project \texttt{CZ.02.01.01/00/23\_020/0008518} of the Czech Ministry of Education. This research was partially supported by SVV~project number \texttt{260 821}.

This work was co-funded by the European Union (ERC, EPICAL, 101141712). Views and opinions expressed are however those of the author(s) only and do not necessarily reflect those of the European Union or the European Research Council. Neither the European Union nor the granting authority can be held responsible for them.

\bibliography{latex/custom}

\appendix

\section{Evaluated Languages}\label{app:langs}

In this article, we work with the~following languages: Afrikaans (\texttt{af}), Amharic (\texttt{am}), Arabic (\texttt{ar}),
Azerbaijani (\texttt{az}), Bulgarian (\texttt{bg}), Bengali (\texttt{bn}), 
Czech (\texttt{cs}), Danish (\texttt{da}), German (\texttt{de}), 
Greek (\texttt{el}), English (\texttt{en}), Spanish (\texttt{es}), 
Farsi (\texttt{fa}), Finnish (\texttt{fi}), French (\texttt{fr}), 
Hebrew (\texttt{he}), Hindi (\texttt{hi}), Hungarian (\texttt{hu}),
Armenian (\texttt{hy}), Indonesian (\texttt{id}), Icelandic (\texttt{is}),
Italian (\texttt{it}), Japanese (\texttt{ja}), Javanese (\texttt{jv}), 
Georgian (\texttt{ka}), Central Khmer (\texttt{km}), Kannada (\texttt{kn}),
Korean (\texttt{ko}), Lithuanian (\texttt{lt}), Malayalam (\texttt{ml}),
Marathi (\texttt{mr}), Dutch (\texttt{nl}), Polish (\texttt{pl}), 
Portuguese (\texttt{pt}), Russian (\texttt{ru}), Swahili (\texttt{sw}), 
Tamil (\texttt{ta}), Telugu (\texttt{te}), Thai (\texttt{th}), 
Tagalog (\texttt{tl}), Turkish (\texttt{tr}), Urdu (\texttt{ur}), 
Vietnamese (\texttt{vi}), and Chinese (\texttt{zh}).

In Section~\ref{sec:related-langs} only, we work with a~special set of seven pairs of closely related languages. Note that we use each pair in both directions, yielding 14 data points:
Czech–Slovak (\texttt{cs-sk}),
Spanish–Catalan (\texttt{es-ca}),
Afrikaans–Dutch (\texttt{af-nl}),
Latvian–Lithuanian (\texttt{lv-lt}),
Turkish–Azerbaijani (\texttt{tr-az}),
Danish–Norwegian (\texttt{da-nb}), and
Urdu–Hindi (\texttt{ur-hi}).

\section{Inference and Prompts}\label{app:prompts}

For high-level inference, we use the vLLM inference engine \citep{kwon2023efficient-vllm}. Elsewhere, we use Transformers \citep{wolf-etal-2020-transformers} with code optimisation for key-value caching. We run the~experiments on 1--2~NVIDIA A40 (48~GB) or A100 (40~GB) GPUs, depending on the task.

The~following are example prompts for each task. The~empty red circle symbol (\(\textcolor{red}{\circ{}}\)) denotes an empty line. The~curled arrow (\textcolor{darkgray}{\(\hookrightarrow\)}) denotes text wrapping in the~box, i.e. there is no newline character in the~place of the~arrow.

\lstset{
  inputencoding=utf8,
  keywords={},
  breaklines=true,
  breakatwhitespace=true,
  basicstyle=\ttfamily\footnotesize,
  breakautoindent=false,
  breakindent=0.5em,
  postbreak=\mbox{\textcolor{darkgray}{$\hookrightarrow$}\space},
  captionpos=b,
  frame=single,
  language={},
  literate={☐}{{$\square{}$}}1 {ù}{{\`{u}}}1 {è}{{\`{e}}}1 {à}{{\`{a}}}1 {◦}{{$\textcolor{red}{\circ{}}$}}1 {USER:}{{$\textcolor{blue}{\texttt{USER:}}$}}5 {ASSISTANT:}{{$\textcolor{violet}{\texttt{ASSISTANT:}}$}}{10}
}

\begin{lstlisting}[caption={Example \textbf{few-shot} prompt for embedding the~English sentence \enquote{All human beings are born small.} The~embeddings would be extracted from the~last token, denoted by~\(\square{}\) (the~symbol is not a~part of the~actual prompt).}, label={lst:fewshot-prompt}]
totality consisting of space, time, matter and energy
universe
◦
third planet from the Sun in the Solar System
Earth
◦
matter capable of extracting energy from the environment for replication
life
◦
permanent cessation of vital functions
death
◦
any single member of Homo sapiens, unique extant species of the genus Homo
human
◦
All human beings are born small.
☐
\end{lstlisting}

\begin{lstlisting}[caption={Prompt for extracting the \emph{SIB-200} label. The~English prompt is used for all text segments regardless of the~language.}, label={lst:sib-200-prompt}]
Classify the following text into one of these topics: "science/technology", "travel", "politics", "sports", "health", "entertainment", "geography". Provide only the topic in English as your response.
◦
text: `<text segment>`
topic: `
\end{lstlisting}

\newpage
\null
\vfill{}
\begin{lstlisting}[caption={Example prompt for a \emph{Belebele} sample in English, using one reference shot (we use three in the~experiments). Supposing that each word is tokenised into a~single token, the~model would only need to generate 4 tokens, which suffices to determine the~answer uniquely (\enquote{To allow for subtitles} vs. \enquote{To allow for simple [...]}).}, label={lst:belebele-prompt}]
Make sure your hand is as relaxed as possible while still hitting all the~notes correctly - also try not to make much extraneous motion with your fingers. This way, you will tire yourself out as little as possible. Remember there's no need to hit the keys with a lot of force for extra volume like on the piano. On the accordion, to get extra volume, you use the bellows with more pressure or speed.
◦
According to the passage, what would not be considered an accurate tip for successfully playing the accordion?
◦
For additional volume, increase the force with which you hit the keys
Keep unnecessary movement to a minimum in order to preserve your stamina
Be mindful of hitting the notes while maintaining a relaxed hand
Increase the speed with which you operate the bellows to achieve extra volume
◦
For additional volume, increase the force with which you hit the keys
◦
◦
One of the most common problems when trying to convert a movie to DVD format is the overscan. Most televisions are made in a way to please the general public. For that reason, everything you see on the TV had the borders cut, top, bottom and sides. This is made to ensure that the image covers the whole screen. That is called overscan. Unfortunately, when you make a DVD, it's borders will most likely be cut too, and if the video had subtitles too close to the bottom, they won't be fully shown.
◦
Why do the images on television have their borders cut?
◦
To allow for subtitles
So the image fills the entire screen
To allow for simple conversion into other formats
To cut subtitles too close to the bottom
\end{lstlisting}
\vfill{}
\newpage
\null
\vfill{}

\begin{lstlisting}[caption={Example prompt for a \emph{Belebele} sample in English for instruction-tuned models, using one reference shot (we use three in the~experiments). Supposing that each word is tokenised into a~single token, the~model would only need to generate 4 tokens, which suffices to determine the~answer uniquely (\enquote{To allow for subtitles} vs. \enquote{To allow for simple [...]}). We further use guided decoding that disables generating the \texttt{*} markdown formatting symbol.}, label={lst:belebele-prompt-instruct}]
USER: Make sure your hand is as relaxed as possible while still hitting all the~notes correctly - also try not to make much extraneous motion with your fingers. This way, you will tire yourself out as little as possible. Remember there's no need to hit the keys with a lot of force for extra volume like on the piano. On the accordion, to get extra volume, you use the bellows with more pressure or speed.
◦
According to the passage, what would not be considered an accurate tip for successfully playing the accordion?
◦
For additional volume, increase the force with which you hit the keys
Keep unnecessary movement to a minimum in order to preserve your stamina
Be mindful of hitting the notes while maintaining a relaxed hand
Increase the speed with which you operate the bellows to achieve extra volume
ASSISTANT: For additional volume, increase the force with which you hit the keys
USER: One of the most common problems when trying to convert a movie to DVD format is the overscan. Most televisions are made in a way to please the general public. For that reason, everything you see on the TV had the borders cut, top, bottom and sides. This is made to ensure that the image covers the whole screen. That is called overscan. Unfortunately, when you make a DVD, it's borders will most likely be cut too, and if the video had subtitles too close to the bottom, they won't be fully shown.
◦
Why do the images on television have their borders cut?
◦
To allow for subtitles
So the image fills the entire screen
To allow for simple conversion into other formats
To cut subtitles too close to the bottom
\end{lstlisting}
\vfill{}
\newpage
\null
\vfill{}

\begin{lstlisting}[caption={Example prompt for translating the~English sentence \enquote{\textit{Previously, Ring's CEO, Jamie Siminoff, remarked the~company started when his doorbell wasn't audible from his shop in his garage.}} to Italian.}, label={lst:translation-prompt}]
"We now have 4-month-old mice that are non-diabetic that used to be diabetic," he added.
"Abbiamo topi di quattro mesi che prima erano diabetici e ora non lo sono più", ha aggiunto.
◦
Dr. Ehud Ur, professor of medicine at Dalhousie University in Halifax, Nova Scotia and chair of the clinical and scientific division of the Canadian Diabetes Association cautioned that the research is still in its early days.
Lo studio è ancora in fase iniziale, come dichiarato cautelativamente dal dottor Ehud Ur, professore di medicina alla Dalhousie University di Halifax, Nuova Scozia, e direttore del dipartimento clinico e scientifico della Canadian Diabetes Association.
◦
Like some other experts, he is skeptical about whether diabetes can be cured, noting that these findings have no relevance to people who already have Type 1 diabetes.
Come altri esperti, è scettico circa la possibilità di curare il diabete, sottolineando che questi risultati non hanno rilevanza per chi è già affetto dal diabete di tipo 1.
◦
Previously, Ring's CEO, Jamie Siminoff, remarked the company started when his doorbell wasn't audible from his shop in his garage.
\end{lstlisting}

\begin{lstlisting}[caption={Prompt for extracting the \emph{SIB-200} label for instruction-tuned \acrshort{llms}. The~English prompt is used for all text segments regardless of the~language.}, label={lst:sib-200-prompt-instruction}]
USER: Classify the following text into one of these topics: "science/technology", "travel", "politics", "sports", "health", "entertainment", "geography". Provide only the topic in English as your response.
◦
text: `<text segment>`
ASSISTANT: The topic is: `
\end{lstlisting}
\vfill{}

\vfill{}
\newpage

\begin{lstlisting}[caption={Example prompt for extracting the~prior probability of the~Italian sentence \enquote{\textit{I media locali riferiscono che durante l'intervento un veicolo aeroportuale antincendio si è ribaltato.}}.}, label={lst:translation-prior-prompt}]
"Abbiamo topi di quattro mesi che prima erano diabetici e ora non lo sono più", ha aggiunto.
Lo studio è ancora in fase iniziale, come dichiarato cautelativamente dal dottor Ehud Ur, professore di medicina alla Dalhousie University di Halifax, Nuova Scozia, e direttore del dipartimento clinico e scientifico della Canadian Diabetes Association.
Come altri esperti, è scettico circa la possibilità di curare il diabete, sottolineando che questi risultati non hanno rilevanza per chi è già affetto dal diabete di tipo 1.
I media locali riferiscono che durante l'intervento un veicolo aeroportuale antincendio si è ribaltato.
\end{lstlisting}

\vspace{-1\baselineskip}
\section{Target-Language Independence of PMI}\label{sec:target-indep}
While the~goal of the \acrshort{pmi} translation metric is target-language independence, it is difficult to validate this property, as \acrshort{llm}s a~priori perform better in some languages, making direct comparisons difficult. Nonetheless, we propose that a~strong correlation within mixed target languages (such as that shown in Table~\ref{tab:corrPmiOverviewCombined}, top; for experimental setup see Section~\ref{sec:corr-analysis-pmi})
indicates target-language independence. Note that this is only a~one-sided implication, as poor correlation can be attributed to various factors.

\begin{figure*}[t]
    \centering
    \includegraphics[width=0.9\textwidth,trim= 10 10 10 0,clip]{figures/pmi_vs_chrf_Qwen3-14B-Base_HIGHEST.pdf}
    \noindent\rule{\textwidth}{0.5pt}
    \includegraphics[width=0.9\textwidth,trim= 10 10 10 0,clip]{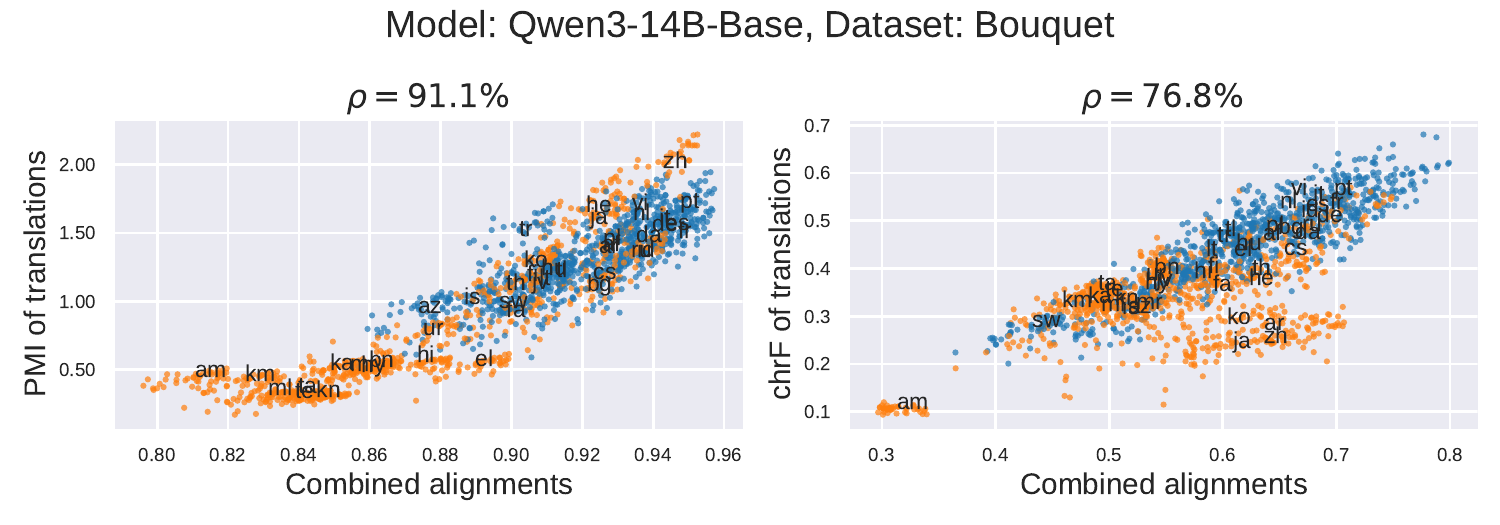}
    \noindent\rule{\textwidth}{0.5pt}
    \includegraphics[width=0.9\textwidth,trim= 10 10 10 0,clip]{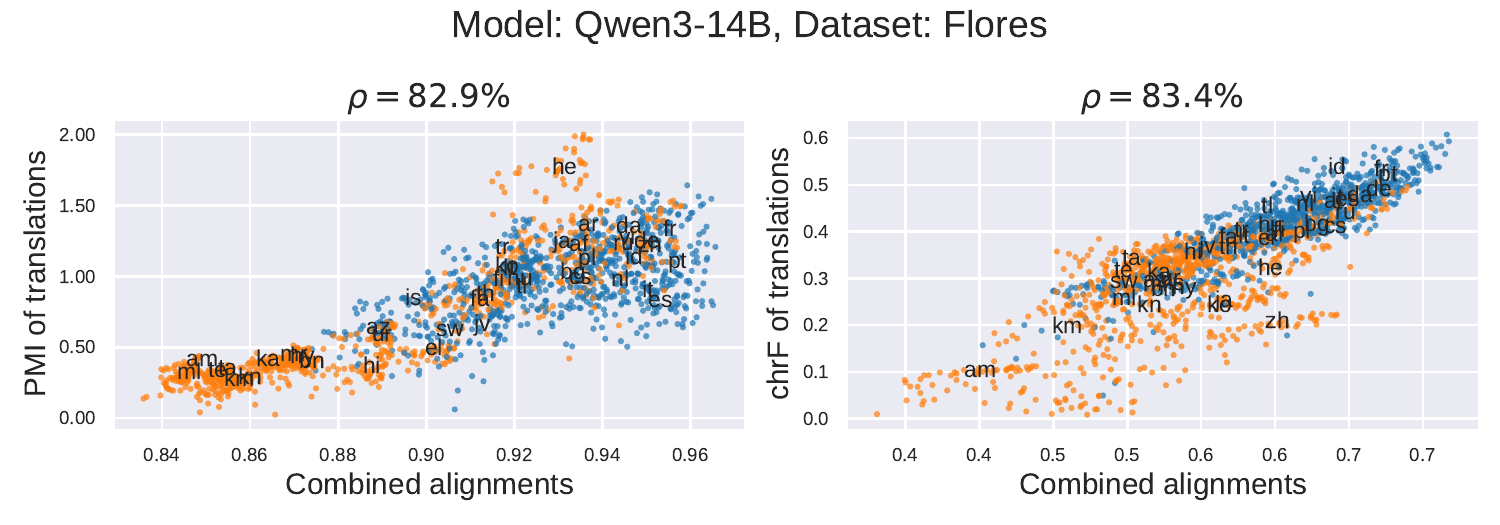}
    \noindent\rule{\textwidth}{0.5pt}
    \includegraphics[width=0.9\textwidth,trim= 10 10 10 0,clip]{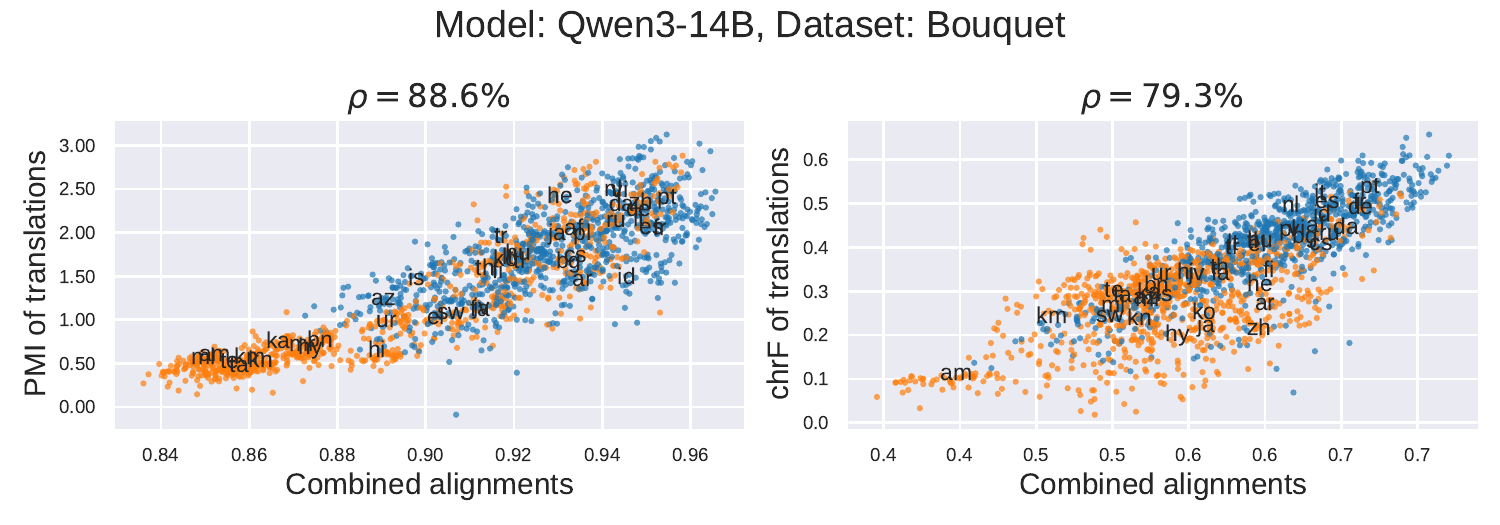}
    \caption{
    Comparison of how translation \acrshort{pmi} (left) and \textsc{chrF} (right) correlate with the \acrshort{cla} scores in the~best-case scenario for the \textbf{\texttt{Qwen3}} family.
    Each point is a~language pair; non-Latin-script target languages are in orange. Both metrics are correlated with their respective optimal convex combination (see Section~\ref{sec:corr-analysis-pmi}) of \texttt{src-en}, \texttt{en-tgt}, and \texttt{src-tgt} \acrshort{cla} scores, using the~best-correlating sentence representation and \acrshort{cla} metric. The optimal convex combination weights, representations, and metrics were calculated on the \textbf{Flores} dataset (odd rows) and reused on the \textbf{BOUQuET} dataset (even rows) for validation.}
    \label{fig:pmi-vs-chrf-qwen}
\end{figure*}

\begin{figure*}[t]
    \centering
    \includegraphics[width=0.88\textwidth,trim= 10 10 10 0,clip]{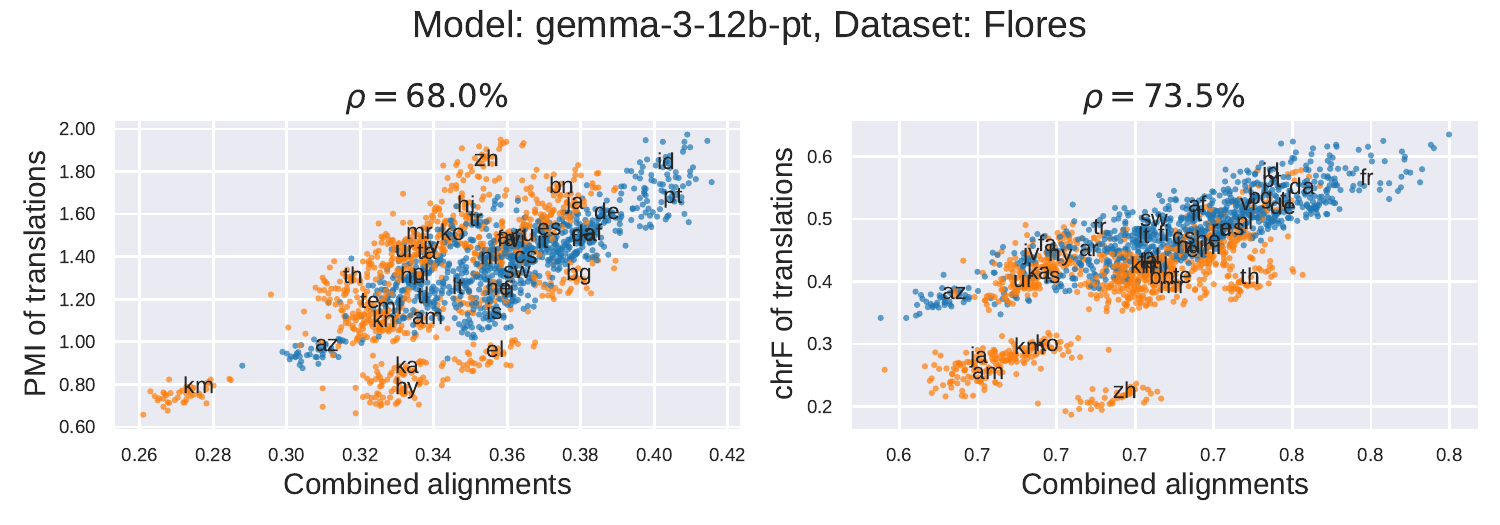}
    \noindent\rule{\textwidth}{0.5pt}
    \includegraphics[width=0.88\textwidth,trim= 10 10 10 0,clip]{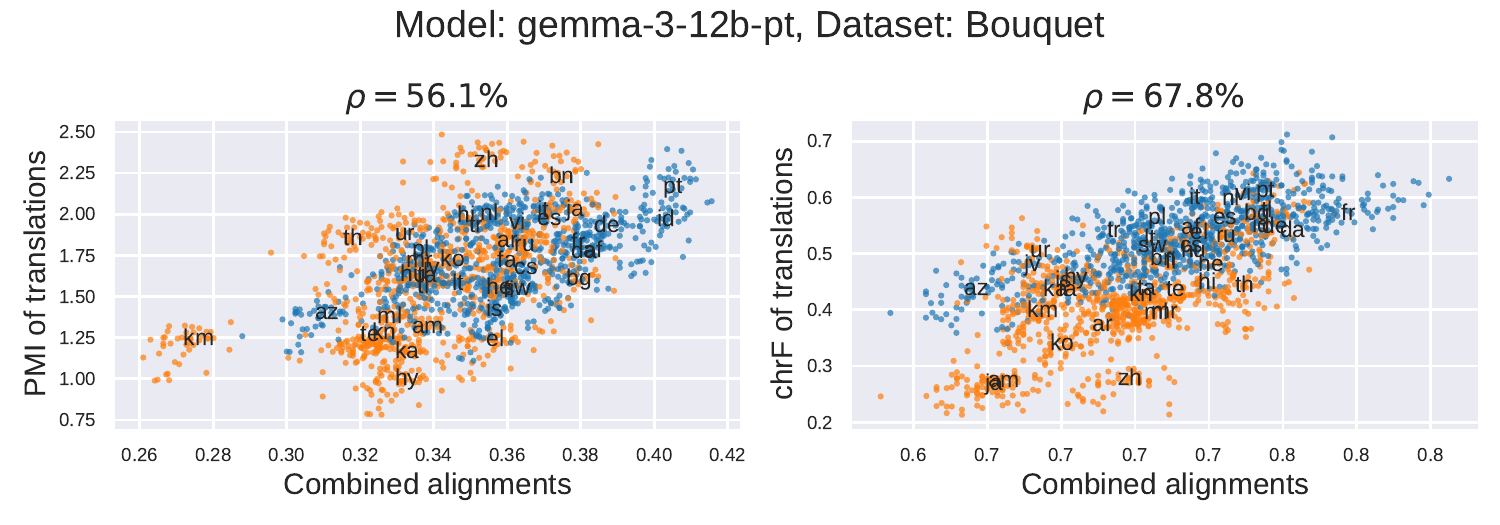}
    \noindent\rule{\textwidth}{0.5pt}
    \includegraphics[width=0.88\textwidth,trim= 10 10 10 0,clip]{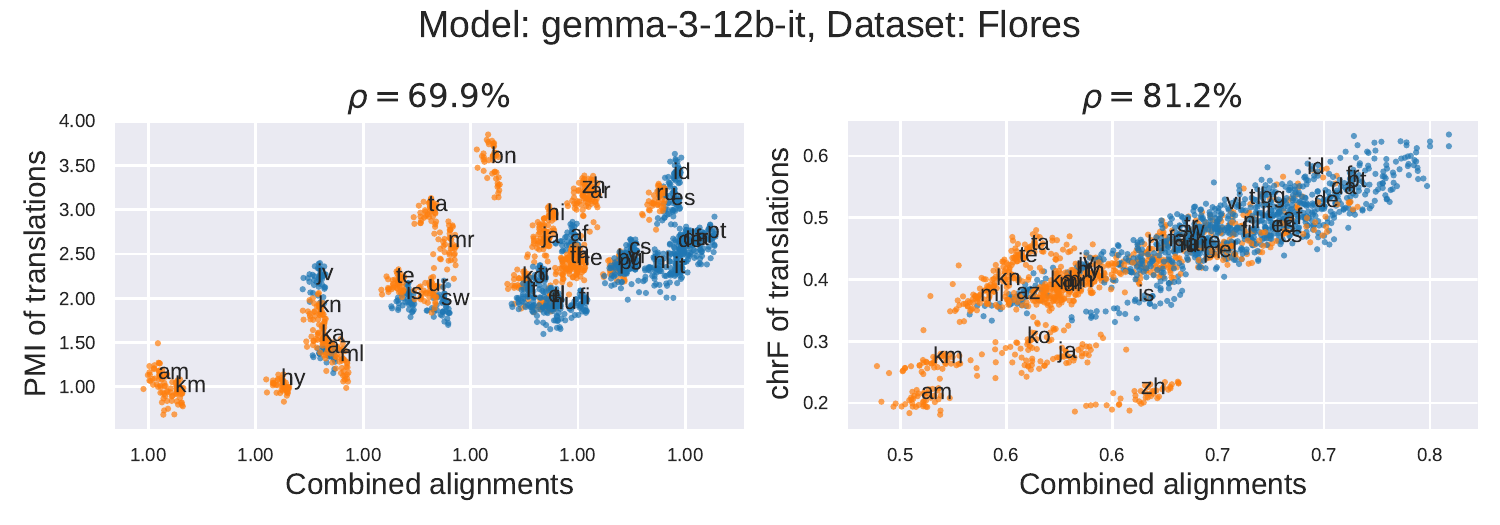}
    \noindent\rule{\textwidth}{0.5pt}
    \includegraphics[width=0.88\textwidth,trim= 10 10 10 0,clip]{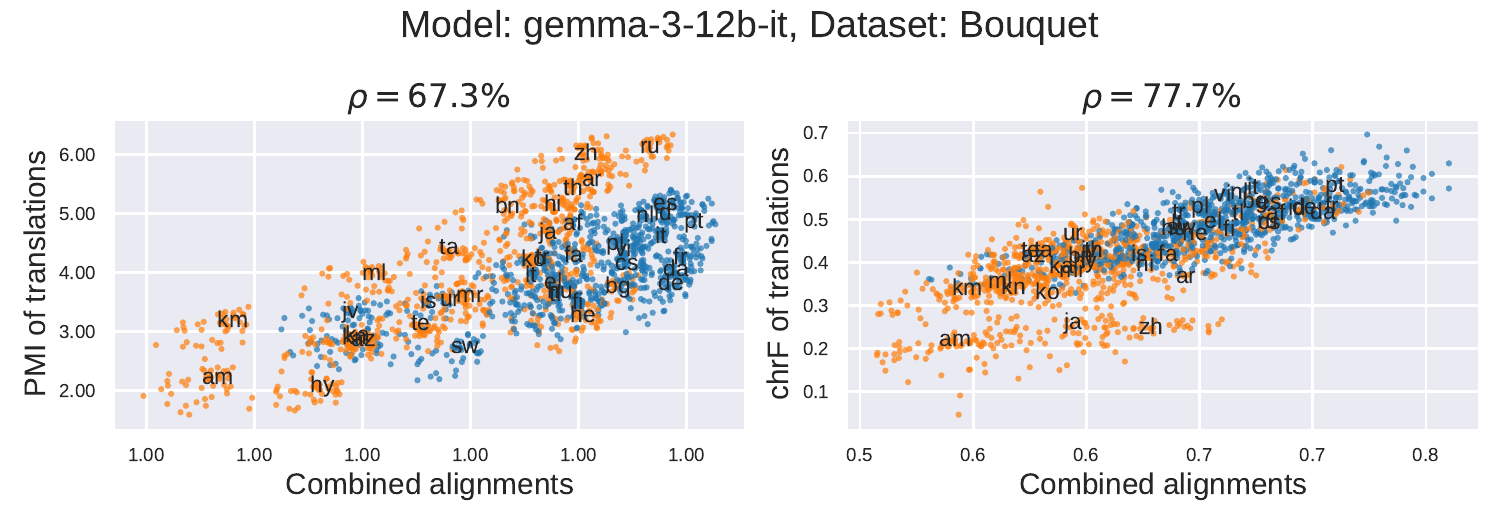}
    \caption{
    Comparison of how translation \acrshort{pmi} (left) and \textsc{chrF} (right) correlate with the \acrshort{cla} scores in the~best-case scenario for the \textbf{\texttt{gemma-3}} family.
    Each point is a~language pair; non-Latin-script target languages are in orange. Both metrics are correlated with their respective optimal convex combination (see Section~\ref{sec:corr-analysis-pmi}) of \texttt{src-en}, \texttt{en-tgt}, and \texttt{src-tgt} \acrshort{cla} scores, using the~best-correlating sentence representation and \acrshort{cla} metric. The optimal convex combination weights, representations, and metrics were calculated on the \textbf{Flores} dataset (odd rows) and reused on the \textbf{BOUQuET} dataset (even rows) for validation. \textbf{Note:} the outlying \texttt{gemma-3-12b-it} PMI performance on Flores is an anomaly observed in Table~\ref{tab:chrf-pmi-corr}, where it correlated weakly with \textsc{chrF}, compared to other models. This further leads to wrong parameter estimation and poor validation performance on BOUQuET.}
    \label{fig:pmi-vs-chrf-gemma}
\end{figure*}

\begin{figure*}[t]
    \centering
    \includegraphics[width=0.9\textwidth,trim= 10 10 10 0,clip]{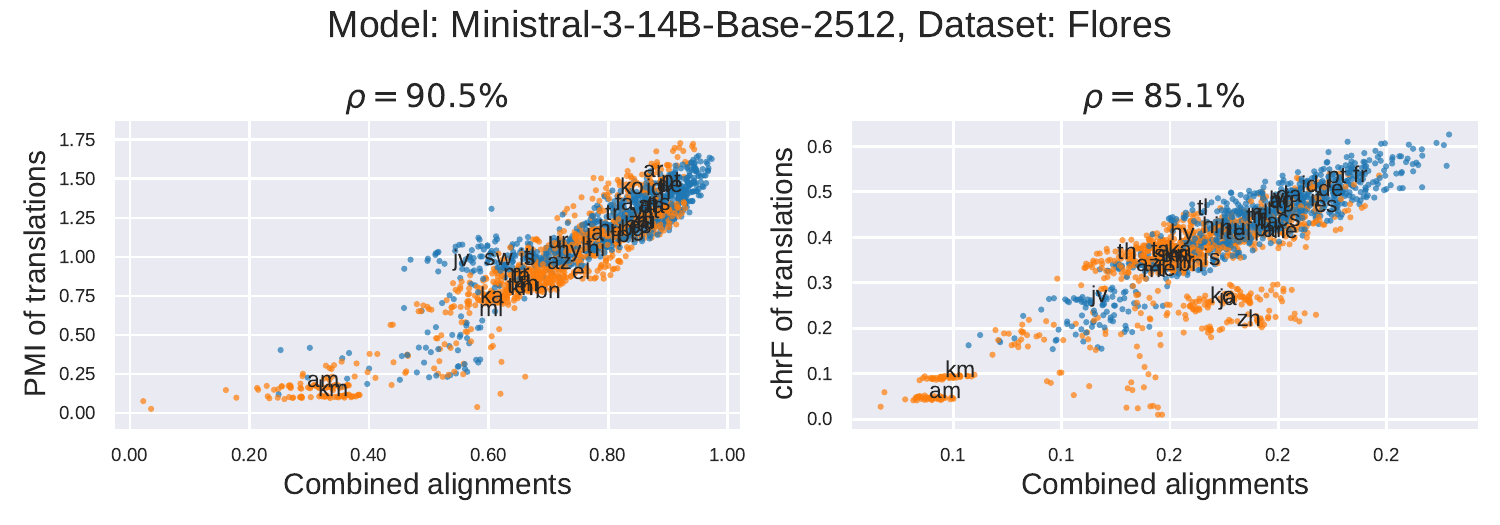}
    \noindent\rule{\textwidth}{0.5pt}
    \includegraphics[width=0.9\textwidth,trim= 10 10 10 0,clip]{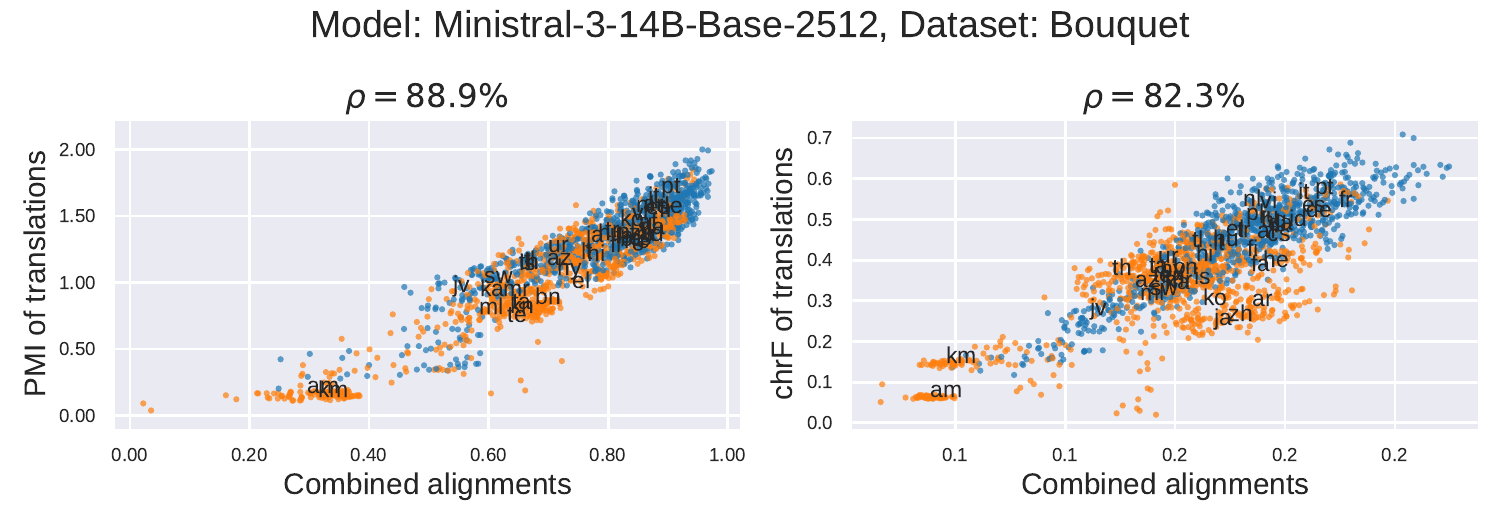}
    \noindent\rule{\textwidth}{0.5pt}
    \includegraphics[width=0.9\textwidth,trim= 10 10 10 0,clip]{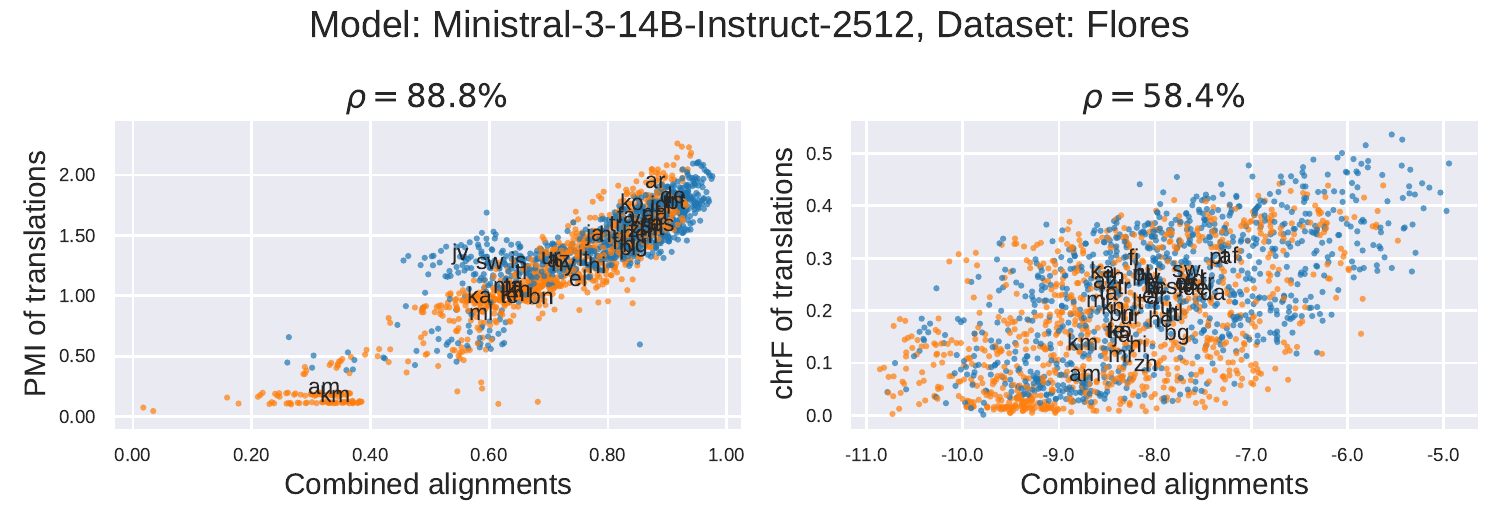}
    \noindent\rule{\textwidth}{0.5pt}
    \includegraphics[width=0.9\textwidth,trim= 10 10 10 0,clip]{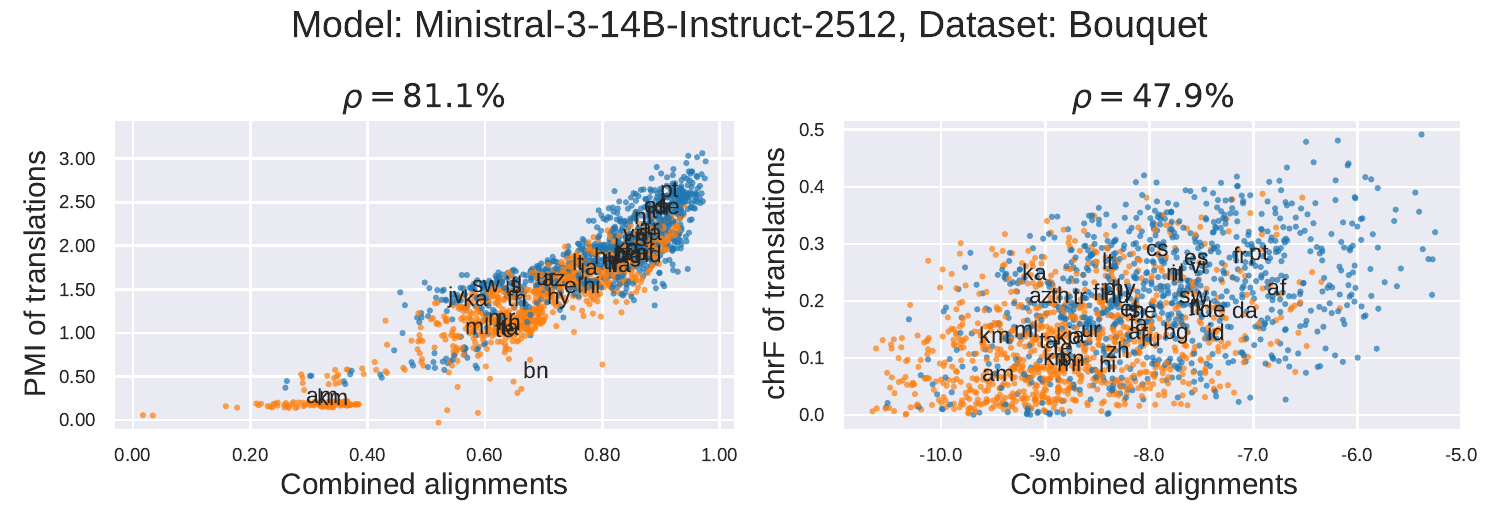}
    \caption{
    Comparison of how translation \acrshort{pmi} (left) and \textsc{chrF} (right) correlate with the \acrshort{cla} scores in the~best-case scenario for the \textbf{\texttt{Ministral-3}} family.
    Each point is a~language pair; non-Latin-script target languages are in orange. Both metrics are correlated with their respective optimal convex combination (see Section~\ref{sec:corr-analysis-pmi}) of \texttt{src-en}, \texttt{en-tgt}, and \texttt{src-tgt} \acrshort{cla} scores, using the~best-correlating sentence representation and \acrshort{cla} metric. The optimal convex combination weights, representations, and metrics were calculated on the \textbf{Flores} dataset (odd rows) and reused on the \textbf{BOUQuET} dataset (even rows) for validation. \textbf{Note} that the noise in \textsc{chrF} for the instr. variant is likely caused by the model often failing to follow the translation instructions, as described in Section~\ref{sec:chrf-analysis}.}
    \label{fig:pmi-vs-chrf-ministral}
\end{figure*}

Figure~\ref{fig:pmi-vs-chrf} shows the~comparison in correlation between \acrshort{pmi} (left) and \textsc{chrF} (right). The~plot for \textsc{chrF} score -- which is target-language-sensitive -- contains clusters of target languages which do not follow the~general pattern: most notably Chinese, for which the \textsc{chrF} score is considerably smaller than a~linear classifier would predict from the \acrshort{cla} scores. Such inconsistencies reduce correlation. While outlying target-language clusters are also present in the \acrshort{pmi} plot, they are less pronounced, resulting in higher correlation scores. Furthermore, as Figures~\ref{fig:pmi-vs-chrf-qwen}--\ref{fig:pmi-vs-chrf-ministral} show, the~clusters are not consistent across models and seem to be nuances of the~models rather than the~metric itself. Therefore, while we could not validate that our proposed \acrshort{pmi} metric is target-language-independent, empirically, it is \emph{less} dependent than \textsc{chrF} and, unlike \textsc{chrF}, does not exhibit systematic biases against specific languages across different models.

\newpage
\section{Detailed Results}\label{app:results}

\input{latex/tables/corr_with_sib-200_all_en-HIGHEST}

\input{latex/tables/corr_with_belebele_combined}

\input{latex/tables/translation_performance_by_language}

\input{latex/tables/translation_performance_by_language-bouquet}
\input{latex/tables/corr_both_with_chrf_combined}

\input{latex/tables/_corr_combination_with_chrf_combined}

\input{latex/tables/corr_pmi_combined}

\input{latex/tables/gammas_pmi_all_HIGHEST_pmi}

\input{latex/tables/corr_combination_with_chrf_all_-HIGHEST-validation}

\input{latex/tables/validation_pmi_all_HIGHEST_pmi}

\end{document}

%% file: latex/tables/corr_with_cla_en_Qwen3-14B.tex
\setlength{\tabcolsep}{4.5pt}
\begin{table}[t]
    \centering\footnotesize
    \begin{tabularx}{\columnwidth}%
    {p{4em} | C  C  C  C  C  C | C}
        \toprule
         & \textbf{cos} & \textbf{ANC} & \textbf{abs} & \textbf{dist} & \textbf{ratio} & \textbf{Dali} & \textbf{Eflo} \\
        \midrule
        \textbf{fewshot} & \percentGrad{95.3}{0} & \percentGrad{63.3}{0} & \percentGrad{87.7}{0} & \percentGrad{87.7}{0} & \percentGrad{87.7}{0} & -- & \percentGrad{58.0}{2} \\
        \textbf{last} & \percentGrad{97.6}{0} & \percentGrad{81.5}{0} & \percentGrad{89.6}{0} & \percentGrad{89.6}{0} & \percentGrad{89.7}{0} & \percentGrad{98.4}{0} & \percentGrad{58.0}{2} \\
        \textbf{p-wm} & \percentGrad{59.8}{0} & \percentGrad{91.5}{0} & \percentGrad{95.2}{0} & \percentGrad{95.2}{0} & \percentGrad{95.2}{0} & -- & \percentGrad{58.0}{0} \\
        \textbf{mean} & \percentGrad{97.7}{0} & \percentGrad{89.6}{0} & \percentGrad{55.3}{0} & \percentGrad{55.3}{0} & \percentGrad{55.4}{0} & -- & \percentGrad{58.0}{2} \\
        \textbf{prompt} & -- & \percentGrad{91.6}{0} & \percentGrad{99.6}{0} & \percentGrad{99.6}{0} & \percentGrad{99.6}{0} & -- & \percentGrad{58.0}{2} \\
        \bottomrule
    \end{tabularx}
    \caption{Strong correlation measured across the \texttt{src} languages between the \texttt{src-en} and \texttt{src-[\textasciitilde{}en]} alignment scores for Qwen3-14B. Values in per cent.}
    \label{tab:corr-Qwen3-14B}
\end{table}

%% file: latex/tables/overview_corr_with_sib-200_en.tex
\setlength{\tabcolsep}{3pt}
\begin{table}[t]
    \centering\footnotesize
    \begin{tabularx}{\columnwidth}%
    {p{2em} | C  C | C  C | C  C}
        \toprule
 & \multicolumn{2}{ c |}{\textbf{Qwen3}} & \multicolumn{2}{ c |}{\textbf{gemma-3}} & \multicolumn{2}{ c }{\textbf{Ministral-3}} \\
 & \textbf{base} & \textbf{inst.} & \textbf{base} & \textbf{inst.} & \textbf{base} & \textbf{inst.} \\
        \midrule
        \textbf{corr.} & \percentGrad{81.8}{0} & \percentGrad{85.9}{0} & \percentGrad{70.9}{0} & \percentGrad{67.6}{0} & \percentGrad{97.5}{1} & \percentGrad{97.4}{0}\\
        \textbf{repr.} & prompt & prompt & p-wm & mean & prompt & prompt \\
        \textbf{metr.} & abs & abs & cos & abs & abs & abs \\
        \bottomrule
    \end{tabularx}
    \caption{Best correlation measured across the \texttt{src} languages of \texttt{src-en} alignment score and the \textbf{{SIB-200}} \(F_1\) score in the \texttt{src} language for all models. Values in per cent.}
    \label{tab:overview-corr-en-task-sib-200}
\end{table}

%% file: latex/tables/overview_corr_with_belebele_en.tex
\setlength{\tabcolsep}{3pt}
\begin{table}[t]
    \centering\footnotesize
    \begin{tabularx}{\columnwidth}%
    {p{2em} | C  C | C  C | C  C}
        \toprule
 & \multicolumn{2}{ c |}{\textbf{Qwen3}} & \multicolumn{2}{ c |}{\textbf{gemma-3}} & \multicolumn{2}{ c }{\textbf{Ministral-3}} \\
 & \textbf{base} & \textbf{inst.} & \textbf{base} & \textbf{inst.} & \textbf{base} & \textbf{inst.} \\
        \midrule
        \textbf{corr.} & \percentGrad{91.7}{0} & \percentGrad{95.2}{0} & \percentGrad{70.0}{0} & \percentGrad{85.0}{0} & \percentGrad{97.4}{1} & \percentGrad{95.6}{0}\\
        \textbf{repr.} & prompt & prompt & p-wm & p-wm & p-wm & fewshot \\
        \textbf{metr.} & ANC & ANC & abs & ANC & ANC & abs \\
        \bottomrule
    \end{tabularx}
    \caption{
    Best correlation measured across the \texttt{src} languages of \texttt{src-en} alignment score and the \textbf{Belebele} accuracy in the \texttt{src} language for all models. Values in per cent.}
    \label{tab:overview-corr-en-task-belebele}
\end{table}

%% file: latex/tables/chrf_pmi_corr.tex
\setlength{\tabcolsep}{3pt}
\begin{table}[t]
    \centering\footnotesize
    \begin{tabularx}{\columnwidth}%
    {p{3.5em} | C  C | C  C | C  C}
        \toprule
 & \multicolumn{2}{ c |}{\textbf{Qwen3}} & \multicolumn{2}{ c |}{\textbf{gemma-3}} & \multicolumn{2}{ c }{\textbf{Ministral-3}} \\
 & \textbf{base} & \textbf{inst.} & \textbf{base} & \textbf{inst.} & \textbf{base} & \textbf{inst.} \\
        \midrule
        \textbf{flores} & \percentGrad{91.0}{0} & \percentGrad{77.6}{0} & \percentGrad{96.3}{1} & \percentGrad{26.2}{0} & \percentGrad{95.8}{0} & \percentGrad{65.8}{0} \\
        \textbf{bouquet} & \percentGrad{95.0}{0} & \percentGrad{83.0}{0} & \percentGrad{96.0}{0} & \percentGrad{78.5}{0} & \percentGrad{95.8}{0} & \percentGrad{77.7}{0} \\
    \bottomrule
    \end{tabularx}
    \caption{Correlations between system-level \textsc{chrF} and PMI, measured within a fixed target language and then averaged across the target languages. Values in per cent.}
    \label{tab:chrf-pmi-corr}
\end{table}

%% file: latex/tables/translation_performance_by_language-short.tex
\newcommand{\translationperformancebylanguageshortGrad}[2]{\gradientcell{#1}{4.591353414078971}{62.435515989119295}{cyan}{yellow}{70}{#2}}
\setlength{\tabcolsep}{4.5pt}
\begin{table}[ht]
    \centering\footnotesize
    \begin{tabularx}{\columnwidth}%
    {p{2em} | C  C | C  C | C  C}
        \toprule
 & \multicolumn{2}{ c |}{\textbf{Qwen3}} & \multicolumn{2}{ c |}{\textbf{gemma-3}} & \multicolumn{2}{ c }{\textbf{Ministral-3}} \\
 & \textbf{base} & \textbf{inst.} & \textbf{base} & \textbf{inst.} & \textbf{base} & \textbf{inst.} \\

        \midrule
        \textbf{en} & \translationperformancebylanguageshortGrad{59.4}{0} & \translationperformancebylanguageshortGrad{58.4}{0} & \translationperformancebylanguageshortGrad{62.4}{1} & \translationperformancebylanguageshortGrad{60.0}{0} & \translationperformancebylanguageshortGrad{59.4}{0} & \translationperformancebylanguageshortGrad{45.6}{0} \\
        \textbf{it} & \translationperformancebylanguageshortGrad{46.0}{0} & \translationperformancebylanguageshortGrad{46.0}{0} & \translationperformancebylanguageshortGrad{49.7}{0} & \translationperformancebylanguageshortGrad{50.1}{1} & \translationperformancebylanguageshortGrad{47.1}{0} & \translationperformancebylanguageshortGrad{18.7}{0} \\
        \textbf{ar} & \translationperformancebylanguageshortGrad{39.1}{0} & \translationperformancebylanguageshortGrad{38.0}{0} & \translationperformancebylanguageshortGrad{44.2}{0} & \translationperformancebylanguageshortGrad{44.7}{1} & \translationperformancebylanguageshortGrad{40.3}{0} & \translationperformancebylanguageshortGrad{21.7}{0} \\
        \textbf{zh} & \translationperformancebylanguageshortGrad{19.5}{0} & \translationperformancebylanguageshortGrad{19.3}{0} & \translationperformancebylanguageshortGrad{21.5}{1} & \translationperformancebylanguageshortGrad{21.4}{1} & \translationperformancebylanguageshortGrad{20.5}{0} & \translationperformancebylanguageshortGrad{8.7}{0} \\
        \textbf{am} & \translationperformancebylanguageshortGrad{7.9}{0} & \translationperformancebylanguageshortGrad{8.8}{0} & \translationperformancebylanguageshortGrad{24.5}{1} & \translationperformancebylanguageshortGrad{20.6}{0} & \translationperformancebylanguageshortGrad{4.6}{0} & \translationperformancebylanguageshortGrad{6.5}{0} \\
        \bottomrule
    \end{tabularx}
    \caption{Average \textbf{Flores} translation performance (\textsc{chrF}; over all source languages) for the displayed target languages. Values in cent.}
    \label{tab:translation-performance-by-language-short}
\end{table}

%% file: latex/tables/table_wide_corr_chrf.tex
\setlength{\tabcolsep}{2pt}
\begin{table}[t]
    \centering\footnotesize
    \begin{tabularx}{\columnwidth}%
    {p{3.1em} | C  C | C  C | C  C}
        \toprule
         & \multicolumn{2}{ c |}{\textbf{Qwen3}} & \multicolumn{2}{ c |}{\textbf{gemma-3}} & \multicolumn{2}{ c }{\textbf{Ministral-3}} \\
         & \textbf{base} & \textbf{inst.} & \textbf{base} & \textbf{inst.} & \textbf{base} & \textbf{inst.} \\
        \midrule
        $\rho_\mathrm{src-en}$ & \percentGrad{81.8}{0} & \percentGrad{81.7}{0} & \percentGrad{66.7}{0} & \percentGrad{77.8}{0} & \percentGrad{87.6}{1} & \percentGrad{63.4}{0} \\
        \textbf{repr.} & fewshot & prompt & p-wm & fewshot & fewshot & last \\
        \textbf{metric} & abs & ANC & ANC & ANC & abs & ANC\\
        \midrule
        $\rho_\mathrm{src-tgt}$ & \percentGrad{59.4}{0} & \percentGrad{63.4}{0} & \percentGrad{70.1}{0} & \percentGrad{75.7}{0} & \percentGrad{81.1}{1} & \percentGrad{42.5}{0} \\
        \textbf{repr.} & p-wm & prompt & p-wm & prompt & last & last \\
        \textbf{metric} & ANC & ANC & ANC & ANC & abs & ANC \\
        \midrule
        $\rho_\oplus$ & \percentGrad{82.0}{0} & \percentGrad{81.7}{0} & \percentGrad{75.8}{0} & \percentGrad{82.9}{0} & \percentGrad{87.6}{1} & \percentGrad{63.4}{0}\\
        \textbf{impr.} & \percentGrad{0.2}{0} & \percentGrad{0}{0} & \percentGrad{4.6}{0} & \percentGrad{5.1}{1} & \percentGrad{0}{0} & \percentGrad{0}{0}\\
        1--$\alpha$ & \percentGrad{20}{0} & \percentGrad{0}{0} & \percentGrad{72}{0} & \percentGrad{64}{0} & \percentGrad{0}{0} & \percentGrad{0}{0} \\
        \textbf{repr.}  & p-wm & prompt & p-wm & fewshot & fewshot & last \\
        \textbf{metric}  & ANC & ANC & ANC & ANC & abs & ANC \\
        $\rho_\oplus^\mathrm{val}$ & \percentGrad{75.2}{0} & \percentGrad{72.2}{0} & \percentGrad{67.2}{0} & \percentGrad{75.1}{0} & \percentGrad{85.2}{1} & \percentGrad{53.6}{0} \\
        \bottomrule
    \end{tabularx}
    \caption{
    Best correlation between \textbf{Flores} translation \textsc{chrF} and \texttt{src-en} (top), \texttt{src-tgt} (middle), and the optimal convex combination of \texttt{en-tgt} and \texttt{src-tgt} (bottom) \acrshort{cla} scores; \textbf{impr}ovement of the combination over \texttt{src-en} alone and the \texttt{src-tgt} weight ($1-\alpha$). The validation score ($\rho_\oplus^\mathrm{val}$) applies the optimal parameters above to the \textbf{BOUQuET} dataset. Correlations are measured across \texttt{src} languages for a fixed \texttt{tgt}, averaged over target languages. Values in per cent.}
\label{tab:overview-all-corr-combined-task-chrf}
\end{table}

%% file: latex/tables/corr_pmi_overview_combined.tex
\setlength{\tabcolsep}{3pt}
\begin{table}[t]
    \centering\footnotesize
    \begin{tabularx}{\columnwidth}%
    {p{3.1em} | C  C | C  C | C  C}
        \toprule
        \multicolumn{7}{c}{\textbf{main language set (44 languages)}} \\
        \midrule
         & \multicolumn{2}{ c |}{\textbf{Qwen3}} & \multicolumn{2}{ c |}{\textbf{gemma-3}} & \multicolumn{2}{ c }{\textbf{Ministral-3}} \\
         & \textbf{base} & \textbf{inst.} & \textbf{base} & \textbf{inst.} & \textbf{base} & \textbf{inst.} \\
        \midrule
        $\rho_\oplus$ & \percentGrad{91.2}{1} & \percentGrad{82.9}{0} & \percentGrad{68.0}{0} & \percentGrad{69.9}{0} & \percentGrad{90.5}{0} & \percentGrad{88.8}{0} \\
        1--$\alpha$--$\beta$ & \percentGrad{0.0}{0} & \percentGrad{8.0}{0} & \percentGrad{0.0}{0} & \percentGrad{12.0}{1} & \percentGrad{4.0}{0} & \percentGrad{0.0}{0} \\
        \textbf{repr.} & p-wm & p-wm & last & p-wm & p-wm & p-wm \\
        \textbf{metric} & cos & cos & ANC & cos & ratio & abs \\
        $\rho_\oplus^\mathrm{val}$ & \percentGrad{91.1}{1} & \percentGrad{88.5}{0} & \percentGrad{56.1}{0} & \percentGrad{66.8}{0} & \percentGrad{88.9}{0} & \percentGrad{81.1}{0} \\
        \midrule
        \multicolumn{7}{c}{\textbf{related language pairs (7 pairs)}} \\
        \midrule
         & \multicolumn{2}{c|}{\textbf{Qwen3}} & \multicolumn{2}{c|}{\textbf{gemma-3}} & \multicolumn{2}{c}{\textbf{Ministral-3}} \\
         & \multicolumn{2}{c|}{\textbf{base}} & \multicolumn{2}{c|}{\textbf{base}} & \multicolumn{2}{c}{\textbf{base}} \\
        \midrule
        $\rho_\oplus$ & \multicolumn{2}{c|}{\percentGrad{84.7}{0}\hspace{0.6em}} & \multicolumn{2}{c|}{\percentGrad{76.8}{0}\hspace{0.6em}} & \multicolumn{2}{c}{\percentGrad{90.0}{1}\hspace{0.6em}} \\
        1--$\alpha$--$\beta$ & \multicolumn{2}{c|}{\percentGrad{88.0}{1}\hspace{0.6em}} & \multicolumn{2}{c|}{\percentGrad{84.0}{0}\hspace{0.6em}} & \multicolumn{2}{c}{\percentGrad{84.0}{0}\hspace{0.6em}} \\
        \textbf{repr.} & \multicolumn{2}{c|}{mean} & \multicolumn{2}{c|}{--} & \multicolumn{2}{c}{mean} \\
        \textbf{metric} & \multicolumn{2}{c|}{cos} & \multicolumn{2}{c|}{Eflo} & \multicolumn{2}{c}{cos} \\
        \bottomrule
    \end{tabularx}
    \caption{Best correlations between the optimal convex combination of \texttt{src-en}, \texttt{en-tgt}, and \texttt{src-tgt} and the \textbf{Flores} translation PMI for our main set of 44 languages (top) and a set of seven closely related languages (bottom; base models only). The validation score ($\rho_\oplus^\mathrm{val}$) applies the optimal parameters above to the \textbf{BOUQuET} dataset. Values in per cent.}
    \label{tab:corrPmiOverviewCombined}
\end{table}

%% file: latex/tables/corr_with_sib-200_all_en-HIGHEST.tex
\begin{table}[!htb]
\centering\footnotesize
\begin{tabularx}{\columnwidth}%
{p{4em} | C  C  C  C  C  C | C}\toprule
\multicolumn{8}{c}{\textbf{Qwen3-14B-Base}} \\
\midrule
         & \textbf{cos} & \textbf{ANC} & \textbf{abs} & \textbf{dist} & \textbf{ratio} & \textbf{Dali} & \textbf{Eflo} \\
        \midrule
        \textbf{fewshot} & \percentGrad{26.9}{0} & \percentGrad{29.6}{0} & \percentGrad{61.0}{0} & \percentGrad{61.0}{0} & \percentGrad{61.3}{0} & -- & \percentGrad{16.8}{2} \\
        \textbf{last} & \percentGrad{25.6}{0} & \percentGrad{29.5}{0} & \percentGrad{61.8}{0} & \percentGrad{61.8}{0} & \percentGrad{62.0}{0} & \percentGrad{18.7}{0} & \percentGrad{16.8}{2} \\
        \textbf{p-wm} & \percentGrad{40.5}{0} & \percentGrad{54.0}{0} & \percentGrad{79.4}{0} & \percentGrad{79.4}{0} & \percentGrad{79.5}{0} & -- & \percentGrad{16.8}{0} \\
        \textbf{mean} & \percentGrad{21.4}{0} & \percentGrad{54.7}{0} & \percentGrad{64.8}{0} & \percentGrad{64.8}{0} & \percentGrad{64.9}{0} & -- & \percentGrad{16.8}{2} \\
        \textbf{prompt} & -- & \percentGrad{59.4}{0} & \percentGrad{81.8}{1} & \percentGrad{81.8}{1} & \percentGrad{81.8}{1} & -- & \percentGrad{16.8}{2} \\
\midrule
\multicolumn{8}{c}{\textbf{Qwen3-14B}} \\
\midrule
         & \textbf{cos} & \textbf{ANC} & \textbf{abs} & \textbf{dist} & \textbf{ratio} & \textbf{Dali} & \textbf{Eflo} \\
        \midrule
        \textbf{fewshot} & -- & \percentGrad{59.7}{0} & \percentGrad{44.1}{0} & \percentGrad{44.1}{0} & \percentGrad{44.1}{0} & -- & \percentGrad{16.6}{2} \\
        \textbf{last} & \percentGrad{16.1}{0} & \percentGrad{28.2}{0} & \percentGrad{63.8}{0} & \percentGrad{63.8}{0} & \percentGrad{63.7}{0} & \percentGrad{53.5}{0} & \percentGrad{16.6}{2} \\
        \textbf{p-wm} & \percentGrad{38.4}{0} & \percentGrad{65.5}{0} & \percentGrad{78.5}{0} & \percentGrad{78.5}{0} & \percentGrad{78.5}{0} & -- & \percentGrad{16.6}{0} \\
        \textbf{mean} & \percentGrad{18.9}{0} & \percentGrad{67.2}{0} & \percentGrad{75.7}{0} & \percentGrad{75.7}{0} & \percentGrad{75.9}{0} & -- & \percentGrad{16.6}{2} \\
        \textbf{prompt} & -- & \percentGrad{81.6}{0} & \percentGrad{85.9}{1} & \percentGrad{85.9}{1} & \percentGrad{85.9}{1} & -- & \percentGrad{16.6}{2} \\
\midrule
\multicolumn{8}{c}{\textbf{gemma-3-12b-pt}} \\
\midrule
         & \textbf{cos} & \textbf{ANC} & \textbf{abs} & \textbf{dist} & \textbf{ratio} & \textbf{Dali} & \textbf{Eflo} \\
        \midrule
        \textbf{fewshot} & \percentGrad{58.2}{0} & \percentGrad{62.1}{0} & \percentGrad{70.0}{0} & \percentGrad{70.0}{0} & \percentGrad{70.0}{0} & -- & \percentGrad{49.9}{2} \\
        \textbf{last} & \percentGrad{20.3}{0} & \percentGrad{65.4}{0} & \percentGrad{65.7}{0} & \percentGrad{65.7}{0} & \percentGrad{65.7}{0} & \percentGrad{62.4}{0} & \percentGrad{49.9}{2} \\
        \textbf{p-wm} & \percentGrad{70.9}{1} & \percentGrad{62.8}{0} & \percentGrad{58.8}{0} & \percentGrad{58.8}{0} & \percentGrad{58.8}{0} & -- & \percentGrad{49.9}{0} \\
        \textbf{mean} & -- & \percentGrad{61.9}{0} & \percentGrad{62.4}{0} & \percentGrad{62.4}{0} & \percentGrad{62.4}{0} & -- & \percentGrad{49.9}{2} \\
        \textbf{prompt} & -- & \percentGrad{33.3}{0} & \percentGrad{55.5}{0} & \percentGrad{55.5}{0} & \percentGrad{55.3}{0} & -- & \percentGrad{49.9}{2} \\
\midrule
\multicolumn{8}{c}{\textbf{gemma-3-12b-it}} \\
\midrule
         & \textbf{cos} & \textbf{ANC} & \textbf{abs} & \textbf{dist} & \textbf{ratio} & \textbf{Dali} & \textbf{Eflo} \\
        \midrule
        \textbf{fewshot} & \percentGrad{12.1}{0} & \percentGrad{40.3}{0} & \percentGrad{53.2}{0} & \percentGrad{53.2}{0} & \percentGrad{53.1}{0} & -- & \percentGrad{24.5}{2} \\
        \textbf{last} & \percentGrad{48.7}{0} & \percentGrad{45.7}{0} & \percentGrad{50.6}{0} & \percentGrad{50.6}{0} & \percentGrad{50.5}{0} & \percentGrad{9.6}{0} & \percentGrad{24.5}{2} \\
        \textbf{p-wm} & \percentGrad{60.9}{0} & \percentGrad{49.6}{0} & \percentGrad{61.0}{0} & \percentGrad{61.0}{0} & \percentGrad{61.0}{0} & -- & \percentGrad{24.5}{0} \\
        \textbf{mean} & -- & \percentGrad{49.4}{0} & \percentGrad{67.6}{1} & \percentGrad{67.6}{1} & \percentGrad{67.6}{1} & -- & \percentGrad{24.5}{2} \\
        \textbf{prompt} & -- & \percentGrad{50.9}{0} & \percentGrad{62.0}{0} & \percentGrad{62.0}{0} & \percentGrad{62.0}{0} & -- & \percentGrad{24.5}{2} \\
\midrule
\multicolumn{8}{c}{\textbf{Ministral-3-14B-Base-2512}} \\
\midrule
         & \textbf{cos} & \textbf{ANC} & \textbf{abs} & \textbf{dist} & \textbf{ratio} & \textbf{Dali} & \textbf{Eflo} \\
        \midrule
        \textbf{fewshot} & \percentGrad{56.4}{0} & \percentGrad{74.7}{0} & \percentGrad{89.5}{0} & \percentGrad{89.5}{0} & \percentGrad{89.5}{0} & -- & \percentGrad{45.2}{2} \\
        \textbf{last} & \percentGrad{73.0}{0} & \percentGrad{66.1}{0} & \percentGrad{73.4}{0} & \percentGrad{73.4}{0} & \percentGrad{73.2}{0} & \percentGrad{92.9}{0} & \percentGrad{45.2}{2} \\
        \textbf{p-wm} & \percentGrad{91.9}{0} & \percentGrad{87.2}{0} & \percentGrad{83.8}{0} & \percentGrad{83.8}{0} & \percentGrad{84.1}{0} & -- & \percentGrad{45.2}{0} \\
        \textbf{mean} & \percentGrad{69.4}{0} & \percentGrad{87.0}{0} & \percentGrad{80.3}{0} & \percentGrad{80.3}{0} & \percentGrad{80.8}{0} & -- & \percentGrad{45.2}{2} \\
        \textbf{prompt} & -- & \percentGrad{92.5}{0} & \percentGrad{97.5}{1} & \percentGrad{97.5}{1} & \percentGrad{97.5}{1} & -- & \percentGrad{45.2}{2} \\
\midrule
\multicolumn{8}{c}{\textbf{Ministral-3-14B-Instruct-2512}} \\
\midrule
         & \textbf{cos} & \textbf{ANC} & \textbf{abs} & \textbf{dist} & \textbf{ratio} & \textbf{Dali} & \textbf{Eflo} \\
        \midrule
        \textbf{fewshot} & -- & \percentGrad{86.7}{0} & \percentGrad{95.2}{0} & \percentGrad{95.2}{0} & \percentGrad{95.2}{0} & -- & \percentGrad{36.7}{2} \\
        \textbf{last} & \percentGrad{70.8}{0} & \percentGrad{64.4}{0} & \percentGrad{92.3}{0} & \percentGrad{92.3}{0} & \percentGrad{92.3}{0} & \percentGrad{88.3}{0} & \percentGrad{36.7}{2} \\
        \textbf{p-wm} & \percentGrad{89.5}{0} & \percentGrad{84.7}{0} & \percentGrad{80.9}{0} & \percentGrad{80.9}{0} & \percentGrad{80.9}{0} & -- & \percentGrad{36.7}{0} \\
        \textbf{mean} & \percentGrad{65.4}{0} & \percentGrad{80.3}{0} & \percentGrad{71.7}{0} & \percentGrad{71.7}{0} & \percentGrad{72.3}{0} & -- & \percentGrad{36.7}{2} \\
        \textbf{prompt} & \percentGrad{9.7}{0} & \percentGrad{87.5}{0} & \percentGrad{97.4}{1} & \percentGrad{97.4}{1} & \percentGrad{97.4}{1} & -- & \percentGrad{36.7}{2} \\
\bottomrule
\end{tabularx}
\caption{Correlation measured across the \texttt{src} languages of \texttt{src-en} alignment score and the \textbf{{SIB-200}} \(F_1\) score in the \texttt{src} language for all models. Valuesin cent.}
\label{tab:corr-en-task-sib-200-all-HIGHEST}
\end{table}

%% file: latex/tables/corr_with_belebele_combined.tex
\begin{table}[ht]
\setlength{\tabcolsep}{3pt}
\centering\footnotesize
\begin{tabularx}{\columnwidth}%
{p{3em} | C  C  C  C  C  C | C}\toprule
\multicolumn{8}{c}{\textbf{Qwen3-14B-Base}} \\
\midrule
         & \textbf{cos} & \textbf{ANC} & \textbf{abs} & \textbf{dist} & \textbf{ratio} & \textbf{Dali} & \textbf{Eflo} \\
        \midrule
        \textbf{fewshot} & \percentGrad{47.8}{0} & \percentGrad{50.5}{0} & \percentGrad{91.4}{0} & \percentGrad{91.4}{0} & \percentGrad{91.5}{0} & -- & \percentGrad{50.2}{2} \\
        \textbf{last} & \percentGrad{42.2}{0} & \percentGrad{51.2}{0} & \percentGrad{81.4}{0} & \percentGrad{81.4}{0} & \percentGrad{81.4}{0} & \percentGrad{24.3}{0} & \percentGrad{50.2}{2} \\
        \textbf{p-wm} & \percentGrad{80.3}{0} & \percentGrad{91.0}{0} & \percentGrad{89.7}{0} & \percentGrad{89.7}{0} & \percentGrad{89.7}{0} & -- & \percentGrad{50.2}{0} \\
        \textbf{mean} & \percentGrad{59.1}{0} & \percentGrad{91.4}{0} & \percentGrad{89.3}{0} & \percentGrad{89.3}{0} & \percentGrad{89.3}{0} & -- & \percentGrad{50.2}{2} \\
        \textbf{prompt} & -- & \percentGrad{91.7}{1} & \percentGrad{85.4}{0} & \percentGrad{85.4}{0} & \percentGrad{85.3}{0} & -- & \percentGrad{50.2}{2} \\
\midrule
\multicolumn{8}{c}{\textbf{Qwen3-14B}} \\
\midrule
         & \textbf{cos} & \textbf{ANC} & \textbf{abs} & \textbf{dist} & \textbf{ratio} & \textbf{Dali} & \textbf{Eflo} \\
        \midrule
        \textbf{fewshot} & -- & \percentGrad{89.6}{0} & \percentGrad{83.3}{0} & \percentGrad{83.3}{0} & \percentGrad{83.3}{0} & -- & \percentGrad{51.8}{2} \\
        \textbf{last} & \percentGrad{42.3}{0} & \percentGrad{56.0}{0} & \percentGrad{81.7}{0} & \percentGrad{81.7}{0} & \percentGrad{81.6}{0} & \percentGrad{80.8}{0} & \percentGrad{51.8}{2} \\
        \textbf{p-wm} & \percentGrad{76.4}{0} & \percentGrad{91.8}{0} & \percentGrad{90.6}{0} & \percentGrad{90.6}{0} & \percentGrad{90.6}{0} & -- & \percentGrad{51.8}{0} \\
        \textbf{mean} & \percentGrad{58.0}{0} & \percentGrad{92.4}{0} & \percentGrad{86.4}{0} & \percentGrad{86.4}{0} & \percentGrad{86.4}{0} & -- & \percentGrad{51.8}{2} \\
        \textbf{prompt} & -- & \percentGrad{95.2}{1} & \percentGrad{81.9}{0} & \percentGrad{81.9}{0} & \percentGrad{81.9}{0} & -- & \percentGrad{51.8}{2} \\
\midrule
\multicolumn{8}{c}{\textbf{gemma-3-12b-pt}} \\
\midrule
         & \textbf{cos} & \textbf{ANC} & \textbf{abs} & \textbf{dist} & \textbf{ratio} & \textbf{Dali} & \textbf{Eflo} \\
        \midrule
        \textbf{fewshot} & \percentGrad{46.5}{0} & \percentGrad{59.3}{0} & \percentGrad{56.4}{0} & \percentGrad{56.4}{0} & \percentGrad{56.3}{0} & -- & \percentGrad{24.2}{2} \\
        \textbf{last} & \percentGrad{32.3}{0} & \percentGrad{49.4}{0} & \percentGrad{59.9}{0} & \percentGrad{59.9}{0} & \percentGrad{59.9}{0} & \percentGrad{42.6}{0} & \percentGrad{24.2}{2} \\
        \textbf{p-wm} & \percentGrad{62.5}{0} & \percentGrad{66.6}{0} & \percentGrad{70.0}{1} & \percentGrad{70.0}{1} & \percentGrad{70.0}{1} & -- & \percentGrad{24.2}{0} \\
        \textbf{mean} & -- & \percentGrad{66.7}{0} & \percentGrad{64.0}{0} & \percentGrad{64.0}{0} & \percentGrad{64.0}{0} & -- & \percentGrad{24.2}{2} \\
        \textbf{prompt} & -- & \percentGrad{31.1}{0} & \percentGrad{59.6}{0} & \percentGrad{59.6}{0} & \percentGrad{59.2}{0} & -- & \percentGrad{24.2}{2} \\
\midrule
\multicolumn{8}{c}{\textbf{gemma-3-12b-it}} \\
\midrule
         & \textbf{cos} & \textbf{ANC} & \textbf{abs} & \textbf{dist} & \textbf{ratio} & \textbf{Dali} & \textbf{Eflo} \\
        \midrule
        \textbf{fewshot} & \percentGrad{20.3}{0} & \percentGrad{81.8}{0} & \percentGrad{82.6}{0} & \percentGrad{82.6}{0} & \percentGrad{82.6}{0} & -- & \percentGrad{58.9}{2} \\
        \textbf{last} & \percentGrad{68.5}{0} & \percentGrad{69.2}{0} & \percentGrad{80.6}{0} & \percentGrad{80.6}{0} & \percentGrad{80.7}{0} & \percentGrad{12.6}{0} & \percentGrad{58.9}{2} \\
        \textbf{p-wm} & \percentGrad{81.7}{0} & \percentGrad{85.0}{1} & \percentGrad{69.2}{0} & \percentGrad{69.2}{0} & \percentGrad{69.2}{0} & -- & \percentGrad{58.9}{0} \\
        \textbf{mean} & -- & \percentGrad{84.4}{0} & \percentGrad{65.2}{0} & \percentGrad{65.2}{0} & \percentGrad{65.2}{0} & -- & \percentGrad{58.9}{2} \\
        \textbf{prompt} & -- & \percentGrad{83.7}{0} & \percentGrad{65.9}{0} & \percentGrad{65.9}{0} & \percentGrad{65.9}{0} & -- & \percentGrad{58.9}{2} \\
\midrule
\multicolumn{8}{c}{\textbf{Ministral-3-14B-Base-2512}} \\
\midrule
         & \textbf{cos} & \textbf{ANC} & \textbf{abs} & \textbf{dist} & \textbf{ratio} & \textbf{Dali} & \textbf{Eflo} \\
        \midrule
        \textbf{fewshot} & \percentGrad{58.8}{0} & \percentGrad{78.0}{0} & \percentGrad{95.4}{0} & \percentGrad{95.4}{0} & \percentGrad{95.4}{0} & -- & \percentGrad{53.3}{2} \\
        \textbf{last} & \percentGrad{88.0}{0} & \percentGrad{84.8}{0} & \percentGrad{89.3}{0} & \percentGrad{89.3}{0} & \percentGrad{89.3}{0} & \percentGrad{89.6}{0} & \percentGrad{53.3}{2} \\
        \textbf{p-wm} & \percentGrad{90.3}{0} & \percentGrad{97.4}{1} & \percentGrad{95.5}{0} & \percentGrad{95.5}{0} & \percentGrad{95.7}{0} & -- & \percentGrad{53.3}{0} \\
        \textbf{mean} & \percentGrad{74.0}{0} & \percentGrad{97.4}{1} & \percentGrad{93.7}{0} & \percentGrad{93.7}{0} & \percentGrad{93.9}{0} & -- & \percentGrad{53.3}{2} \\
        \textbf{prompt} & -- & \percentGrad{93.5}{0} & \percentGrad{87.8}{0} & \percentGrad{87.8}{0} & \percentGrad{87.8}{0} & -- & \percentGrad{53.3}{2} \\
\midrule
\multicolumn{8}{c}{\textbf{Ministral-3-14B-Instruct-2512}} \\
\midrule
         & \textbf{cos} & \textbf{ANC} & \textbf{abs} & \textbf{dist} & \textbf{ratio} & \textbf{Dali} & \textbf{Eflo} \\
        \midrule
        \textbf{fewshot} & -- & \percentGrad{94.6}{0} & \percentGrad{95.7}{1} & \percentGrad{95.7}{1} & \percentGrad{95.7}{1} & -- & \percentGrad{50.8}{2} \\
        \textbf{last} & \percentGrad{72.8}{0} & \percentGrad{73.6}{0} & \percentGrad{94.3}{0} & \percentGrad{94.3}{0} & \percentGrad{94.4}{0} & \percentGrad{93.1}{0} & \percentGrad{50.8}{2} \\
        \textbf{p-wm} & \percentGrad{92.5}{0} & \percentGrad{94.2}{0} & \percentGrad{91.8}{0} & \percentGrad{91.8}{0} & \percentGrad{91.8}{0} & -- & \percentGrad{50.8}{0} \\
        \textbf{mean} & \percentGrad{74.2}{0} & \percentGrad{91.8}{0} & \percentGrad{84.8}{0} & \percentGrad{84.8}{0} & \percentGrad{85.1}{0} & -- & \percentGrad{50.8}{2} \\
        \textbf{prompt} & \percentGrad{4.2}{0} & \percentGrad{95.6}{1} & \percentGrad{94.3}{0} & \percentGrad{94.3}{0} & \percentGrad{94.3}{0} & -- & \percentGrad{50.8}{2} \\
\bottomrule
\end{tabularx}
\caption{Correlation measured across the \texttt{src} languages of \texttt{src-en} alignment score and the \textbf{{Belebele}} accuracy in the \texttt{src} language for all models. Values in per cent.}
\label{tab:corr-en-task-belebele-all-HIGHEST}
\end{table}

%% file: latex/tables/translation_performance_by_language.tex
\newcommand{\translationperformancebylanguageGrad}[2]{\gradientcell{#1}{4.591353414078971}{62.435515989119295}{cyan}{yellow}{70}{#2}}
\setlength{\tabcolsep}{4.5pt}
\begin{table}[ht]
    \centering\footnotesize
    \begin{tabularx}{\columnwidth}%
    {p{2em} | C  C  C  C  C  C}
        \toprule
 & \multicolumn{2}{ c }{\textbf{Qwen3}} & \multicolumn{2}{ c }{\textbf{gemma-3}} & \multicolumn{2}{ c }{\textbf{Ministral-3}} \\
 & \textbf{base} & \textbf{inst.} & \textbf{base} & \textbf{inst.} & \textbf{base} & \textbf{inst.} \\

        \midrule
        \textbf{af} & \translationperformancebylanguageGrad{45.6}{0} & \translationperformancebylanguageGrad{45.3}{0} & \translationperformancebylanguageGrad{51.4}{1} & \translationperformancebylanguageGrad{49.3}{0} & \translationperformancebylanguageGrad{46.7}{0} & \translationperformancebylanguageGrad{29.6}{0} \\
        \textbf{am} & \translationperformancebylanguageGrad{7.9}{0} & \translationperformancebylanguageGrad{8.8}{0} & \translationperformancebylanguageGrad{24.5}{1} & \translationperformancebylanguageGrad{20.6}{0} & \translationperformancebylanguageGrad{4.6}{0} & \translationperformancebylanguageGrad{6.5}{0} \\
        \textbf{ar} & \translationperformancebylanguageGrad{39.1}{0} & \translationperformancebylanguageGrad{38.0}{0} & \translationperformancebylanguageGrad{44.2}{0} & \translationperformancebylanguageGrad{44.7}{1} & \translationperformancebylanguageGrad{40.3}{0} & \translationperformancebylanguageGrad{21.7}{0} \\
        \textbf{az} & \translationperformancebylanguageGrad{29.1}{0} & \translationperformancebylanguageGrad{28.1}{0} & \translationperformancebylanguageGrad{37.2}{1} & \translationperformancebylanguageGrad{36.8}{0} & \translationperformancebylanguageGrad{32.6}{0} & \translationperformancebylanguageGrad{24.5}{0} \\
        \textbf{bg} & \translationperformancebylanguageGrad{44.8}{0} & \translationperformancebylanguageGrad{40.4}{0} & \translationperformancebylanguageGrad{52.6}{1} & \translationperformancebylanguageGrad{52.6}{1} & \translationperformancebylanguageGrad{46.8}{0} & \translationperformancebylanguageGrad{15.1}{0} \\
        \textbf{bn} & \translationperformancebylanguageGrad{29.8}{0} & \translationperformancebylanguageGrad{26.4}{0} & \translationperformancebylanguageGrad{39.7}{1} & \translationperformancebylanguageGrad{38.8}{0} & \translationperformancebylanguageGrad{32.7}{0} & \translationperformancebylanguageGrad{18.4}{0} \\
        \textbf{cs} & \translationperformancebylanguageGrad{40.2}{0} & \translationperformancebylanguageGrad{39.9}{0} & \translationperformancebylanguageGrad{46.1}{1} & \translationperformancebylanguageGrad{46.2}{1} & \translationperformancebylanguageGrad{42.7}{0} & \translationperformancebylanguageGrad{23.7}{0} \\
        \textbf{da} & \translationperformancebylanguageGrad{46.5}{0} & \translationperformancebylanguageGrad{46.6}{0} & \translationperformancebylanguageGrad{54.3}{1} & \translationperformancebylanguageGrad{54.0}{0} & \translationperformancebylanguageGrad{48.2}{0} & \translationperformancebylanguageGrad{22.8}{0} \\
        \textbf{de} & \translationperformancebylanguageGrad{47.6}{0} & \translationperformancebylanguageGrad{47.7}{0} & \translationperformancebylanguageGrad{51.0}{0} & \translationperformancebylanguageGrad{51.9}{1} & \translationperformancebylanguageGrad{49.4}{0} & \translationperformancebylanguageGrad{23.7}{0} \\
        \textbf{el} & \translationperformancebylanguageGrad{36.7}{0} & \translationperformancebylanguageGrad{37.2}{0} & \translationperformancebylanguageGrad{44.0}{1} & \translationperformancebylanguageGrad{43.8}{0} & \translationperformancebylanguageGrad{39.6}{0} & \translationperformancebylanguageGrad{21.8}{0} \\
        \textbf{en} & \translationperformancebylanguageGrad{59.4}{0} & \translationperformancebylanguageGrad{58.4}{0} & \translationperformancebylanguageGrad{62.4}{1} & \translationperformancebylanguageGrad{60.0}{0} & \translationperformancebylanguageGrad{59.4}{0} & \translationperformancebylanguageGrad{45.6}{0} \\
        \textbf{es} & \translationperformancebylanguageGrad{45.3}{0} & \translationperformancebylanguageGrad{45.5}{0} & \translationperformancebylanguageGrad{47.5}{0} & \translationperformancebylanguageGrad{47.8}{1} & \translationperformancebylanguageGrad{45.7}{0} & \translationperformancebylanguageGrad{24.4}{0} \\
        \textbf{fa} & \translationperformancebylanguageGrad{38.3}{0} & \translationperformancebylanguageGrad{37.5}{0} & \translationperformancebylanguageGrad{45.0}{0} & \translationperformancebylanguageGrad{45.6}{1} & \translationperformancebylanguageGrad{41.9}{0} & \translationperformancebylanguageGrad{23.7}{0} \\
        \textbf{fi} & \translationperformancebylanguageGrad{39.2}{0} & \translationperformancebylanguageGrad{38.8}{0} & \translationperformancebylanguageGrad{46.6}{0} & \translationperformancebylanguageGrad{47.1}{1} & \translationperformancebylanguageGrad{42.7}{0} & \translationperformancebylanguageGrad{29.1}{0} \\
        \textbf{fr} & \translationperformancebylanguageGrad{52.2}{0} & \translationperformancebylanguageGrad{52.0}{0} & \translationperformancebylanguageGrad{55.7}{0} & \translationperformancebylanguageGrad{56.0}{1} & \translationperformancebylanguageGrad{52.4}{0} & \translationperformancebylanguageGrad{24.9}{0} \\
        \textbf{he} & \translationperformancebylanguageGrad{33.1}{0} & \translationperformancebylanguageGrad{30.9}{0} & \translationperformancebylanguageGrad{45.7}{1} & \translationperformancebylanguageGrad{45.3}{0} & \translationperformancebylanguageGrad{39.9}{0} & \translationperformancebylanguageGrad{17.4}{0} \\
        \textbf{hi} & \translationperformancebylanguageGrad{36.2}{0} & \translationperformancebylanguageGrad{34.4}{0} & \translationperformancebylanguageGrad{44.5}{0} & \translationperformancebylanguageGrad{44.7}{1} & \translationperformancebylanguageGrad{41.2}{0} & \translationperformancebylanguageGrad{12.7}{0} \\
        \textbf{hu} & \translationperformancebylanguageGrad{39.8}{0} & \translationperformancebylanguageGrad{40.1}{0} & \translationperformancebylanguageGrad{44.6}{1} & \translationperformancebylanguageGrad{44.6}{1} & \translationperformancebylanguageGrad{40.6}{0} & \translationperformancebylanguageGrad{26.3}{0} \\
        \textbf{hy} & \translationperformancebylanguageGrad{33.8}{0} & \translationperformancebylanguageGrad{26.9}{0} & \translationperformancebylanguageGrad{43.4}{1} & \translationperformancebylanguageGrad{40.4}{0} & \translationperformancebylanguageGrad{39.5}{0} & \translationperformancebylanguageGrad{25.4}{0} \\
        \textbf{id} & \translationperformancebylanguageGrad{52.8}{0} & \translationperformancebylanguageGrad{52.4}{0} & \translationperformancebylanguageGrad{56.6}{0} & \translationperformancebylanguageGrad{57.3}{1} & \translationperformancebylanguageGrad{50.1}{0} & \translationperformancebylanguageGrad{25.4}{0} \\
        \textbf{is} & \translationperformancebylanguageGrad{27.9}{0} & \translationperformancebylanguageGrad{27.5}{0} & \translationperformancebylanguageGrad{39.8}{1} & \translationperformancebylanguageGrad{36.5}{0} & \translationperformancebylanguageGrad{34.0}{0} & \translationperformancebylanguageGrad{23.7}{0} \\
        \textbf{it} & \translationperformancebylanguageGrad{46.0}{0} & \translationperformancebylanguageGrad{46.0}{0} & \translationperformancebylanguageGrad{49.7}{0} & \translationperformancebylanguageGrad{50.1}{1} & \translationperformancebylanguageGrad{47.1}{0} & \translationperformancebylanguageGrad{18.7}{0} \\
        \textbf{ja} & \translationperformancebylanguageGrad{23.8}{0} & \translationperformancebylanguageGrad{23.9}{0} & \translationperformancebylanguageGrad{27.0}{0} & \translationperformancebylanguageGrad{27.3}{1} & \translationperformancebylanguageGrad{25.2}{0} & \translationperformancebylanguageGrad{14.3}{0} \\
        \textbf{jv} & \translationperformancebylanguageGrad{35.8}{0} & \translationperformancebylanguageGrad{35.6}{0} & \translationperformancebylanguageGrad{43.6}{1} & \translationperformancebylanguageGrad{41.8}{0} & \translationperformancebylanguageGrad{25.7}{0} & \translationperformancebylanguageGrad{23.7}{0} \\
        \textbf{ka} & \translationperformancebylanguageGrad{31.5}{0} & \translationperformancebylanguageGrad{29.9}{0} & \translationperformancebylanguageGrad{40.4}{1} & \translationperformancebylanguageGrad{38.8}{0} & \translationperformancebylanguageGrad{35.7}{0} & \translationperformancebylanguageGrad{26.5}{0} \\
        \textbf{km} & \translationperformancebylanguageGrad{21.7}{0} & \translationperformancebylanguageGrad{18.3}{0} & \translationperformancebylanguageGrad{28.4}{1} & \translationperformancebylanguageGrad{26.4}{0} & \translationperformancebylanguageGrad{9.1}{0} & \translationperformancebylanguageGrad{12.4}{0} \\
        \textbf{kn} & \translationperformancebylanguageGrad{27.3}{0} & \translationperformancebylanguageGrad{23.1}{0} & \translationperformancebylanguageGrad{41.4}{1} & \translationperformancebylanguageGrad{39.2}{0} & \translationperformancebylanguageGrad{34.8}{0} & \translationperformancebylanguageGrad{19.8}{0} \\
        \textbf{ko} & \translationperformancebylanguageGrad{25.4}{0} & \translationperformancebylanguageGrad{22.9}{0} & \translationperformancebylanguageGrad{29.0}{0} & \translationperformancebylanguageGrad{29.8}{1} & \translationperformancebylanguageGrad{25.6}{0} & \translationperformancebylanguageGrad{15.0}{0} \\
        \textbf{lt} & \translationperformancebylanguageGrad{38.6}{0} & \translationperformancebylanguageGrad{37.5}{0} & \translationperformancebylanguageGrad{46.3}{1} & \translationperformancebylanguageGrad{44.9}{0} & \translationperformancebylanguageGrad{39.8}{0} & \translationperformancebylanguageGrad{20.9}{0} \\
        \textbf{ml} & \translationperformancebylanguageGrad{26.2}{0} & \translationperformancebylanguageGrad{24.6}{0} & \translationperformancebylanguageGrad{41.5}{1} & \translationperformancebylanguageGrad{36.6}{0} & \translationperformancebylanguageGrad{31.5}{0} & \translationperformancebylanguageGrad{21.0}{0} \\
        \textbf{mr} & \translationperformancebylanguageGrad{26.8}{0} & \translationperformancebylanguageGrad{27.4}{0} & \translationperformancebylanguageGrad{38.3}{0} & \translationperformancebylanguageGrad{38.6}{1} & \translationperformancebylanguageGrad{33.2}{0} & \translationperformancebylanguageGrad{10.7}{0} \\
        \textbf{nl} & \translationperformancebylanguageGrad{44.3}{0} & \translationperformancebylanguageGrad{44.5}{0} & \translationperformancebylanguageGrad{48.4}{1} & \translationperformancebylanguageGrad{48.5}{1} & \translationperformancebylanguageGrad{43.8}{0} & \translationperformancebylanguageGrad{18.9}{0} \\
        \textbf{pl} & \translationperformancebylanguageGrad{38.4}{0} & \translationperformancebylanguageGrad{38.8}{0} & \translationperformancebylanguageGrad{42.8}{0} & \translationperformancebylanguageGrad{43.5}{1} & \translationperformancebylanguageGrad{40.3}{0} & \translationperformancebylanguageGrad{26.0}{0} \\
        \textbf{pt} & \translationperformancebylanguageGrad{51.7}{0} & \translationperformancebylanguageGrad{51.0}{0} & \translationperformancebylanguageGrad{55.4}{1} & \translationperformancebylanguageGrad{55.3}{0} & \translationperformancebylanguageGrad{52.3}{0} & \translationperformancebylanguageGrad{29.6}{0} \\
        \textbf{ru} & \translationperformancebylanguageGrad{43.8}{0} & \translationperformancebylanguageGrad{42.9}{0} & \translationperformancebylanguageGrad{47.5}{0} & \translationperformancebylanguageGrad{47.8}{1} & \translationperformancebylanguageGrad{45.2}{0} & \translationperformancebylanguageGrad{19.9}{0} \\
        \textbf{sw} & \translationperformancebylanguageGrad{28.9}{0} & \translationperformancebylanguageGrad{28.0}{0} & \translationperformancebylanguageGrad{49.0}{1} & \translationperformancebylanguageGrad{47.0}{0} & \translationperformancebylanguageGrad{34.9}{0} & \translationperformancebylanguageGrad{26.6}{0} \\
        \textbf{ta} & \translationperformancebylanguageGrad{30.6}{0} & \translationperformancebylanguageGrad{32.9}{0} & \translationperformancebylanguageGrad{42.2}{0} & \translationperformancebylanguageGrad{44.9}{1} & \translationperformancebylanguageGrad{35.7}{0} & \translationperformancebylanguageGrad{22.3}{0} \\
        \textbf{te} & \translationperformancebylanguageGrad{29.8}{0} & \translationperformancebylanguageGrad{30.4}{0} & \translationperformancebylanguageGrad{40.0}{0} & \translationperformancebylanguageGrad{43.0}{1} & \translationperformancebylanguageGrad{31.6}{0} & \translationperformancebylanguageGrad{15.3}{0} \\
        \textbf{th} & \translationperformancebylanguageGrad{35.4}{0} & \translationperformancebylanguageGrad{35.2}{0} & \translationperformancebylanguageGrad{39.7}{0} & \translationperformancebylanguageGrad{40.2}{1} & \translationperformancebylanguageGrad{35.2}{0} & \translationperformancebylanguageGrad{25.2}{0} \\
        \textbf{tl} & \translationperformancebylanguageGrad{45.2}{0} & \translationperformancebylanguageGrad{44.0}{0} & \translationperformancebylanguageGrad{52.0}{0} & \translationperformancebylanguageGrad{52.3}{1} & \translationperformancebylanguageGrad{45.0}{0} & \translationperformancebylanguageGrad{23.9}{0} \\
        \textbf{tr} & \translationperformancebylanguageGrad{41.7}{0} & \translationperformancebylanguageGrad{39.0}{0} & \translationperformancebylanguageGrad{47.7}{0} & \translationperformancebylanguageGrad{47.8}{1} & \translationperformancebylanguageGrad{43.3}{0} & \translationperformancebylanguageGrad{23.6}{0} \\
        \textbf{ur} & \translationperformancebylanguageGrad{27.7}{0} & \translationperformancebylanguageGrad{28.7}{0} & \translationperformancebylanguageGrad{39.2}{1} & \translationperformancebylanguageGrad{38.2}{0} & \translationperformancebylanguageGrad{34.6}{0} & \translationperformancebylanguageGrad{17.9}{0} \\
        \textbf{vi} & \translationperformancebylanguageGrad{48.2}{0} & \translationperformancebylanguageGrad{46.2}{0} & \translationperformancebylanguageGrad{51.6}{1} & \translationperformancebylanguageGrad{51.6}{1} & \translationperformancebylanguageGrad{47.3}{0} & \translationperformancebylanguageGrad{24.5}{0} \\
        \textbf{zh} & \translationperformancebylanguageGrad{19.5}{0} & \translationperformancebylanguageGrad{19.3}{0} & \translationperformancebylanguageGrad{21.5}{1} & \translationperformancebylanguageGrad{21.4}{1} & \translationperformancebylanguageGrad{20.5}{0} & \translationperformancebylanguageGrad{8.7}{0} \\
        \bottomrule
    \end{tabularx}
    \caption{Average \textbf{Flores} translation performance (\textsc{chrF}; over all source languages) for the displayed target languages. Values in cent.}
    \label{tab:translation-performance-by-language}
\end{table}

%% file: latex/tables/translation_performance_by_language-bouquet.tex
\setlength{\tabcolsep}{4.5pt}
\begin{table}[ht]
    \centering\footnotesize
    \begin{tabularx}{\columnwidth}%
    {p{2em} | C  C  C  C  C  C}
        \toprule
 & \multicolumn{2}{ c }{\textbf{Qwen3}} & \multicolumn{2}{ c }{\textbf{gemma-3}} & \multicolumn{2}{ c }{\textbf{Ministral-3}} \\
 & \textbf{base} & \textbf{inst.} & \textbf{base} & \textbf{inst.} & \textbf{base} & \textbf{inst.} \\

        \midrule
        \textbf{af} & \translationperformancebylanguageGrad{46.3}{0} & \translationperformancebylanguageGrad{43.8}{0} & \translationperformancebylanguageGrad{53.7}{1} & \translationperformancebylanguageGrad{49.9}{0} & \translationperformancebylanguageGrad{45.6}{0} & \translationperformancebylanguageGrad{21.5}{0} \\
        \textbf{am} & \translationperformancebylanguageGrad{10.6}{0} & \translationperformancebylanguageGrad{9.8}{0} & \translationperformancebylanguageGrad{26.3}{1} & \translationperformancebylanguageGrad{20.5}{0} & \translationperformancebylanguageGrad{6.3}{0} & \translationperformancebylanguageGrad{5.8}{0} \\
        \textbf{ar} & \translationperformancebylanguageGrad{27.1}{0} & \translationperformancebylanguageGrad{25.8}{0} & \translationperformancebylanguageGrad{36.3}{1} & \translationperformancebylanguageGrad{35.4}{0} & \translationperformancebylanguageGrad{28.7}{0} & \translationperformancebylanguageGrad{13.5}{0} \\
        \textbf{az} & \translationperformancebylanguageGrad{30.7}{0} & \translationperformancebylanguageGrad{27.1}{0} & \translationperformancebylanguageGrad{42.8}{1} & \translationperformancebylanguageGrad{40.1}{0} & \translationperformancebylanguageGrad{33.6}{0} & \translationperformancebylanguageGrad{19.8}{0} \\
        \textbf{bg} & \translationperformancebylanguageGrad{47.5}{0} & \translationperformancebylanguageGrad{41.3}{0} & \translationperformancebylanguageGrad{56.1}{1} & \translationperformancebylanguageGrad{52.7}{0} & \translationperformancebylanguageGrad{48.4}{0} & \translationperformancebylanguageGrad{14.0}{0} \\
        \textbf{bn} & \translationperformancebylanguageGrad{38.9}{0} & \translationperformancebylanguageGrad{30.0}{0} & \translationperformancebylanguageGrad{48.0}{1} & \translationperformancebylanguageGrad{39.8}{0} & \translationperformancebylanguageGrad{37.0}{0} & \translationperformancebylanguageGrad{9.0}{0} \\
        \textbf{cs} & \translationperformancebylanguageGrad{43.0}{0} & \translationperformancebylanguageGrad{39.6}{0} & \translationperformancebylanguageGrad{50.6}{1} & \translationperformancebylanguageGrad{47.9}{0} & \translationperformancebylanguageGrad{44.9}{0} & \translationperformancebylanguageGrad{28.2}{0} \\
        \textbf{da} & \translationperformancebylanguageGrad{46.3}{0} & \translationperformancebylanguageGrad{43.2}{0} & \translationperformancebylanguageGrad{53.1}{1} & \translationperformancebylanguageGrad{50.2}{0} & \translationperformancebylanguageGrad{47.6}{0} & \translationperformancebylanguageGrad{17.4}{0} \\
        \textbf{de} & \translationperformancebylanguageGrad{49.9}{0} & \translationperformancebylanguageGrad{47.7}{0} & \translationperformancebylanguageGrad{53.8}{1} & \translationperformancebylanguageGrad{51.2}{0} & \translationperformancebylanguageGrad{50.8}{0} & \translationperformancebylanguageGrad{17.6}{0} \\
        \textbf{el} & \translationperformancebylanguageGrad{42.8}{0} & \translationperformancebylanguageGrad{39.3}{0} & \translationperformancebylanguageGrad{52.8}{1} & \translationperformancebylanguageGrad{48.1}{0} & \translationperformancebylanguageGrad{46.1}{0} & \translationperformancebylanguageGrad{17.8}{0} \\
        \textbf{en} & \translationperformancebylanguageGrad{60.1}{0} & \translationperformancebylanguageGrad{56.0}{0} & \translationperformancebylanguageGrad{63.0}{1} & \translationperformancebylanguageGrad{56.9}{0} & \translationperformancebylanguageGrad{59.2}{0} & \translationperformancebylanguageGrad{25.3}{0} \\
        \textbf{es} & \translationperformancebylanguageGrad{52.3}{0} & \translationperformancebylanguageGrad{49.2}{0} & \translationperformancebylanguageGrad{55.4}{1} & \translationperformancebylanguageGrad{52.5}{0} & \translationperformancebylanguageGrad{52.0}{0} & \translationperformancebylanguageGrad{26.6}{0} \\
        \textbf{fa} & \translationperformancebylanguageGrad{35.3}{0} & \translationperformancebylanguageGrad{32.7}{0} & \translationperformancebylanguageGrad{42.4}{1} & \translationperformancebylanguageGrad{40.4}{0} & \translationperformancebylanguageGrad{37.3}{0} & \translationperformancebylanguageGrad{15.0}{0} \\
        \textbf{fi} & \translationperformancebylanguageGrad{39.1}{0} & \translationperformancebylanguageGrad{33.4}{0} & \translationperformancebylanguageGrad{47.6}{1} & \translationperformancebylanguageGrad{46.1}{0} & \translationperformancebylanguageGrad{41.0}{0} & \translationperformancebylanguageGrad{20.5}{0} \\
        \textbf{fr} & \translationperformancebylanguageGrad{52.6}{0} & \translationperformancebylanguageGrad{48.9}{0} & \translationperformancebylanguageGrad{56.1}{1} & \translationperformancebylanguageGrad{51.5}{0} & \translationperformancebylanguageGrad{53.2}{0} & \translationperformancebylanguageGrad{27.0}{0} \\
        \textbf{he} & \translationperformancebylanguageGrad{36.8}{0} & \translationperformancebylanguageGrad{30.3}{0} & \translationperformancebylanguageGrad{47.1}{1} & \translationperformancebylanguageGrad{45.4}{0} & \translationperformancebylanguageGrad{38.6}{0} & \translationperformancebylanguageGrad{17.3}{0} \\
        \textbf{hi} & \translationperformancebylanguageGrad{38.0}{0} & \translationperformancebylanguageGrad{32.9}{0} & \translationperformancebylanguageGrad{43.6}{1} & \translationperformancebylanguageGrad{38.1}{0} & \translationperformancebylanguageGrad{39.9}{0} & \translationperformancebylanguageGrad{8.0}{0} \\
        \textbf{hu} & \translationperformancebylanguageGrad{44.1}{0} & \translationperformancebylanguageGrad{40.3}{0} & \translationperformancebylanguageGrad{49.6}{1} & \translationperformancebylanguageGrad{46.5}{0} & \translationperformancebylanguageGrad{43.6}{0} & \translationperformancebylanguageGrad{20.0}{0} \\
        \textbf{hy} & \translationperformancebylanguageGrad{35.8}{0} & \translationperformancebylanguageGrad{19.0}{0} & \translationperformancebylanguageGrad{44.7}{1} & \translationperformancebylanguageGrad{38.8}{0} & \translationperformancebylanguageGrad{36.3}{0} & \translationperformancebylanguageGrad{21.3}{0} \\
        \textbf{id} & \translationperformancebylanguageGrad{50.9}{0} & \translationperformancebylanguageGrad{46.1}{0} & \translationperformancebylanguageGrad{53.9}{1} & \translationperformancebylanguageGrad{51.1}{0} & \translationperformancebylanguageGrad{48.5}{0} & \translationperformancebylanguageGrad{13.6}{0} \\
        \textbf{is} & \translationperformancebylanguageGrad{30.5}{0} & \translationperformancebylanguageGrad{27.9}{0} & \translationperformancebylanguageGrad{44.2}{1} & \translationperformancebylanguageGrad{40.3}{0} & \translationperformancebylanguageGrad{34.2}{0} & \translationperformancebylanguageGrad{17.4}{0} \\
        \textbf{it} & \translationperformancebylanguageGrad{54.4}{0} & \translationperformancebylanguageGrad{51.1}{0} & \translationperformancebylanguageGrad{59.0}{1} & \translationperformancebylanguageGrad{56.2}{0} & \translationperformancebylanguageGrad{55.3}{0} & \translationperformancebylanguageGrad{23.9}{0} \\
        \textbf{ja} & \translationperformancebylanguageGrad{23.3}{0} & \translationperformancebylanguageGrad{20.8}{0} & \translationperformancebylanguageGrad{25.8}{1} & \translationperformancebylanguageGrad{24.5}{0} & \translationperformancebylanguageGrad{24.0}{0} & \translationperformancebylanguageGrad{12.8}{0} \\
        \textbf{jv} & \translationperformancebylanguageGrad{36.4}{0} & \translationperformancebylanguageGrad{32.6}{0} & \translationperformancebylanguageGrad{47.1}{1} & \translationperformancebylanguageGrad{41.2}{0} & \translationperformancebylanguageGrad{26.5}{0} & \translationperformancebylanguageGrad{18.4}{0} \\
        \textbf{ka} & \translationperformancebylanguageGrad{32.8}{0} & \translationperformancebylanguageGrad{28.3}{0} & \translationperformancebylanguageGrad{42.5}{1} & \translationperformancebylanguageGrad{37.5}{0} & \translationperformancebylanguageGrad{32.9}{0} & \translationperformancebylanguageGrad{23.7}{0} \\
        \textbf{km} & \translationperformancebylanguageGrad{31.9}{0} & \translationperformancebylanguageGrad{23.0}{0} & \translationperformancebylanguageGrad{38.7}{1} & \translationperformancebylanguageGrad{32.3}{0} & \translationperformancebylanguageGrad{14.6}{0} & \translationperformancebylanguageGrad{12.6}{0} \\
        \textbf{kn} & \translationperformancebylanguageGrad{32.4}{0} & \translationperformancebylanguageGrad{22.5}{0} & \translationperformancebylanguageGrad{41.5}{1} & \translationperformancebylanguageGrad{32.5}{0} & \translationperformancebylanguageGrad{34.5}{0} & \translationperformancebylanguageGrad{9.1}{0} \\
        \textbf{ko} & \translationperformancebylanguageGrad{28.5}{0} & \translationperformancebylanguageGrad{23.9}{0} & \translationperformancebylanguageGrad{33.0}{1} & \translationperformancebylanguageGrad{31.6}{0} & \translationperformancebylanguageGrad{28.9}{0} & \translationperformancebylanguageGrad{12.7}{0} \\
        \textbf{lt} & \translationperformancebylanguageGrad{42.6}{0} & \translationperformancebylanguageGrad{39.6}{0} & \translationperformancebylanguageGrad{51.7}{1} & \translationperformancebylanguageGrad{47.5}{0} & \translationperformancebylanguageGrad{42.7}{0} & \translationperformancebylanguageGrad{25.8}{0} \\
        \textbf{ml} & \translationperformancebylanguageGrad{31.2}{0} & \translationperformancebylanguageGrad{25.3}{0} & \translationperformancebylanguageGrad{38.4}{1} & \translationperformancebylanguageGrad{34.0}{0} & \translationperformancebylanguageGrad{30.2}{0} & \translationperformancebylanguageGrad{14.0}{0} \\
        \textbf{mr} & \translationperformancebylanguageGrad{31.5}{0} & \translationperformancebylanguageGrad{26.6}{0} & \translationperformancebylanguageGrad{38.5}{1} & \translationperformancebylanguageGrad{36.5}{0} & \translationperformancebylanguageGrad{33.3}{0} & \translationperformancebylanguageGrad{7.9}{0} \\
        \textbf{nl} & \translationperformancebylanguageGrad{53.0}{0} & \translationperformancebylanguageGrad{48.5}{0} & \translationperformancebylanguageGrad{59.0}{1} & \translationperformancebylanguageGrad{54.9}{0} & \translationperformancebylanguageGrad{53.6}{0} & \translationperformancebylanguageGrad{24.1}{0} \\
        \textbf{pl} & \translationperformancebylanguageGrad{47.9}{0} & \translationperformancebylanguageGrad{42.9}{0} & \translationperformancebylanguageGrad{55.4}{1} & \translationperformancebylanguageGrad{51.5}{0} & \translationperformancebylanguageGrad{50.3}{0} & \translationperformancebylanguageGrad{21.5}{0} \\
        \textbf{pt} & \translationperformancebylanguageGrad{55.7}{0} & \translationperformancebylanguageGrad{52.7}{0} & \translationperformancebylanguageGrad{60.4}{1} & \translationperformancebylanguageGrad{56.6}{0} & \translationperformancebylanguageGrad{56.6}{0} & \translationperformancebylanguageGrad{27.6}{0} \\
        \textbf{ru} & \translationperformancebylanguageGrad{47.9}{0} & \translationperformancebylanguageGrad{42.0}{0} & \translationperformancebylanguageGrad{52.2}{1} & \translationperformancebylanguageGrad{48.1}{0} & \translationperformancebylanguageGrad{49.2}{0} & \translationperformancebylanguageGrad{12.6}{0} \\
        \textbf{sw} & \translationperformancebylanguageGrad{27.7}{0} & \translationperformancebylanguageGrad{23.2}{0} & \translationperformancebylanguageGrad{50.5}{1} & \translationperformancebylanguageGrad{46.6}{0} & \translationperformancebylanguageGrad{31.5}{0} & \translationperformancebylanguageGrad{19.8}{0} \\
        \textbf{ta} & \translationperformancebylanguageGrad{35.4}{0} & \translationperformancebylanguageGrad{27.6}{0} & \translationperformancebylanguageGrad{42.5}{1} & \translationperformancebylanguageGrad{41.2}{0} & \translationperformancebylanguageGrad{36.8}{0} & \translationperformancebylanguageGrad{11.9}{0} \\
        \textbf{te} & \translationperformancebylanguageGrad{34.2}{0} & \translationperformancebylanguageGrad{28.8}{0} & \translationperformancebylanguageGrad{42.4}{1} & \translationperformancebylanguageGrad{41.2}{0} & \translationperformancebylanguageGrad{34.5}{0} & \translationperformancebylanguageGrad{10.6}{0} \\
        \textbf{th} & \translationperformancebylanguageGrad{38.8}{0} & \translationperformancebylanguageGrad{34.0}{0} & \translationperformancebylanguageGrad{43.3}{1} & \translationperformancebylanguageGrad{41.2}{0} & \translationperformancebylanguageGrad{36.4}{0} & \translationperformancebylanguageGrad{19.8}{0} \\
        \textbf{tl} & \translationperformancebylanguageGrad{47.1}{0} & \translationperformancebylanguageGrad{40.2}{0} & \translationperformancebylanguageGrad{56.8}{1} & \translationperformancebylanguageGrad{50.1}{0} & \translationperformancebylanguageGrad{43.4}{0} & \translationperformancebylanguageGrad{17.9}{0} \\
        \textbf{tr} & \translationperformancebylanguageGrad{45.8}{0} & \translationperformancebylanguageGrad{38.7}{0} & \translationperformancebylanguageGrad{53.0}{1} & \translationperformancebylanguageGrad{50.2}{0} & \translationperformancebylanguageGrad{45.9}{0} & \translationperformancebylanguageGrad{19.9}{0} \\
        \textbf{ur} & \translationperformancebylanguageGrad{37.8}{0} & \translationperformancebylanguageGrad{32.8}{0} & \translationperformancebylanguageGrad{49.6}{1} & \translationperformancebylanguageGrad{45.3}{0} & \translationperformancebylanguageGrad{39.5}{0} & \translationperformancebylanguageGrad{14.1}{0} \\
        \textbf{vi} & \translationperformancebylanguageGrad{55.7}{0} & \translationperformancebylanguageGrad{43.0}{0} & \translationperformancebylanguageGrad{59.8}{1} & \translationperformancebylanguageGrad{54.4}{0} & \translationperformancebylanguageGrad{53.2}{0} & \translationperformancebylanguageGrad{25.2}{0} \\
        \textbf{zh} & \translationperformancebylanguageGrad{24.5}{0} & \translationperformancebylanguageGrad{20.2}{0} & \translationperformancebylanguageGrad{26.6}{1} & \translationperformancebylanguageGrad{23.4}{0} & \translationperformancebylanguageGrad{25.1}{0} & \translationperformancebylanguageGrad{10.1}{0} \\
        \bottomrule
    \end{tabularx}
    \caption{Average \textbf{BOUQuET} translation performance (\textsc{chrF}; over all source languages) for the displayed target languages. Values in cent.}
    \label{tab:translation-performance-by-language-bouquet}
\end{table}

%% file: latex/tables/corr_both_with_chrf_combined.tex
\setlength{\tabcolsep}{3pt}\begin{table}[ht]
\centering\footnotesize
\begin{tabularx}{\columnwidth}%
{X | c  c  c  c  c  c | c}\toprule
\multicolumn{8}{c}{\textbf{Qwen3-14B-Base}} \\
\midrule
         & \textbf{cos} & \textbf{ANC} & \textbf{abs} & \textbf{dist} & \textbf{ratio} & \textbf{Dali} & \textbf{Eflo} \\
        \midrule
        \textbf{fewshot} & \percentGrad{43}{0}\hspace{0.6em}/20 & \percentGrad{56}{0}\hspace{0.6em}/48 & \percentGrad{82}{1}\hspace{0.6em}/56 & \percentGrad{82}{1}\hspace{0.6em}/56 & \percentGrad{82}{1}\hspace{0.6em}/56 & --/-- & \percentGrad{62}{2} \\
        \textbf{last} & \percentGrad{34}{0}\hspace{0.6em}/16 & \percentGrad{57}{0}\hspace{0.6em}/48 & \percentGrad{72}{0}\hspace{0.6em}/52 & \percentGrad{72}{0}\hspace{0.6em}/52 & \percentGrad{72}{0}\hspace{0.6em}/53 & \percentGrad{17}{0}\hspace{0.6em}/14 & \percentGrad{62}{2} \\
        \textbf{p-wm} & \percentGrad{77}{0}\hspace{0.6em}/36 & \percentGrad{82}{1}\hspace{0.6em}/\textbf{59} & \percentGrad{72}{0}\hspace{0.6em}/\textbf{59} & \percentGrad{72}{0}\hspace{0.6em}/\textbf{59} & \percentGrad{72}{0}\hspace{0.6em}/\textbf{59} & --/-- & \percentGrad{62}{0}\hspace{0.6em}/15 \\
        \textbf{mean} & \percentGrad{61}{0}\hspace{0.6em}/30 & \percentGrad{81}{0}\hspace{0.6em}/57 & \percentGrad{77}{0}\hspace{0.6em}/28 & \percentGrad{77}{0}\hspace{0.6em}/28 & \percentGrad{76}{0}\hspace{0.6em}/28 & --/-- & \percentGrad{62}{2} \\
        \textbf{prompt} & --/-- & \percentGrad{80}{0}\hspace{0.6em}/58 & \percentGrad{67}{0}\hspace{0.6em}/58 & \percentGrad{67}{0}\hspace{0.6em}/58 & \percentGrad{67}{0}\hspace{0.6em}/58 & --/-- & \percentGrad{62}{2} \\
\midrule
\multicolumn{8}{c}{\textbf{Qwen3-14B}} \\
\midrule
         & \textbf{cos} & \textbf{ANC} & \textbf{abs} & \textbf{dist} & \textbf{ratio} & \textbf{Dali} & \textbf{Eflo} \\
        \midrule
        \textbf{fewshot} & --/-- & \percentGrad{80}{0}\hspace{0.6em}/50 & \percentGrad{71}{0}\hspace{0.6em}/43 & \percentGrad{71}{0}\hspace{0.6em}/43 & \percentGrad{71}{0}\hspace{0.6em}/43 & --/-- & \percentGrad{50}{2} \\
        \textbf{last} & \percentGrad{34}{0}\hspace{0.6em}/18 & \percentGrad{54}{0}\hspace{0.6em}/44 & \percentGrad{69}{0}\hspace{0.6em}/45 & \percentGrad{69}{0}\hspace{0.6em}/45 & \percentGrad{69}{0}\hspace{0.6em}/45 & \percentGrad{69}{0}\hspace{0.6em}/58 & \percentGrad{50}{2} \\
        \textbf{p-wm} & \percentGrad{68}{0}\hspace{0.6em}/30 & \percentGrad{80}{0}\hspace{0.6em}/58 & \percentGrad{74}{0}\hspace{0.6em}/58 & \percentGrad{74}{0}\hspace{0.6em}/58 & \percentGrad{74}{0}\hspace{0.6em}/58 & --/-- & \percentGrad{50}{0}\hspace{0.6em}/7 \\
        \textbf{mean} & \percentGrad{52}{0}\hspace{0.6em}/29 & \percentGrad{81}{0}\hspace{0.6em}/57 & \percentGrad{73}{0}\hspace{0.6em}/27 & \percentGrad{73}{0}\hspace{0.6em}/27 & \percentGrad{73}{0}\hspace{0.6em}/27 & --/-- & \percentGrad{50}{2} \\
        \textbf{prompt} & --/-- & \percentGrad{82}{1}\hspace{0.6em}/\textbf{63} & \percentGrad{63}{0}\hspace{0.6em}/55 & \percentGrad{63}{0}\hspace{0.6em}/55 & \percentGrad{63}{0}\hspace{0.6em}/55 & --/-- & \percentGrad{50}{2} \\
\midrule
\multicolumn{8}{c}{\textbf{gemma-3-12b-pt}} \\
\midrule
         & \textbf{cos} & \textbf{ANC} & \textbf{abs} & \textbf{dist} & \textbf{ratio} & \textbf{Dali} & \textbf{Eflo} \\
        \midrule
        \textbf{fewshot} & \percentGrad{43}{0}\hspace{0.6em}/28 & \percentGrad{57}{0}\hspace{0.6em}/42 & \percentGrad{51}{0}\hspace{0.6em}/22 & \percentGrad{51}{0}\hspace{0.6em}/22 & \percentGrad{51}{0}\hspace{0.6em}/22 & --/-- & \percentGrad{65}{2} \\
        \textbf{last} & \percentGrad{22}{0}\hspace{0.6em}/10 & \percentGrad{57}{0}\hspace{0.6em}/50 & \percentGrad{64}{0}\hspace{0.6em}/58 & \percentGrad{64}{0}\hspace{0.6em}/58 & \percentGrad{64}{0}\hspace{0.6em}/58 & \percentGrad{53}{0}\hspace{0.6em}/50 & \percentGrad{65}{2} \\
        \textbf{p-wm} & \percentGrad{57}{0}\hspace{0.6em}/53 & \percentGrad{67}{1}\hspace{0.6em}/\textbf{70} & \percentGrad{54}{0}\hspace{0.6em}/52 & \percentGrad{54}{0}\hspace{0.6em}/52 & \percentGrad{54}{0}\hspace{0.6em}/52 & --/-- & \percentGrad{65}{0}\hspace{0.6em}/53 \\
        \textbf{mean} & --/20 & \percentGrad{67}{1}\hspace{0.6em}/68 & \percentGrad{49}{0}\hspace{0.6em}/46 & \percentGrad{49}{0}\hspace{0.6em}/46 & \percentGrad{49}{0}\hspace{0.6em}/46 & --/-- & \percentGrad{65}{2} \\
        \textbf{prompt} & --/-- & \percentGrad{60}{0}\hspace{0.6em}/66 & \percentGrad{50}{0}\hspace{0.6em}/54 & \percentGrad{50}{0}\hspace{0.6em}/54 & \percentGrad{51}{0}\hspace{0.6em}/54 & --/-- & \percentGrad{65}{2} \\
\midrule
\multicolumn{8}{c}{\textbf{gemma-3-12b-it}} \\
\midrule
         & \textbf{cos} & \textbf{ANC} & \textbf{abs} & \textbf{dist} & \textbf{ratio} & \textbf{Dali} & \textbf{Eflo} \\
        \midrule
        \textbf{fewshot} & \percentGrad{20}{0}\hspace{0.6em}/25 & \percentGrad{78}{1}\hspace{0.6em}/70 & \percentGrad{71}{0}\hspace{0.6em}/53 & \percentGrad{71}{0}\hspace{0.6em}/53 & \percentGrad{71}{0}\hspace{0.6em}/53 & --/-- & \percentGrad{68}{2} \\
        \textbf{last} & \percentGrad{52}{0}\hspace{0.6em}/44 & \percentGrad{71}{0}\hspace{0.6em}/51 & \percentGrad{71}{0}\hspace{0.6em}/54 & \percentGrad{71}{0}\hspace{0.6em}/54 & \percentGrad{71}{0}\hspace{0.6em}/54 & \percentGrad{26}{0}\hspace{0.6em}/26 & \percentGrad{68}{2} \\
        \textbf{p-wm} & \percentGrad{66}{0}\hspace{0.6em}/53 & \percentGrad{77}{0}\hspace{0.6em}/70 & \percentGrad{58}{0}\hspace{0.6em}/48 & \percentGrad{58}{0}\hspace{0.6em}/48 & \percentGrad{58}{0}\hspace{0.6em}/48 & --/-- & \percentGrad{68}{0}\hspace{0.6em}/48 \\
        \textbf{mean} & --/27 & \percentGrad{77}{0}\hspace{0.6em}/68 & \percentGrad{55}{0}\hspace{0.6em}/43 & \percentGrad{55}{0}\hspace{0.6em}/43 & \percentGrad{55}{0}\hspace{0.6em}/42 & --/-- & \percentGrad{68}{2} \\
        \textbf{prompt} & --/-- & \percentGrad{73}{0}\hspace{0.6em}/\textbf{76} & \percentGrad{62}{0}\hspace{0.6em}/63 & \percentGrad{62}{0}\hspace{0.6em}/63 & \percentGrad{62}{0}\hspace{0.6em}/63 & --/-- & \percentGrad{68}{2} \\
\midrule
\multicolumn{8}{c}{\textbf{Ministral-3-14B-Base-2512}} \\
\midrule
         & \textbf{cos} & \textbf{ANC} & \textbf{abs} & \textbf{dist} & \textbf{ratio} & \textbf{Dali} & \textbf{Eflo} \\
        \midrule
        \textbf{fewshot} & \percentGrad{56}{0}\hspace{0.6em}/43 & \percentGrad{77}{0}\hspace{0.6em}/74 & \percentGrad{88}{1}\hspace{0.6em}/77 & \percentGrad{88}{1}\hspace{0.6em}/77 & \percentGrad{88}{1}\hspace{0.6em}/77 & --/-- & \percentGrad{63}{2} \\
        \textbf{last} & \percentGrad{82}{0}\hspace{0.6em}/64 & \percentGrad{79}{0}\hspace{0.6em}/70 & \percentGrad{83}{0}\hspace{0.6em}/\textbf{81} & \percentGrad{83}{0}\hspace{0.6em}/\textbf{81} & \percentGrad{83}{0}\hspace{0.6em}/\textbf{81} & \percentGrad{85}{0}\hspace{0.6em}/78 & \percentGrad{63}{2} \\
        \textbf{p-wm} & \percentGrad{84}{0}\hspace{0.6em}/72 & \percentGrad{87}{0}\hspace{0.6em}/77 & \percentGrad{86}{0}\hspace{0.6em}/75 & \percentGrad{86}{0}\hspace{0.6em}/75 & \percentGrad{86}{0}\hspace{0.6em}/75 & --/-- & \percentGrad{63}{0}\hspace{0.6em}/30 \\
        \textbf{mean} & \percentGrad{67}{0}\hspace{0.6em}/60 & \percentGrad{87}{0}\hspace{0.6em}/77 & \percentGrad{84}{0}\hspace{0.6em}/64 & \percentGrad{84}{0}\hspace{0.6em}/64 & \percentGrad{84}{0}\hspace{0.6em}/64 & --/-- & \percentGrad{63}{2} \\
        \textbf{prompt} & --/-- & \percentGrad{87}{0}\hspace{0.6em}/78 & \percentGrad{81}{0}\hspace{0.6em}/\textbf{81} & \percentGrad{81}{0}\hspace{0.6em}/\textbf{81} & \percentGrad{81}{0}\hspace{0.6em}/\textbf{81} & --/-- & \percentGrad{63}{2} \\
\midrule
\multicolumn{8}{c}{\textbf{Ministral-3-14B-Instruct-2512}} \\
\midrule
         & \textbf{cos} & \textbf{ANC} & \textbf{abs} & \textbf{dist} & \textbf{ratio} & \textbf{Dali} & \textbf{Eflo} \\
        \midrule
        \textbf{fewshot} & --/-- & \percentGrad{56}{0}\hspace{0.6em}/32 & \percentGrad{35}{0}\hspace{0.6em}/28 & \percentGrad{35}{0}\hspace{0.6em}/28 & \percentGrad{35}{0}\hspace{0.6em}/28 & --/-- & \percentGrad{62}{2} \\
        \textbf{last} & \percentGrad{46}{0}\hspace{0.6em}/33 & \percentGrad{63}{1}\hspace{0.6em}/\textbf{43} & \percentGrad{42}{0}\hspace{0.6em}/36 & \percentGrad{42}{0}\hspace{0.6em}/36 & \percentGrad{42}{0}\hspace{0.6em}/36 & \percentGrad{43}{0}\hspace{0.6em}/33 & \percentGrad{62}{2} \\
        \textbf{p-wm} & \percentGrad{39}{0}\hspace{0.6em}/23 & \percentGrad{52}{0}\hspace{0.6em}/28 & \percentGrad{54}{0}\hspace{0.6em}/27 & \percentGrad{54}{0}\hspace{0.6em}/27 & \percentGrad{54}{0}\hspace{0.6em}/27 & --/-- & \percentGrad{62}{0}\hspace{0.6em}/29 \\
        \textbf{mean} & \percentGrad{27}{0}\hspace{0.6em}/21 & \percentGrad{55}{0}\hspace{0.6em}/29 & \percentGrad{60}{0}\hspace{0.6em}/23 & \percentGrad{60}{0}\hspace{0.6em}/23 & \percentGrad{59}{0}\hspace{0.6em}/22 & --/-- & \percentGrad{62}{2} \\
        \textbf{prompt} & \percentGrad{2}{0}\hspace{0.6em}/-15 & \percentGrad{50}{0}\hspace{0.6em}/31 & \percentGrad{28}{0}\hspace{0.6em}/24 & \percentGrad{28}{0}\hspace{0.6em}/24 & \percentGrad{28}{0}\hspace{0.6em}/24 & --/-- & \percentGrad{62}{2} \\
\bottomrule
\end{tabularx}
\caption{Correlation between the \acrshort{cla} and the \textbf{Flores} translation \textsc{chrF} measured first across the \texttt{src} languages with a fixed \texttt{tgt} language and then averaged over the target languages. The correlation is measured between the \texttt{src-en} (left of the `/' symbol) or \texttt{src-tgt} (right of the `/' symbol) alignment score and the \textsc{chrF} score in the \texttt{tgt} language. Values in per cent. Cell colours are based on the \texttt{src-en} scores.}
\label{tab:corr-both-with-chrf-all-HIGHEST}
\end{table}

%% file: latex/tables/_corr_combination_with_chrf_combined.tex
\setlength{\tabcolsep}{4.5pt}
\setlength{\tabcolsep}{3pt}\begin{table}[ht]
\centering\footnotesize
\begin{tabularx}{\columnwidth}%
{X | c  c  c  c  c  c | c}\toprule
\multicolumn{8}{c}{\textbf{Qwen3-14B-Base}} \\
\midrule
         & \textbf{cos} & \textbf{ANC} & \textbf{abs} & \textbf{dist} & \textbf{ratio} & \textbf{Dali} & \textbf{Eflo} \\
        \midrule
        \textbf{fewshot} & \percentGrad{43}{0}\hspace{0.6em}/0 & \percentGrad{63}{0}\hspace{0.6em}/\textbf{7} & \percentGrad{82}{1}\hspace{0.6em}/0 & \percentGrad{82}{1}\hspace{0.6em}/0 & \percentGrad{82}{1}\hspace{0.6em}/0 & --/-- & \percentGrad{62}{2} \\
        \textbf{last} & \percentGrad{34}{0}\hspace{0.6em}/0 & \percentGrad{59}{0}\hspace{0.6em}/2 & \percentGrad{72}{0}\hspace{0.6em}/0 & \percentGrad{72}{0}\hspace{0.6em}/0 & \percentGrad{72}{0}\hspace{0.6em}/0 & \percentGrad{17}{0}\hspace{0.6em}/0 & \percentGrad{62}{2} \\
        \textbf{p-wm} & \percentGrad{79}{0}\hspace{0.6em}/2 & \percentGrad{82}{1}\hspace{0.6em}/0 & \percentGrad{72}{0}\hspace{0.6em}/0 & \percentGrad{72}{0}\hspace{0.6em}/0 & \percentGrad{72}{0}\hspace{0.6em}/0 & --/-- & \percentGrad{62}{0}\hspace{0.6em}/0 \\
        \textbf{mean} & \percentGrad{61}{0}\hspace{0.6em}/0 & \percentGrad{82}{1}\hspace{0.6em}/0 & \percentGrad{77}{0}\hspace{0.6em}/0 & \percentGrad{77}{0}\hspace{0.6em}/0 & \percentGrad{76}{0}\hspace{0.6em}/0 & --/-- & \percentGrad{62}{2} \\
        \textbf{prompt} & --/-- & \percentGrad{80}{0}\hspace{0.6em}/0 & \percentGrad{67}{0}\hspace{0.6em}/0 & \percentGrad{67}{0}\hspace{0.6em}/0 & \percentGrad{67}{0}\hspace{0.6em}/0 & --/-- & \percentGrad{62}{2} \\
\midrule
\multicolumn{8}{c}{\textbf{Qwen3-14B}} \\
\midrule
         & \textbf{cos} & \textbf{ANC} & \textbf{abs} & \textbf{dist} & \textbf{ratio} & \textbf{Dali} & \textbf{Eflo} \\
        \midrule
        \textbf{fewshot} & --/-- & \percentGrad{80}{0}\hspace{0.6em}/0 & \percentGrad{73}{0}\hspace{0.6em}/2 & \percentGrad{73}{0}\hspace{0.6em}/2 & \percentGrad{73}{0}\hspace{0.6em}/2 & --/-- & \percentGrad{50}{2} \\
        \textbf{last} & \percentGrad{34}{0}\hspace{0.6em}/0 & \percentGrad{56}{0}\hspace{0.6em}/2 & \percentGrad{69}{0}\hspace{0.6em}/0 & \percentGrad{69}{0}\hspace{0.6em}/0 & \percentGrad{69}{0}\hspace{0.6em}/0 & \percentGrad{69}{0}\hspace{0.6em}/1 & \percentGrad{50}{2} \\
        \textbf{p-wm} & \percentGrad{71}{0}\hspace{0.6em}/\textbf{3} & \percentGrad{81}{0}\hspace{0.6em}/0 & \percentGrad{74}{0}\hspace{0.6em}/0 & \percentGrad{74}{0}\hspace{0.6em}/0 & \percentGrad{74}{0}\hspace{0.6em}/0 & --/-- & \percentGrad{50}{0}\hspace{0.6em}/0 \\
        \textbf{mean} & \percentGrad{52}{0}\hspace{0.6em}/0 & \percentGrad{81}{0}\hspace{0.6em}/0 & \percentGrad{73}{0}\hspace{0.6em}/0 & \percentGrad{73}{0}\hspace{0.6em}/0 & \percentGrad{73}{0}\hspace{0.6em}/0 & --/-- & \percentGrad{50}{2} \\
        \textbf{prompt} & --/-- & \percentGrad{82}{1}\hspace{0.6em}/0 & \percentGrad{63}{0}\hspace{0.6em}/0 & \percentGrad{63}{0}\hspace{0.6em}/0 & \percentGrad{63}{0}\hspace{0.6em}/0 & --/-- & \percentGrad{50}{2} \\
\midrule
\multicolumn{8}{c}{\textbf{gemma-3-12b-pt}} \\
\midrule
         & \textbf{cos} & \textbf{ANC} & \textbf{abs} & \textbf{dist} & \textbf{ratio} & \textbf{Dali} & \textbf{Eflo} \\
        \midrule
        \textbf{fewshot} & \percentGrad{45}{0}\hspace{0.6em}/2 & \percentGrad{59}{0}\hspace{0.6em}/3 & \percentGrad{51}{0}\hspace{0.6em}/0 & \percentGrad{51}{0}\hspace{0.6em}/0 & \percentGrad{51}{0}\hspace{0.6em}/0 & --/-- & \percentGrad{69}{2} \\
        \textbf{last} & \percentGrad{22}{0}\hspace{0.6em}/0 & \percentGrad{62}{0}\hspace{0.6em}/5 & \percentGrad{72}{0}\hspace{0.6em}/7 & \percentGrad{72}{0}\hspace{0.6em}/7 & \percentGrad{72}{0}\hspace{0.6em}/7 & \percentGrad{58}{0}\hspace{0.6em}/5 & \percentGrad{69}{2} \\
        \textbf{p-wm} & \percentGrad{65}{0}\hspace{0.6em}/8 & \percentGrad{76}{1}\hspace{0.6em}/9 & \percentGrad{57}{0}\hspace{0.6em}/3 & \percentGrad{57}{0}\hspace{0.6em}/3 & \percentGrad{57}{0}\hspace{0.6em}/3 & --/-- & \percentGrad{69}{0}\hspace{0.6em}/4 \\
        \textbf{mean} & \percentGrad{20}{0}\hspace{0.6em}/-- & \percentGrad{75}{0}\hspace{0.6em}/8 & \percentGrad{52}{0}\hspace{0.6em}/3 & \percentGrad{52}{0}\hspace{0.6em}/3 & \percentGrad{52}{0}\hspace{0.6em}/3 & --/-- & \percentGrad{69}{2} \\
        \textbf{prompt} & --/-- & \percentGrad{74}{0}\hspace{0.6em}/\textbf{14} & \percentGrad{57}{0}\hspace{0.6em}/7 & \percentGrad{57}{0}\hspace{0.6em}/7 & \percentGrad{58}{0}\hspace{0.6em}/7 & --/-- & \percentGrad{69}{2} \\
\midrule
\multicolumn{8}{c}{\textbf{gemma-3-12b-it}} \\
\midrule
         & \textbf{cos} & \textbf{ANC} & \textbf{abs} & \textbf{dist} & \textbf{ratio} & \textbf{Dali} & \textbf{Eflo} \\
        \midrule
        \textbf{fewshot} & \percentGrad{27}{0}\hspace{0.6em}/\textbf{7} & \percentGrad{83}{1}\hspace{0.6em}/5 & \percentGrad{72}{0}\hspace{0.6em}/1 & \percentGrad{72}{0}\hspace{0.6em}/1 & \percentGrad{72}{0}\hspace{0.6em}/1 & --/-- & \percentGrad{70}{2} \\
        \textbf{last} & \percentGrad{54}{0}\hspace{0.6em}/2 & \percentGrad{72}{0}\hspace{0.6em}/1 & \percentGrad{76}{0}\hspace{0.6em}/4 & \percentGrad{76}{0}\hspace{0.6em}/4 & \percentGrad{75}{0}\hspace{0.6em}/4 & \percentGrad{26}{0}\hspace{0.6em}/0 & \percentGrad{70}{2} \\
        \textbf{p-wm} & \percentGrad{71}{0}\hspace{0.6em}/5 & \percentGrad{80}{0}\hspace{0.6em}/3 & \percentGrad{59}{0}\hspace{0.6em}/1 & \percentGrad{59}{0}\hspace{0.6em}/1 & \percentGrad{59}{0}\hspace{0.6em}/1 & --/-- & \percentGrad{70}{0}\hspace{0.6em}/2 \\
        \textbf{mean} & \percentGrad{27}{0}\hspace{0.6em}/-- & \percentGrad{80}{0}\hspace{0.6em}/3 & \percentGrad{55}{0}\hspace{0.6em}/0 & \percentGrad{55}{0}\hspace{0.6em}/0 & \percentGrad{55}{0}\hspace{0.6em}/0 & --/-- & \percentGrad{70}{2} \\
        \textbf{prompt} & --/-- & \percentGrad{79}{0}\hspace{0.6em}/6 & \percentGrad{65}{0}\hspace{0.6em}/2 & \percentGrad{65}{0}\hspace{0.6em}/2 & \percentGrad{65}{0}\hspace{0.6em}/2 & --/-- & \percentGrad{70}{2} \\
\midrule
\multicolumn{8}{c}{\textbf{Ministral-3-14B-Base-2512}} \\
\midrule
         & \textbf{cos} & \textbf{ANC} & \textbf{abs} & \textbf{dist} & \textbf{ratio} & \textbf{Dali} & \textbf{Eflo} \\
        \midrule
        \textbf{fewshot} & \percentGrad{56}{0}\hspace{0.6em}/0 & \percentGrad{80}{0}\hspace{0.6em}/\textbf{3} & \percentGrad{88}{1}\hspace{0.6em}/0 & \percentGrad{88}{1}\hspace{0.6em}/0 & \percentGrad{88}{1}\hspace{0.6em}/0 & --/-- & \percentGrad{63}{2} \\
        \textbf{last} & \percentGrad{82}{0}\hspace{0.6em}/0 & \percentGrad{81}{0}\hspace{0.6em}/1 & \percentGrad{84}{0}\hspace{0.6em}/1 & \percentGrad{84}{0}\hspace{0.6em}/1 & \percentGrad{84}{0}\hspace{0.6em}/1 & \percentGrad{85}{0}\hspace{0.6em}/0 & \percentGrad{63}{2} \\
        \textbf{p-wm} & \percentGrad{84}{0}\hspace{0.6em}/0 & \percentGrad{87}{0}\hspace{0.6em}/0 & \percentGrad{86}{0}\hspace{0.6em}/0 & \percentGrad{86}{0}\hspace{0.6em}/0 & \percentGrad{86}{0}\hspace{0.6em}/0 & --/-- & \percentGrad{63}{0}\hspace{0.6em}/0 \\
        \textbf{mean} & \percentGrad{67}{0}\hspace{0.6em}/0 & \percentGrad{87}{0}\hspace{0.6em}/0 & \percentGrad{85}{0}\hspace{0.6em}/0 & \percentGrad{85}{0}\hspace{0.6em}/0 & \percentGrad{85}{0}\hspace{0.6em}/0 & --/-- & \percentGrad{63}{2} \\
        \textbf{prompt} & --/-- & \percentGrad{87}{0}\hspace{0.6em}/0 & \percentGrad{82}{0}\hspace{0.6em}/0 & \percentGrad{82}{0}\hspace{0.6em}/0 & \percentGrad{82}{0}\hspace{0.6em}/0 & --/-- & \percentGrad{63}{2} \\
\midrule
\multicolumn{8}{c}{\textbf{Ministral-3-14B-Instruct-2512}} \\
\midrule
         & \textbf{cos} & \textbf{ANC} & \textbf{abs} & \textbf{dist} & \textbf{ratio} & \textbf{Dali} & \textbf{Eflo} \\
        \midrule
        \textbf{fewshot} & --/-- & \percentGrad{56}{0}\hspace{0.6em}/\textbf{0} & \percentGrad{35}{0}\hspace{0.6em}/\textbf{0} & \percentGrad{35}{0}\hspace{0.6em}/\textbf{0} & \percentGrad{35}{0}\hspace{0.6em}/\textbf{0} & --/-- & \percentGrad{62}{2} \\
        \textbf{last} & \percentGrad{46}{0}\hspace{0.6em}/\textbf{0} & \percentGrad{63}{1}\hspace{0.6em}/\textbf{0} & \percentGrad{42}{0}\hspace{0.6em}/\textbf{0} & \percentGrad{42}{0}\hspace{0.6em}/\textbf{0} & \percentGrad{42}{0}\hspace{0.6em}/\textbf{0} & \percentGrad{43}{0}\hspace{0.6em}/\textbf{0} & \percentGrad{62}{2} \\
        \textbf{p-wm} & \percentGrad{39}{0}\hspace{0.6em}/\textbf{0} & \percentGrad{52}{0}\hspace{0.6em}/\textbf{0} & \percentGrad{54}{0}\hspace{0.6em}/\textbf{0} & \percentGrad{54}{0}\hspace{0.6em}/\textbf{0} & \percentGrad{54}{0}\hspace{0.6em}/\textbf{0} & --/-- & \percentGrad{62}{0}\hspace{0.6em}/\textbf{0} \\
        \textbf{mean} & \percentGrad{27}{0}\hspace{0.6em}/\textbf{0} & \percentGrad{55}{0}\hspace{0.6em}/\textbf{0} & \percentGrad{60}{0}\hspace{0.6em}/\textbf{0} & \percentGrad{60}{0}\hspace{0.6em}/\textbf{0} & \percentGrad{59}{0}\hspace{0.6em}/\textbf{0} & --/-- & \percentGrad{62}{2} \\
        \textbf{prompt} & \percentGrad{2}{0}\hspace{0.6em}/\textbf{0} & \percentGrad{50}{0}\hspace{0.6em}/\textbf{0} & \percentGrad{28}{0}\hspace{0.6em}/\textbf{0} & \percentGrad{28}{0}\hspace{0.6em}/\textbf{0} & \percentGrad{28}{0}\hspace{0.6em}/\textbf{0} & --/-- & \percentGrad{62}{2} \\
\bottomrule
\end{tabularx}
\caption{Correlation between the \acrshort{cla} and the \textbf{Flores} translation \textsc{chrF} measured first across the \texttt{src} languages with a fixed \texttt{tgt} language and then averaged over the target languages. The correlation is measured between the optimal convex combination of \texttt{src-en} and \texttt{src-tgt} alignment scores and the \textsc{chrF} score in the \texttt{tgt} language. Values are displayed as per cent. The improvement of the combination over the \texttt{en-tgt} (within the same metric-representation) is shown right to the `/' symbol.}
\label{tab:corr-combination-with-chrf-combined}
\end{table}

%% file: latex/tables/corr_pmi_combined.tex
\begin{table}[ht]
\centering\footnotesize
\begin{tabularx}{\columnwidth}%
{p{4em} | C  C  C  C  C  C | C}\toprule
\multicolumn{8}{c}{\textbf{Qwen3-14B-Base}} \\
\midrule
         & \textbf{cos} & \textbf{ANC} & \textbf{abs} & \textbf{dist} & \textbf{ratio} & \textbf{Dali} & \textbf{Eflo} \\
        \midrule
        \textbf{fewshot} & \percentGrad{60.7}{0} & \percentGrad{63.3}{0} & \percentGrad{80.1}{0} & \percentGrad{80.1}{0} & \percentGrad{80.1}{0} & -- & \percentGrad{84.3}{2} \\
        \textbf{last} & \percentGrad{50.2}{0} & \percentGrad{62.3}{0} & \percentGrad{54.2}{0} & \percentGrad{54.2}{0} & \percentGrad{54.4}{0} & \percentGrad{28.1}{0} & \percentGrad{84.3}{2} \\
        \textbf{p-wm} & \percentGrad{91.2}{1} & \percentGrad{81.1}{0} & \percentGrad{57.5}{0} & \percentGrad{57.5}{0} & \percentGrad{57.5}{0} & -- & \percentGrad{84.3}{0} \\
        \textbf{mean} & \percentGrad{85.2}{0} & \percentGrad{79.5}{0} & \percentGrad{90.8}{0} & \percentGrad{90.8}{0} & \percentGrad{90.7}{0} & -- & \percentGrad{84.3}{2} \\
        \textbf{prompt} & -- & \percentGrad{71.8}{0} & \percentGrad{43.0}{0} & \percentGrad{43.0}{0} & \percentGrad{42.9}{0} & -- & \percentGrad{84.3}{2} \\
\midrule
\multicolumn{8}{c}{\textbf{Qwen3-14B}} \\
\midrule
         & \textbf{cos} & \textbf{ANC} & \textbf{abs} & \textbf{dist} & \textbf{ratio} & \textbf{Dali} & \textbf{Eflo} \\
        \midrule
        \textbf{fewshot} & -- & \percentGrad{74.2}{0} & \percentGrad{79.3}{0} & \percentGrad{79.3}{0} & \percentGrad{79.3}{0} & -- & \percentGrad{74.9}{2} \\
        \textbf{last} & \percentGrad{44.8}{0} & \percentGrad{55.6}{0} & \percentGrad{52.0}{0} & \percentGrad{52.0}{0} & \percentGrad{52.0}{0} & \percentGrad{66.5}{0} & \percentGrad{74.9}{2} \\
        \textbf{p-wm} & \percentGrad{82.9}{1} & \percentGrad{75.1}{0} & \percentGrad{55.6}{0} & \percentGrad{55.6}{0} & \percentGrad{55.6}{0} & -- & \percentGrad{74.9}{0} \\
        \textbf{mean} & \percentGrad{74.8}{0} & \percentGrad{74.1}{0} & \percentGrad{79.9}{0} & \percentGrad{79.9}{0} & \percentGrad{79.9}{0} & -- & \percentGrad{74.9}{2} \\
        \textbf{prompt} & -- & \percentGrad{64.0}{0} & \percentGrad{36.2}{0} & \percentGrad{36.2}{0} & \percentGrad{36.3}{0} & -- & \percentGrad{74.9}{2} \\
\midrule
\multicolumn{8}{c}{\textbf{gemma-3-12b-pt}} \\
\midrule
         & \textbf{cos} & \textbf{ANC} & \textbf{abs} & \textbf{dist} & \textbf{ratio} & \textbf{Dali} & \textbf{Eflo} \\
        \midrule
        \textbf{fewshot} & \percentGrad{30.2}{0} & \percentGrad{46.0}{0} & \percentGrad{48.2}{0} & \percentGrad{48.2}{0} & \percentGrad{48.2}{0} & -- & \percentGrad{54.5}{2} \\
        \textbf{last} & \percentGrad{15.8}{0} & \percentGrad{68.0}{1} & \percentGrad{67.3}{0} & \percentGrad{67.3}{0} & \percentGrad{65.7}{0} & \percentGrad{37.8}{0} & \percentGrad{54.5}{2} \\
        \textbf{p-wm} & \percentGrad{64.8}{0} & \percentGrad{42.0}{0} & \percentGrad{53.7}{0} & \percentGrad{53.7}{0} & \percentGrad{53.7}{0} & -- & \percentGrad{54.5}{0} \\
        \textbf{mean} & \percentGrad{2.9}{0} & \percentGrad{41.8}{0} & \percentGrad{56.0}{0} & \percentGrad{56.0}{0} & \percentGrad{56.0}{0} & -- & \percentGrad{54.5}{2} \\
        \textbf{prompt} & -- & \percentGrad{39.4}{0} & \percentGrad{47.5}{0} & \percentGrad{47.5}{0} & \percentGrad{47.4}{0} & -- & \percentGrad{54.5}{2} \\
\midrule
\multicolumn{8}{c}{\textbf{gemma-3-12b-it}} \\
\midrule
         & \textbf{cos} & \textbf{ANC} & \textbf{abs} & \textbf{dist} & \textbf{ratio} & \textbf{Dali} & \textbf{Eflo} \\
        \midrule
        \textbf{fewshot} & \percentGrad{26.4}{0} & \percentGrad{50.7}{0} & \percentGrad{60.7}{0} & \percentGrad{60.7}{0} & \percentGrad{60.7}{0} & -- & \percentGrad{48.0}{2} \\
        \textbf{last} & \percentGrad{55.6}{0} & \percentGrad{59.0}{0} & \percentGrad{67.2}{0} & \percentGrad{67.2}{0} & \percentGrad{67.7}{0} & \percentGrad{32.9}{0} & \percentGrad{48.0}{2} \\
        \textbf{p-wm} & \percentGrad{69.9}{1} & \percentGrad{48.8}{0} & \percentGrad{62.2}{0} & \percentGrad{62.2}{0} & \percentGrad{62.2}{0} & -- & \percentGrad{48.0}{0} \\
        \textbf{mean} & \percentGrad{8.4}{0} & \percentGrad{48.1}{0} & \percentGrad{63.7}{0} & \percentGrad{63.7}{0} & \percentGrad{63.7}{0} & -- & \percentGrad{48.0}{2} \\
        \textbf{prompt} & -- & \percentGrad{50.8}{0} & \percentGrad{38.9}{0} & \percentGrad{38.9}{0} & \percentGrad{38.9}{0} & -- & \percentGrad{48.0}{2} \\
\midrule
\multicolumn{8}{c}{\textbf{Ministral-3-14B-Base-2512}} \\
\midrule
         & \textbf{cos} & \textbf{ANC} & \textbf{abs} & \textbf{dist} & \textbf{ratio} & \textbf{Dali} & \textbf{Eflo} \\
        \midrule
        \textbf{fewshot} & \percentGrad{57.4}{0} & \percentGrad{79.6}{0} & \percentGrad{85.7}{0} & \percentGrad{85.7}{0} & \percentGrad{85.8}{0} & -- & \percentGrad{69.7}{2} \\
        \textbf{last} & \percentGrad{84.6}{0} & \percentGrad{83.9}{0} & \percentGrad{87.0}{0} & \percentGrad{87.0}{0} & \percentGrad{87.1}{0} & \percentGrad{80.8}{0} & \percentGrad{69.7}{2} \\
        \textbf{p-wm} & \percentGrad{81.3}{0} & \percentGrad{84.2}{0} & \percentGrad{90.4}{1} & \percentGrad{90.4}{1} & \percentGrad{90.5}{1} & -- & \percentGrad{69.7}{0} \\
        \textbf{mean} & \percentGrad{63.3}{0} & \percentGrad{83.9}{0} & \percentGrad{88.8}{0} & \percentGrad{88.8}{0} & \percentGrad{88.9}{0} & -- & \percentGrad{69.7}{2} \\
        \textbf{prompt} & -- & \percentGrad{79.6}{0} & \percentGrad{73.2}{0} & \percentGrad{73.2}{0} & \percentGrad{73.1}{0} & -- & \percentGrad{69.7}{2} \\
\midrule
\multicolumn{8}{c}{\textbf{Ministral-3-14B-Instruct-2512}} \\
\midrule
         & \textbf{cos} & \textbf{ANC} & \textbf{abs} & \textbf{dist} & \textbf{ratio} & \textbf{Dali} & \textbf{Eflo} \\
        \midrule
        \textbf{fewshot} & -- & \percentGrad{84.2}{0} & \percentGrad{76.5}{0} & \percentGrad{76.5}{0} & \percentGrad{76.5}{0} & -- & \percentGrad{69.4}{2} \\
        \textbf{last} & \percentGrad{67.0}{0} & \percentGrad{78.4}{0} & \percentGrad{75.1}{0} & \percentGrad{75.1}{0} & \percentGrad{75.1}{0} & \percentGrad{79.1}{0} & \percentGrad{69.4}{2} \\
        \textbf{p-wm} & \percentGrad{79.3}{0} & \percentGrad{82.2}{0} & \percentGrad{88.8}{1} & \percentGrad{88.8}{1} & \percentGrad{88.7}{1} & -- & \percentGrad{69.4}{0} \\
        \textbf{mean} & \percentGrad{60.4}{0} & \percentGrad{82.1}{0} & \percentGrad{85.2}{0} & \percentGrad{85.2}{0} & \percentGrad{85.2}{0} & -- & \percentGrad{69.4}{2} \\
        \textbf{prompt} & \percentGrad{19.3}{0} & \percentGrad{80.8}{0} & \percentGrad{70.5}{0} & \percentGrad{70.5}{0} & \percentGrad{70.5}{0} & -- & \percentGrad{69.4}{2} \\
\bottomrule
\end{tabularx}
    \caption{Correlation between the optimal convex combination of alignments \texttt{src-en}, \texttt{en-tgt} and \texttt{src-tgt} and the \textbf{Flores} translation PMI. Values in per cent.}
    \label{tab:corr-pmi-combined}
\end{table}

%% file: latex/tables/gammas_pmi_all_HIGHEST_pmi.tex
\begin{table}[ht]
\centering\footnotesize
\begin{tabularx}{\columnwidth}%
{p{4em} | C  C  C  C  C  C | C}\toprule
\multicolumn{8}{c}{\textbf{Qwen3-14B-Base}} \\
\midrule
         & \textbf{cos} & \textbf{ANC} & \textbf{abs} & \textbf{dist} & \textbf{ratio} & \textbf{Dali} & \textbf{Eflo} \\
        \midrule
        \textbf{fewshot} & \percentGrad{0.0}{0} & \percentGrad{40.0}{1} & \percentGrad{0.0}{0} & \percentGrad{0.0}{0} & \percentGrad{0.0}{0} & -- & \percentGrad{0.0}{2} \\
        \textbf{last} & \percentGrad{0.0}{0} & \percentGrad{20.0}{0} & \percentGrad{0.0}{0} & \percentGrad{0.0}{0} & \percentGrad{0.0}{0} & \percentGrad{0.0}{0} & \percentGrad{0.0}{2} \\
        \textbf{p-wm} & \percentGrad{0.0}{0} & \percentGrad{0.0}{0} & \percentGrad{0.0}{0} & \percentGrad{0.0}{0} & \percentGrad{0.0}{0} & -- & \percentGrad{0.0}{0} \\
        \textbf{mean} & \percentGrad{0.0}{0} & \percentGrad{0.0}{0} & \percentGrad{24.0}{0} & \percentGrad{24.0}{0} & \percentGrad{24.0}{0} & -- & \percentGrad{0.0}{2} \\
        \textbf{prompt} & -- & \percentGrad{0.0}{0} & \percentGrad{0.0}{0} & \percentGrad{0.0}{0} & \percentGrad{0.0}{0} & -- & \percentGrad{0.0}{2} \\
\midrule
\multicolumn{8}{c}{\textbf{Qwen3-14B}} \\
\midrule
         & \textbf{cos} & \textbf{ANC} & \textbf{abs} & \textbf{dist} & \textbf{ratio} & \textbf{Dali} & \textbf{Eflo} \\
        \midrule
        \textbf{fewshot} & -- & \percentGrad{0.0}{0} & \percentGrad{8.0}{0} & \percentGrad{8.0}{0} & \percentGrad{8.0}{0} & -- & \percentGrad{0.0}{2} \\
        \textbf{last} & \percentGrad{0.0}{0} & \percentGrad{16.0}{0} & \percentGrad{0.0}{0} & \percentGrad{0.0}{0} & \percentGrad{0.0}{0} & \percentGrad{0.0}{0} & \percentGrad{0.0}{2} \\
        \textbf{p-wm} & \percentGrad{8.0}{0} & \percentGrad{0.0}{0} & \percentGrad{0.0}{0} & \percentGrad{0.0}{0} & \percentGrad{0.0}{0} & -- & \percentGrad{0.0}{0} \\
        \textbf{mean} & \percentGrad{0.0}{0} & \percentGrad{0.0}{0} & \percentGrad{36.0}{1} & \percentGrad{36.0}{1} & \percentGrad{36.0}{1} & -- & \percentGrad{0.0}{2} \\
        \textbf{prompt} & -- & \percentGrad{0.0}{0} & \percentGrad{0.0}{0} & \percentGrad{0.0}{0} & \percentGrad{0.0}{0} & -- & \percentGrad{0.0}{2} \\
\midrule
\multicolumn{8}{c}{\textbf{gemma-3-12b-pt}} \\
\midrule
         & \textbf{cos} & \textbf{ANC} & \textbf{abs} & \textbf{dist} & \textbf{ratio} & \textbf{Dali} & \textbf{Eflo} \\
        \midrule
        \textbf{fewshot} & \percentGrad{0.0}{0} & \percentGrad{0.0}{0} & \percentGrad{0.0}{0} & \percentGrad{0.0}{0} & \percentGrad{0.0}{0} & -- & \percentGrad{52.0}{2} \\
        \textbf{last} & \percentGrad{0.0}{0} & \percentGrad{0.0}{0} & \percentGrad{0.0}{0} & \percentGrad{0.0}{0} & \percentGrad{0.0}{0} & \percentGrad{24.0}{0} & \percentGrad{52.0}{2} \\
        \textbf{p-wm} & \percentGrad{0.0}{0} & \percentGrad{0.0}{0} & \percentGrad{0.0}{0} & \percentGrad{0.0}{0} & \percentGrad{0.0}{0} & -- & \percentGrad{52.0}{0} \\
        \textbf{mean} & \percentGrad{72.0}{1} & \percentGrad{0.0}{0} & \percentGrad{0.0}{0} & \percentGrad{0.0}{0} & \percentGrad{0.0}{0} & -- & \percentGrad{52.0}{2} \\
        \textbf{prompt} & -- & \percentGrad{16.0}{0} & \percentGrad{0.0}{0} & \percentGrad{0.0}{0} & \percentGrad{0.0}{0} & -- & \percentGrad{52.0}{2} \\
\midrule
\multicolumn{8}{c}{\textbf{gemma-3-12b-it}} \\
\midrule
         & \textbf{cos} & \textbf{ANC} & \textbf{abs} & \textbf{dist} & \textbf{ratio} & \textbf{Dali} & \textbf{Eflo} \\
        \midrule
        \textbf{fewshot} & \percentGrad{0.0}{0} & \percentGrad{12.0}{0} & \percentGrad{0.0}{0} & \percentGrad{0.0}{0} & \percentGrad{0.0}{0} & -- & \percentGrad{36.0}{2} \\
        \textbf{last} & \percentGrad{12.0}{0} & \percentGrad{0.0}{0} & \percentGrad{0.0}{0} & \percentGrad{0.0}{0} & \percentGrad{0.0}{0} & \percentGrad{0.0}{0} & \percentGrad{36.0}{2} \\
        \textbf{p-wm} & \percentGrad{0.0}{0} & \percentGrad{12.0}{0} & \percentGrad{0.0}{0} & \percentGrad{0.0}{0} & \percentGrad{0.0}{0} & -- & \percentGrad{36.0}{0} \\
        \textbf{mean} & \percentGrad{72.0}{1} & \percentGrad{16.0}{0} & \percentGrad{0.0}{0} & \percentGrad{0.0}{0} & \percentGrad{0.0}{0} & -- & \percentGrad{36.0}{2} \\
        \textbf{prompt} & -- & \percentGrad{0.0}{0} & \percentGrad{0.0}{0} & \percentGrad{0.0}{0} & \percentGrad{0.0}{0} & -- & \percentGrad{36.0}{2} \\
\midrule
\multicolumn{8}{c}{\textbf{Ministral-3-14B-Base-2512}} \\
\midrule
         & \textbf{cos} & \textbf{ANC} & \textbf{abs} & \textbf{dist} & \textbf{ratio} & \textbf{Dali} & \textbf{Eflo} \\
        \midrule
        \textbf{fewshot} & \percentGrad{0.0}{0} & \percentGrad{40.0}{0} & \percentGrad{0.0}{0} & \percentGrad{0.0}{0} & \percentGrad{0.0}{0} & -- & \percentGrad{4.0}{2} \\
        \textbf{last} & \percentGrad{4.0}{0} & \percentGrad{32.0}{0} & \percentGrad{32.0}{0} & \percentGrad{32.0}{0} & \percentGrad{32.0}{0} & \percentGrad{0.0}{0} & \percentGrad{4.0}{2} \\
        \textbf{p-wm} & \percentGrad{0.0}{0} & \percentGrad{0.0}{0} & \percentGrad{4.0}{0} & \percentGrad{4.0}{0} & \percentGrad{4.0}{0} & -- & \percentGrad{4.0}{0} \\
        \textbf{mean} & \percentGrad{4.0}{0} & \percentGrad{0.0}{0} & \percentGrad{24.0}{0} & \percentGrad{24.0}{0} & \percentGrad{24.0}{0} & -- & \percentGrad{4.0}{2} \\
        \textbf{prompt} & -- & \percentGrad{4.0}{0} & \percentGrad{56.0}{0} & \percentGrad{56.0}{0} & \percentGrad{64.0}{1} & -- & \percentGrad{4.0}{2} \\
\midrule
\multicolumn{8}{c}{\textbf{Ministral-3-14B-Instruct-2512}} \\
\midrule
         & \textbf{cos} & \textbf{ANC} & \textbf{abs} & \textbf{dist} & \textbf{ratio} & \textbf{Dali} & \textbf{Eflo} \\
        \midrule
        \textbf{fewshot} & -- & \percentGrad{0.0}{0} & \percentGrad{0.0}{0} & \percentGrad{0.0}{0} & \percentGrad{0.0}{0} & -- & \percentGrad{0.0}{2} \\
        \textbf{last} & \percentGrad{4.0}{0} & \percentGrad{24.0}{0} & \percentGrad{0.0}{0} & \percentGrad{0.0}{0} & \percentGrad{0.0}{0} & \percentGrad{0.0}{0} & \percentGrad{0.0}{2} \\
        \textbf{p-wm} & \percentGrad{0.0}{0} & \percentGrad{0.0}{0} & \percentGrad{0.0}{0} & \percentGrad{0.0}{0} & \percentGrad{0.0}{0} & -- & \percentGrad{0.0}{0} \\
        \textbf{mean} & \percentGrad{0.0}{0} & \percentGrad{0.0}{0} & \percentGrad{24.0}{0} & \percentGrad{24.0}{0} & \percentGrad{24.0}{0} & -- & \percentGrad{0.0}{2} \\
        \textbf{prompt} & \percentGrad{0.0}{0} & \percentGrad{0.0}{0} & \percentGrad{32.0}{1} & \percentGrad{32.0}{1} & \percentGrad{32.0}{1} & -- & \percentGrad{0.0}{2} \\
\bottomrule
\end{tabularx}
\caption{Optimal convex-combination weight for the \texttt{src-tgt} alignment ($1-\alpha-\beta$) for predicting the \textbf{Flores} translation PMI scores for all models. Values are shown as per cent.}
\label{gammas-pmi-all-HIGHEST-pmi}
\end{table}

%% file: latex/tables/corr_combination_with_chrf_all_-HIGHEST-validation.tex
\setlength{\tabcolsep}{3pt}\begin{table}[ht]
\centering\footnotesize
\begin{tabularx}{\columnwidth}%
{X | c  c  c  c  c  c | c}\toprule
\multicolumn{8}{c}{\textbf{Qwen3-14B-Base}} \\
\midrule
         & \textbf{cos} & \textbf{ANC} & \textbf{abs} & \textbf{dist} & \textbf{ratio} & \textbf{Dali} & \textbf{Eflo} \\
        \midrule
        \textbf{fewshot} & \percentGrad{33}{0}\hspace{0.6em}/0 & \percentGrad{61}{0}\hspace{0.6em}/\textbf{13} & \percentGrad{73}{0}\hspace{0.6em}/1 & \percentGrad{73}{0}\hspace{0.6em}/1 & \percentGrad{73}{0}\hspace{0.6em}/0 & --/-- & \percentGrad{51}{2} \\
        \textbf{last} & \percentGrad{23}{0}\hspace{0.6em}/0 & \percentGrad{55}{0}\hspace{0.6em}/6 & \percentGrad{69}{0}\hspace{0.6em}/0 & \percentGrad{69}{0}\hspace{0.6em}/0 & \percentGrad{69}{0}\hspace{0.6em}/0 & \percentGrad{23}{0}\hspace{0.6em}/0 & \percentGrad{51}{2} \\
        \textbf{p-wm} & \percentGrad{70}{0}\hspace{0.6em}/4 & \percentGrad{75}{0}\hspace{0.6em}/2 & \percentGrad{68}{0}\hspace{0.6em}/0 & \percentGrad{68}{0}\hspace{0.6em}/0 & \percentGrad{68}{0}\hspace{0.6em}/0 & --/-- & \percentGrad{51}{0}\hspace{0.6em}/0 \\
        \textbf{mean} & \percentGrad{50}{0}\hspace{0.6em}/1 & \percentGrad{74}{0}\hspace{0.6em}/1 & \percentGrad{70}{0}\hspace{0.6em}/0 & \percentGrad{70}{0}\hspace{0.6em}/0 & \percentGrad{70}{0}\hspace{0.6em}/0 & --/-- & \percentGrad{51}{2} \\
        \textbf{prompt} & --/-- & \percentGrad{76}{1}\hspace{0.6em}/0 & \percentGrad{65}{0}\hspace{0.6em}/0 & \percentGrad{65}{0}\hspace{0.6em}/0 & \percentGrad{65}{0}\hspace{0.6em}/0 & --/-- & \percentGrad{51}{2} \\
\midrule
\multicolumn{8}{c}{\textbf{Qwen3-14B}} \\
\midrule
         & \textbf{cos} & \textbf{ANC} & \textbf{abs} & \textbf{dist} & \textbf{ratio} & \textbf{Dali} & \textbf{Eflo} \\
        \midrule
        \textbf{fewshot} & --/-- & \percentGrad{69}{0}\hspace{0.6em}/1 & \percentGrad{61}{0}\hspace{0.6em}/2 & \percentGrad{61}{0}\hspace{0.6em}/2 & \percentGrad{61}{0}\hspace{0.6em}/2 & --/-- & \percentGrad{37}{2} \\
        \textbf{last} & \percentGrad{23}{0}\hspace{0.6em}/0 & \percentGrad{48}{0}\hspace{0.6em}/\textbf{3} & \percentGrad{61}{0}\hspace{0.6em}/0 & \percentGrad{61}{0}\hspace{0.6em}/0 & \percentGrad{61}{0}\hspace{0.6em}/0 & \percentGrad{59}{0}\hspace{0.6em}/1 & \percentGrad{37}{2} \\
        \textbf{p-wm} & \percentGrad{57}{0}\hspace{0.6em}/\textbf{3} & \percentGrad{70}{0}\hspace{0.6em}/1 & \percentGrad{65}{0}\hspace{0.6em}/0 & \percentGrad{65}{0}\hspace{0.6em}/0 & \percentGrad{65}{0}\hspace{0.6em}/0 & --/-- & \percentGrad{37}{0}\hspace{0.6em}/0 \\
        \textbf{mean} & \percentGrad{39}{0}\hspace{0.6em}/0 & \percentGrad{70}{0}\hspace{0.6em}/1 & \percentGrad{62}{0}\hspace{0.6em}/0 & \percentGrad{62}{0}\hspace{0.6em}/0 & \percentGrad{62}{0}\hspace{0.6em}/0 & --/-- & \percentGrad{37}{2} \\
        \textbf{prompt} & --/-- & \percentGrad{72}{1}\hspace{0.6em}/0 & \percentGrad{59}{0}\hspace{0.6em}/0 & \percentGrad{59}{0}\hspace{0.6em}/0 & \percentGrad{59}{0}\hspace{0.6em}/0 & --/-- & \percentGrad{37}{2} \\
\midrule
\multicolumn{8}{c}{\textbf{gemma-3-12b-pt}} \\
\midrule
         & \textbf{cos} & \textbf{ANC} & \textbf{abs} & \textbf{dist} & \textbf{ratio} & \textbf{Dali} & \textbf{Eflo} \\
        \midrule
        \textbf{fewshot} & \percentGrad{47}{0}\hspace{0.6em}/0 & \percentGrad{60}{0}\hspace{0.6em}/4 & \percentGrad{50}{0}\hspace{0.6em}/0 & \percentGrad{50}{0}\hspace{0.6em}/0 & \percentGrad{50}{0}\hspace{0.6em}/0 & --/-- & \percentGrad{52}{2} \\
        \textbf{last} & \percentGrad{19}{0}\hspace{0.6em}/0 & \percentGrad{42}{0}\hspace{0.6em}/6 & \percentGrad{58}{0}\hspace{0.6em}/4 & \percentGrad{58}{0}\hspace{0.6em}/4 & \percentGrad{58}{0}\hspace{0.6em}/4 & \percentGrad{62}{0}\hspace{0.6em}/2 & \percentGrad{52}{2} \\
        \textbf{p-wm} & \percentGrad{58}{0}\hspace{0.6em}/6 & \percentGrad{67}{1}\hspace{0.6em}/5 & \percentGrad{48}{0}\hspace{0.6em}/2 & \percentGrad{48}{0}\hspace{0.6em}/2 & \percentGrad{48}{0}\hspace{0.6em}/2 & --/-- & \percentGrad{52}{0}\hspace{0.6em}/1 \\
        \textbf{mean} & \percentGrad{18}{0}\hspace{0.6em}/-- & \percentGrad{66}{0}\hspace{0.6em}/4 & \percentGrad{40}{0}\hspace{0.6em}/1 & \percentGrad{40}{0}\hspace{0.6em}/1 & \percentGrad{40}{0}\hspace{0.6em}/1 & --/-- & \percentGrad{52}{2} \\
        \textbf{prompt} & --/-- & \percentGrad{58}{0}\hspace{0.6em}/\textbf{18} & \percentGrad{44}{0}\hspace{0.6em}/3 & \percentGrad{44}{0}\hspace{0.6em}/3 & \percentGrad{44}{0}\hspace{0.6em}/3 & --/-- & \percentGrad{52}{2} \\
\midrule
\multicolumn{8}{c}{\textbf{gemma-3-12b-it}} \\
\midrule
         & \textbf{cos} & \textbf{ANC} & \textbf{abs} & \textbf{dist} & \textbf{ratio} & \textbf{Dali} & \textbf{Eflo} \\
        \midrule
        \textbf{fewshot} & \percentGrad{40}{0}\hspace{0.6em}/\textbf{3} & \percentGrad{75}{1}\hspace{0.6em}/0 & \percentGrad{71}{0}\hspace{0.6em}/0 & \percentGrad{71}{0}\hspace{0.6em}/0 & \percentGrad{71}{0}\hspace{0.6em}/0 & --/-- & \percentGrad{57}{2} \\
        \textbf{last} & \percentGrad{57}{0}\hspace{0.6em}/1 & \percentGrad{55}{0}\hspace{0.6em}/-1 & \percentGrad{64}{0}\hspace{0.6em}/0 & \percentGrad{64}{0}\hspace{0.6em}/0 & \percentGrad{64}{0}\hspace{0.6em}/0 & \percentGrad{9}{0}\hspace{0.6em}/-1 & \percentGrad{57}{2} \\
        \textbf{p-wm} & \percentGrad{68}{0}\hspace{0.6em}/0 & \percentGrad{72}{0}\hspace{0.6em}/-2 & \percentGrad{52}{0}\hspace{0.6em}/-2 & \percentGrad{52}{0}\hspace{0.6em}/-2 & \percentGrad{52}{0}\hspace{0.6em}/-2 & --/-- & \percentGrad{57}{0}\hspace{0.6em}/0 \\
        \textbf{mean} & \percentGrad{24}{0}\hspace{0.6em}/-- & \percentGrad{71}{0}\hspace{0.6em}/-2 & \percentGrad{49}{0}\hspace{0.6em}/-1 & \percentGrad{49}{0}\hspace{0.6em}/-1 & \percentGrad{49}{0}\hspace{0.6em}/-1 & --/-- & \percentGrad{57}{2} \\
        \textbf{prompt} & --/-- & \percentGrad{71}{0}\hspace{0.6em}/-2 & \percentGrad{51}{0}\hspace{0.6em}/-4 & \percentGrad{51}{0}\hspace{0.6em}/-4 & \percentGrad{51}{0}\hspace{0.6em}/-4 & --/-- & \percentGrad{57}{2} \\
\midrule
\multicolumn{8}{c}{\textbf{Ministral-3-14B-Base-2512}} \\
\midrule
         & \textbf{cos} & \textbf{ANC} & \textbf{abs} & \textbf{dist} & \textbf{ratio} & \textbf{Dali} & \textbf{Eflo} \\
        \midrule
        \textbf{fewshot} & \percentGrad{55}{0}\hspace{0.6em}/0 & \percentGrad{80}{0}\hspace{0.6em}/\textbf{3} & \percentGrad{85}{1}\hspace{0.6em}/0 & \percentGrad{85}{1}\hspace{0.6em}/0 & \percentGrad{85}{1}\hspace{0.6em}/0 & --/-- & \percentGrad{60}{2} \\
        \textbf{last} & \percentGrad{79}{0}\hspace{0.6em}/0 & \percentGrad{80}{0}\hspace{0.6em}/2 & \percentGrad{81}{0}\hspace{0.6em}/2 & \percentGrad{81}{0}\hspace{0.6em}/2 & \percentGrad{81}{0}\hspace{0.6em}/2 & \percentGrad{83}{0}\hspace{0.6em}/0 & \percentGrad{60}{2} \\
        \textbf{p-wm} & \percentGrad{79}{0}\hspace{0.6em}/0 & \percentGrad{84}{0}\hspace{0.6em}/0 & \percentGrad{83}{0}\hspace{0.6em}/1 & \percentGrad{83}{0}\hspace{0.6em}/1 & \percentGrad{83}{0}\hspace{0.6em}/1 & --/-- & \percentGrad{60}{0}\hspace{0.6em}/0 \\
        \textbf{mean} & \percentGrad{63}{0}\hspace{0.6em}/0 & \percentGrad{84}{0}\hspace{0.6em}/0 & \percentGrad{81}{0}\hspace{0.6em}/1 & \percentGrad{81}{0}\hspace{0.6em}/1 & \percentGrad{81}{0}\hspace{0.6em}/1 & --/-- & \percentGrad{60}{2} \\
        \textbf{prompt} & --/-- & \percentGrad{82}{0}\hspace{0.6em}/0 & \percentGrad{77}{0}\hspace{0.6em}/0 & \percentGrad{77}{0}\hspace{0.6em}/0 & \percentGrad{77}{0}\hspace{0.6em}/0 & --/-- & \percentGrad{60}{2} \\
\midrule
\multicolumn{8}{c}{\textbf{Ministral-3-14B-Instruct-2512}} \\
\midrule
         & \textbf{cos} & \textbf{ANC} & \textbf{abs} & \textbf{dist} & \textbf{ratio} & \textbf{Dali} & \textbf{Eflo} \\
        \midrule
        \textbf{fewshot} & --/-- & \percentGrad{59}{1}\hspace{0.6em}/\textbf{0} & \percentGrad{46}{0}\hspace{0.6em}/\textbf{0} & \percentGrad{46}{0}\hspace{0.6em}/\textbf{0} & \percentGrad{46}{0}\hspace{0.6em}/\textbf{0} & --/-- & \percentGrad{45}{2} \\
        \textbf{last} & \percentGrad{44}{0}\hspace{0.6em}/\textbf{0} & \percentGrad{54}{0}\hspace{0.6em}/\textbf{0} & \percentGrad{49}{0}\hspace{0.6em}/\textbf{0} & \percentGrad{49}{0}\hspace{0.6em}/\textbf{0} & \percentGrad{49}{0}\hspace{0.6em}/\textbf{0} & \percentGrad{46}{0}\hspace{0.6em}/\textbf{0} & \percentGrad{45}{2} \\
        \textbf{p-wm} & \percentGrad{48}{0}\hspace{0.6em}/\textbf{0} & \percentGrad{57}{0}\hspace{0.6em}/\textbf{0} & \percentGrad{58}{0}\hspace{0.6em}/\textbf{0} & \percentGrad{58}{0}\hspace{0.6em}/\textbf{0} & \percentGrad{57}{0}\hspace{0.6em}/\textbf{0} & --/-- & \percentGrad{45}{0}\hspace{0.6em}/\textbf{0} \\
        \textbf{mean} & \percentGrad{39}{0}\hspace{0.6em}/\textbf{0} & \percentGrad{59}{1}\hspace{0.6em}/\textbf{0} & \percentGrad{57}{0}\hspace{0.6em}/\textbf{0} & \percentGrad{57}{0}\hspace{0.6em}/\textbf{0} & \percentGrad{57}{0}\hspace{0.6em}/\textbf{0} & --/-- & \percentGrad{45}{2} \\
        \textbf{prompt} & \percentGrad{1}{0}\hspace{0.6em}/\textbf{0} & \percentGrad{57}{0}\hspace{0.6em}/\textbf{0} & \percentGrad{43}{0}\hspace{0.6em}/\textbf{0} & \percentGrad{43}{0}\hspace{0.6em}/\textbf{0} & \percentGrad{43}{0}\hspace{0.6em}/\textbf{0} & --/-- & \percentGrad{45}{2} \\
\bottomrule
\end{tabularx}
\caption{Validation counterpart to Table~\ref{tab:corr-combination-with-chrf-combined}. The optimal convex combination weights are evaluated on the \textbf{BOUQuET} dataset. Right to the `/' symbol is the improvement against using \texttt{src-en} alone.}
\label{tab:corr-combination-with-chrf-all-HIGHEST-validation}
\end{table}

%% file: latex/tables/validation_pmi_all_HIGHEST_pmi.tex
\begin{table}[ht]
\centering\footnotesize
\begin{tabularx}{\columnwidth}%
{p{4em} | C  C  C  C  C  C | C}\toprule
\multicolumn{8}{c}{\textbf{Qwen3-14B-Base}} \\
\midrule
         & \textbf{cos} & \textbf{ANC} & \textbf{abs} & \textbf{dist} & \textbf{ratio} & \textbf{Dali} & \textbf{Eflo} \\
        \midrule
        \textbf{fewshot} & \percentGrad{58.1}{0} & \percentGrad{62.6}{0} & \percentGrad{79.0}{0} & \percentGrad{79.0}{0} & \percentGrad{79.0}{0} & -- & \percentGrad{81.5}{2} \\
        \textbf{last} & \percentGrad{43.6}{0} & \percentGrad{59.3}{0} & \percentGrad{51.6}{0} & \percentGrad{51.6}{0} & \percentGrad{51.7}{0} & \percentGrad{31.0}{0} & \percentGrad{81.5}{2} \\
        \textbf{p-wm} & \percentGrad{91.1}{1} & \percentGrad{82.1}{0} & \percentGrad{56.7}{0} & \percentGrad{56.7}{0} & \percentGrad{56.7}{0} & -- & \percentGrad{81.5}{0} \\
        \textbf{mean} & \percentGrad{84.4}{0} & \percentGrad{80.6}{0} & \percentGrad{91.2}{1} & \percentGrad{91.2}{1} & \percentGrad{91.1}{1} & -- & \percentGrad{81.5}{2} \\
        \textbf{prompt} & -- & \percentGrad{74.5}{0} & \percentGrad{44.2}{0} & \percentGrad{44.2}{0} & \percentGrad{44.1}{0} & -- & \percentGrad{81.5}{2} \\
\midrule
\multicolumn{8}{c}{\textbf{Qwen3-14B}} \\
\midrule
         & \textbf{cos} & \textbf{ANC} & \textbf{abs} & \textbf{dist} & \textbf{ratio} & \textbf{Dali} & \textbf{Eflo} \\
        \midrule
        \textbf{fewshot} & -- & \percentGrad{80.4}{0} & \percentGrad{83.9}{0} & \percentGrad{83.9}{0} & \percentGrad{83.9}{0} & -- & \percentGrad{79.0}{2} \\
        \textbf{last} & \percentGrad{45.5}{0} & \percentGrad{62.4}{0} & \percentGrad{54.6}{0} & \percentGrad{54.6}{0} & \percentGrad{54.6}{0} & \percentGrad{72.3}{0} & \percentGrad{79.0}{2} \\
        \textbf{p-wm} & \percentGrad{88.5}{1} & \percentGrad{81.7}{0} & \percentGrad{56.8}{0} & \percentGrad{56.8}{0} & \percentGrad{56.8}{0} & -- & \percentGrad{79.0}{0} \\
        \textbf{mean} & \percentGrad{76.9}{0} & \percentGrad{80.8}{0} & \percentGrad{85.0}{0} & \percentGrad{85.0}{0} & \percentGrad{84.9}{0} & -- & \percentGrad{79.0}{2} \\
        \textbf{prompt} & -- & \percentGrad{68.5}{0} & \percentGrad{37.7}{0} & \percentGrad{37.7}{0} & \percentGrad{37.8}{0} & -- & \percentGrad{79.0}{2} \\
\midrule
\multicolumn{8}{c}{\textbf{gemma-3-12b-pt}} \\
\midrule
         & \textbf{cos} & \textbf{ANC} & \textbf{abs} & \textbf{dist} & \textbf{ratio} & \textbf{Dali} & \textbf{Eflo} \\
        \midrule
        \textbf{fewshot} & \percentGrad{32.5}{0} & \percentGrad{50.1}{0} & \percentGrad{52.6}{0} & \percentGrad{52.6}{0} & \percentGrad{52.6}{0} & -- & \percentGrad{51.4}{2} \\
        \textbf{last} & \percentGrad{19.0}{0} & \percentGrad{56.1}{0} & \percentGrad{64.9}{0} & \percentGrad{64.9}{0} & \percentGrad{63.7}{0} & \percentGrad{45.8}{0} & \percentGrad{51.4}{2} \\
        \textbf{p-wm} & \percentGrad{66.2}{1} & \percentGrad{46.9}{0} & \percentGrad{52.9}{0} & \percentGrad{52.9}{0} & \percentGrad{52.9}{0} & -- & \percentGrad{51.4}{0} \\
        \textbf{mean} & \percentGrad{5.2}{0} & \percentGrad{46.6}{0} & \percentGrad{53.0}{0} & \percentGrad{53.0}{0} & \percentGrad{53.0}{0} & -- & \percentGrad{51.4}{2} \\
        \textbf{prompt} & -- & \percentGrad{36.8}{0} & \percentGrad{49.4}{0} & \percentGrad{49.4}{0} & \percentGrad{48.9}{0} & -- & \percentGrad{51.4}{2} \\
\midrule
\multicolumn{8}{c}{\textbf{gemma-3-12b-it}} \\
\midrule
         & \textbf{cos} & \textbf{ANC} & \textbf{abs} & \textbf{dist} & \textbf{ratio} & \textbf{Dali} & \textbf{Eflo} \\
        \midrule
        \textbf{fewshot} & \percentGrad{15.3}{0} & \percentGrad{43.9}{0} & \percentGrad{56.5}{0} & \percentGrad{56.5}{0} & \percentGrad{56.5}{0} & -- & \percentGrad{38.3}{2} \\
        \textbf{last} & \percentGrad{53.6}{0} & \percentGrad{39.3}{0} & \percentGrad{62.4}{0} & \percentGrad{62.4}{0} & \percentGrad{62.4}{0} & \percentGrad{25.9}{0} & \percentGrad{38.3}{2} \\
        \textbf{p-wm} & \percentGrad{66.8}{1} & \percentGrad{49.9}{0} & \percentGrad{53.8}{0} & \percentGrad{53.8}{0} & \percentGrad{53.8}{0} & -- & \percentGrad{38.3}{0} \\
        \textbf{mean} & \percentGrad{9.1}{0} & \percentGrad{49.4}{0} & \percentGrad{52.1}{0} & \percentGrad{52.1}{0} & \percentGrad{52.1}{0} & -- & \percentGrad{38.3}{2} \\
        \textbf{prompt} & -- & \percentGrad{56.0}{0} & \percentGrad{50.4}{0} & \percentGrad{50.4}{0} & \percentGrad{50.4}{0} & -- & \percentGrad{38.3}{2} \\
\midrule
\multicolumn{8}{c}{\textbf{Ministral-3-14B-Base-2512}} \\
\midrule
         & \textbf{cos} & \textbf{ANC} & \textbf{abs} & \textbf{dist} & \textbf{ratio} & \textbf{Dali} & \textbf{Eflo} \\
        \midrule
        \textbf{fewshot} & \percentGrad{56.5}{0} & \percentGrad{80.8}{0} & \percentGrad{86.4}{0} & \percentGrad{86.4}{0} & \percentGrad{86.4}{0} & -- & \percentGrad{67.2}{2} \\
        \textbf{last} & \percentGrad{82.0}{0} & \percentGrad{84.0}{0} & \percentGrad{85.6}{0} & \percentGrad{85.6}{0} & \percentGrad{85.6}{0} & \percentGrad{80.7}{0} & \percentGrad{67.2}{2} \\
        \textbf{p-wm} & \percentGrad{78.3}{0} & \percentGrad{84.2}{0} & \percentGrad{88.8}{1} & \percentGrad{88.8}{1} & \percentGrad{88.9}{1} & -- & \percentGrad{67.2}{0} \\
        \textbf{mean} & \percentGrad{60.9}{0} & \percentGrad{83.9}{0} & \percentGrad{87.6}{0} & \percentGrad{87.6}{0} & \percentGrad{87.7}{0} & -- & \percentGrad{67.2}{2} \\
        \textbf{prompt} & -- & \percentGrad{78.0}{0} & \percentGrad{70.9}{0} & \percentGrad{70.9}{0} & \percentGrad{70.9}{0} & -- & \percentGrad{67.2}{2} \\
\midrule
\multicolumn{8}{c}{\textbf{Ministral-3-14B-Instruct-2512}} \\
\midrule
         & \textbf{cos} & \textbf{ANC} & \textbf{abs} & \textbf{dist} & \textbf{ratio} & \textbf{Dali} & \textbf{Eflo} \\
        \midrule
        \textbf{fewshot} & -- & \percentGrad{75.4}{0} & \percentGrad{65.0}{0} & \percentGrad{65.0}{0} & \percentGrad{65.0}{0} & -- & \percentGrad{65.6}{2} \\
        \textbf{last} & \percentGrad{58.5}{0} & \percentGrad{71.6}{0} & \percentGrad{64.4}{0} & \percentGrad{64.4}{0} & \percentGrad{64.4}{0} & \percentGrad{70.2}{0} & \percentGrad{65.6}{2} \\
        \textbf{p-wm} & \percentGrad{66.5}{0} & \percentGrad{75.1}{0} & \percentGrad{81.1}{1} & \percentGrad{81.1}{1} & \percentGrad{81.1}{1} & -- & \percentGrad{65.6}{0} \\
        \textbf{mean} & \percentGrad{49.3}{0} & \percentGrad{76.1}{0} & \percentGrad{79.3}{0} & \percentGrad{79.3}{0} & \percentGrad{79.2}{0} & -- & \percentGrad{65.6}{2} \\
        \textbf{prompt} & \percentGrad{6.2}{0} & \percentGrad{73.5}{0} & \percentGrad{58.4}{0} & \percentGrad{58.4}{0} & \percentGrad{58.4}{0} & -- & \percentGrad{65.6}{2} \\
\bottomrule
\end{tabularx}
\caption{Validation counterpart to Table~\ref{tab:corr-pmi-combined}. The optimal convex combination weights are evaluated on the \textbf{BOUQuET} dataset.}
\label{tab:validation-pmi-all-HIGHEST-pmi}
\end{table}